\documentclass[]{techreport}

\usepackage[utf8]{inputenc}
\usepackage[T1]{fontenc}
\usepackage{xspace}
\usepackage{graphicx}
\usepackage{amsmath}
\usepackage{amssymb}
\usepackage{booktabs}
\usepackage{url}
\usepackage{amsfonts}
\usepackage{nicefrac}
\usepackage{microtype}
\usepackage{xcolor}

\newcommand{\todo}[1]{{\color{red}#1}}
\newcommand{\TODO}[1]{\textbf{\color{red}[TODO: #1]}}

\usepackage{adjustbox}
\usepackage{graphicx}
\usepackage{caption}
\usepackage{subcaption}
\usepackage{printlen}
\usepackage{float}
\usepackage{wrapfig}
\usepackage{multirow}
\usepackage{booktabs}
\usepackage{pifont}
\usepackage{makecell}
\usepackage{nicefrac}
\usepackage{url}
\usepackage{yhmath}
\usepackage{stmaryrd}
\usepackage{verbatim}
\usepackage{enumitem}
\usepackage{amsmath}
\usepackage{lipsum}
\usepackage{algorithm}
\usepackage{algpseudocode}
\usepackage{amsfonts}
\usepackage{microtype}
\usepackage{pgf}
\usepackage{pgfplots}
\usepackage{tabularx}
\usepackage{scalerel}
\usepackage{xspace}
\usepackage{ifthen}
\usepackage{soul}
\usepackage{cuted}
\usepackage{mathtools}
\usepackage{tikz}
\usetikzlibrary{angles,quotes,3d,math,arrows.meta,calc,positioning,fit,backgrounds,decorations.pathreplacing,calligraphy,shapes,shapes.multipart}
\usepackage[table]{xcolor}
\usepackage{lineno}

\usepackage{multirow}
\usepackage{booktabs}
\usepackage{array}
\usepackage{rotating}
\usepackage[labelfont=bf]{caption}
\usepackage{afterpage}
\usepackage{xr}
\usepackage{cancel}

\newif\ifeccv
\eccvfalse

\newcommand{\methodname}{RayDer\xspace}

\newcommand{\inapprox}{\mathrel{\tilde{\in}}}

\ifeccv
    \newcommand{\smalltableheaderfont}{\scriptsize}
\else
    \newcommand{\smalltableheaderfont}{\footnotesize}
\fi

\newcommand{\image}{\mathbf{I}}
\newcommand{\images}{\mathcal{I}}
\newcommand{\pose}{\mathbf{p}}
\newcommand{\state}{\mathbf{s}}
\newcommand{\model}{\mathcal{M}}
\newcommand{\config}[1]{\mbox{\textsc{Config} \textbf{#1}}}

\ifeccv
    \newcommand{\samebf}[1]{\begingroup\sbox0{#1}\sbox2{\textbf{#1}}\resizebox{\wd0}{\ht2}{\usebox2}\endgroup}
\else
    \newcommand{\samebf}[1]{\textbf{#1}}
\fi

\makeatletter
\newcommand{\cancelasX}[1]{\mathpalette\cancelasX@{#1}}
\newcommand{\cancelasX@}[2]{%
  \ooalign{%
    \hfil$\m@th#1\cancel{\phantom{x}}$\hfil\cr
    $\m@th#1#2$\cr
  }%
}
\makeatother

\newcommand{\cmark}{\ding{51}}
\newcommand{\xmark}{\ding{55}}

\ifeccv\else
    \setlist[itemize]{nosep}
\fi

\renewcommand{\todo}[1]{{\color{red}#1}}
\renewcommand{\TODO}[1]{\textbf{\color{red}[TODO: #1]}}

\newcommand{\stefan}[1]{\textbf{{\color{ourgreen}[Stefan: #1]}}}
\newcommand{\uli}[1]{\textbf{{\color{ourblue}[Uli: #1]}}}
\newcommand{\nick}[1]{\textbf{{\color{ourturquoise}[Nick: #1]}}}
\newcommand{\olya}[1]{\textbf{{\color{ourturquoise}[Olya: #1]}}}
\newcommand{\kosta}[1]{\textbf{{\color{ourpurple}[Kosta: #1]}}}
\newcommand{\bjoern}[1]{\textbf{{\color{ourpurple}[Björn: #1]}}}

\newcommand{\remove}[1]{{\color{ourorange}#1}}
\newcommand{\draft}[1]{{\color{gray}#1}}
\newcommand{\compactversion}[1]{#1}

\renewcommand{\TODO}[1]{}
\renewcommand{\todo}[1]{#1}
\renewcommand{\uli}[1]{}
\renewcommand{\stefan}[1]{}
\renewcommand{\nick}[1]{}
\renewcommand{\olya}[1]{}
\renewcommand{\kosta}[1]{}
\renewcommand{\bjoern}[1]{}
\renewcommand{\compactversion}[1]{}
\renewcommand{\draft}[1]{}
\renewcommand{\remove}[1]{}

\definecolor{ourgreen}{RGB}{46, 204, 113}
\definecolor{ourgreenborder}{RGB}{39, 174, 96}
\definecolor{ourblue}{RGB}{52, 152, 219}
\definecolor{ourblueborder}{RGB}{41, 128, 185}
\definecolor{ourorange}{RGB}{230, 126, 34}
\definecolor{ourorangeborder}{RGB}{211, 84, 0}
\definecolor{ourred}{RGB}{231, 76, 60}
\definecolor{ourredborder}{RGB}{192, 57, 43}
\definecolor{ouryellow}{RGB}{241, 196, 15}
\definecolor{ouryellowborder}{RGB}{243, 156, 18}
\definecolor{ourpurple}{RGB}{155, 89, 182}
\definecolor{ourpurpleborder}{RGB}{142, 68, 173}
\definecolor{ourturquoise}{RGB}{26, 188, 156}
\definecolor{ourturquoiseborder}{RGB}{22, 160, 133}
\definecolor{ourturquoise}{RGB}{26, 188, 156}
\definecolor{ourturquoiseborder}{RGB}{22, 160, 133}
\definecolor{ourwhite}{RGB}{236, 240, 241}
\definecolor{ourwhiteborder}{RGB}{189, 195, 199}
\definecolor{ourgray}{RGB}{149, 165, 166}
\definecolor{ourgrayborder}{RGB}{127, 140, 141}

\definecolor{ourwhite2}{RGB}{246, 247, 248}

\definecolor{matplotlibblue}{HTML}{1f77b4}
\definecolor{matplotliborange}{HTML}{ff7f0e}
\definecolor{matplotlibgreen}{HTML}{2ca02c}

\definecolor{ourhighlightcolor}{RGB}{46, 204, 113}

\newcommand{\shorttabulara}[1]{\begin{tabular}{c}#1\\[-1mm]\\\end{tabular}}

\newcolumntype{H}{>{\setbox0=\hbox\bgroup}c<{\egroup}@{}}

\newcommand{\tikzstylenodedistance}{4mm}
\newcommand{\tikzstyleinnersep}{2mm}
\newcommand{\tikzstyleminimumheight}{8.75mm}
\newcommand{\tikzstyleminimumwidth}{12mm}

\tikzset{
    node distance=\tikzstylenodedistance,
    text centered,
    anchor=center,
}
\tikzset{
    standard node/.style n args={1}{%
        rectangle,
        rounded corners=0.1cm,
        fill=our#1,
        draw=our#1border,
        line width=0.04cm,
        minimum height=\tikzstyleminimumheight,
        minimum width=\tikzstyleminimumwidth,
        inner sep=\tikzstyleinnersep,
        text centered,
        anchor=center,
        align=center,
    }
}
\tikzset{
    standard node module/.style n args={0}{%
        rectangle,
        rounded corners=0.1cm,
        fill=ourturquoise,
        draw=ourturquoiseborder,
        line width=0.04cm,
        minimum height=\tikzstyleminimumheight,
        minimum width=12mm,
        inner xsep=\tikzstyleinnersep,
        inner ysep=1mm,
        text centered,
        anchor=center,
        align=center,
    }
}
\tikzset{
    standard node image/.style n args={1}{%
        rectangle,
        fill=our#1,
        draw=our#1border,
        line width=0.04cm,
        minimum height=\tikzstyleminimumheight,
        minimum width=\tikzstyleminimumwidth,
        inner sep=0,
        text centered,
        anchor=center,
        align=center,
    }
}
\tikzset{
    standard node circle/.style n args={1}{%
        fill=our#1,
        draw=our#1border,
        circle,
        inner sep=0.1cm,
        minimum height=0,
        minimum width=0,
    }
}
\tikzset{
    standard node circle/.prefix style = standard node
}

\tikzset{
    standard line/.style n args={0}{%
        line width=0.04cm,
        rounded corners=0.1cm,
    }
}
\tikzset{
    standard arrow/.style n args={0}{%
        -latex,
    }
}
\tikzset{
    standard arrow/.prefix style = standard line
}

\tikzset{
    simple node image/.style n args={0}{%
        rectangle,
        inner sep=0,
        text centered,
        anchor=center,
        align=center,
        node distance=0mm
    }
}

\makeatletter
\newcommand{\printfnsymbol}[1]{%
  \textsuperscript{\@fnsymbol{#1}}%
}
\makeatother

\crefname{appsec}{Supp.\ Sec.}{Supp.\ Secs.}
\Crefname{appsec}{Supp.\ Section}{Supp.\ Sections}
\crefname{appfig}{Supp.\ Fig.}{Supp.\ Figs.}
\Crefname{appfig}{Supp.\ Figure}{Supp.\ Figures}
\crefname{apptab}{Supp.\ Tab.}{Supp.\ Tabs.}
\Crefname{apptab}{Supp.\ Table}{Supp.\ Tables}
\crefname{appeq}{Supp.\ Eq.}{Supp.\ Eqs.}
\Crefname{appeq}{Supp.\ Equation}{Supp.\ Equations}

\title{\methodname: Scalable Self-Supervised Novel View Synthesis from Real-World Video}

\author{
    {\large \textbf{Ulrich Prestel}\NoHyper\thanks{Equal Contribution.}\endNoHyper, \hspace{0.3em} \textbf{Stefan Andreas Baumann}\footnotemark[1], \hspace{0.3em} \textbf{Nick Stracke}, \hspace{0.3em} \textbf{Bj\"orn Ommer}}\\
    CompVis @ LMU Munich, Munich Center for Machine Learning (MCML)
}

\abstract{%
    Self-supervised novel view synthesis (NVS) remains challenging to scale, despite the abundance of video data, largely due to the brittleness of training on realistic videos and the hard-to-predict scaling behavior of multi-network system designs.
We introduce \methodname, a unified, feed-forward transformer that consolidates camera estimation, scene reconstruction, and rendering into a single backbone, turning self-supervised NVS into a well-posed single-model scaling problem.
A minimal dynamic state, treated as a nuisance factor, absorbs time-varying content and enables stable training on unconstrained real-world video.
Importantly, \methodname keeps \emph{static}-scene NVS as its target task: dynamic content is leveraged purely as scalable supervision, not reconstructed as in dynamic-scene (4D) NVS.
Across multiple model sizes and orders of magnitude in data, \methodname exhibits clean power-law scaling with data and compute, and outperforms static-scene data mixtures.
On a large number of benchmarks, \methodname achieves strong zero-shot open-set performance competitive with state-of-the-art supervised approaches.

    \textbf{Project Page:} \url{https://compvis.github.io/rayder}
    \\
    \textbf{Code:} \url{https://github.com/compvis/rayder}
    \vspace{-3.2em}
}

\begin{document}
\maketitle

\begin{figure}[H]
    \centering
    \ifeccv\vspace{-1em}\fi
    \includegraphics[width=\linewidth]{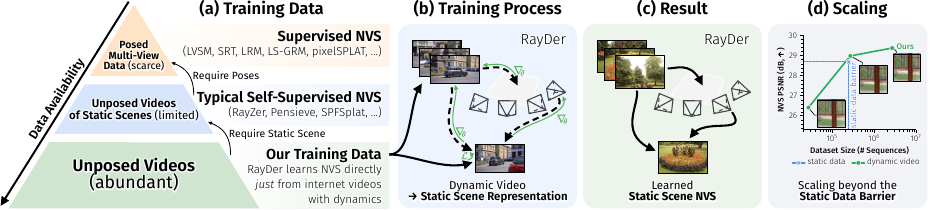}
    \caption{
        \textbf{Training Static-scene Novel View Synthesis from Abundant Video.}
        Existing approaches rely on scarce data sources: supervised NVS requires posed multi-view images, while prior self-supervised methods require unposed videos/image collections of static scenes.
        Our method instead trains from generic unposed videos that may contain dynamic objects, enabling learning from the dominant form of visual data.
        This removes the static-scene data bottleneck and unlocks improved scaling with dataset size.
    }
    \vspace{-1em}
    \label{fig:teaser}
\end{figure}
\setcounter{footnote}{0}
\RenewDocumentCommand{\paragraph}{s m}{\vspace{.25em}\noindent\textbf{#2\IfBooleanF{#1}{.}}}

\section{Introduction}
\label{sec:intro}
Novel view synthesis (NVS) should, in principle, be a highly scalable learning problem:
given a set of posed views of a (static) scene, learn to predict other views.
However, posed multi-view data is extremely scarce.
Self-supervised NVS removes this pose requirement by learning camera geometry jointly with view synthesis, with recent methods even rivaling pose-supervised ones under real-world conditions~\citep{jiang2025rayzer}.
Here, view synthesis is itself the \emph{target} task, not a pretext task for learning transferable features:
camera-pose labels are noisy and expensive to obtain on real video at scale, so removing them is precisely what makes abundant, unlabeled video usable for training the synthesis model itself.
Despite this, and large amounts of video being available on the internet~\citep[cf.][]{youtube2025press}, these methods rely on restrictive assumptions -- most notably, static scenes from curated datasets -- that prevent reliable training at scale.
We argue that the main obstacle to scalable self-supervised NVS is not data availability, but rather how current systems are designed to use that data.

Most existing approaches to self-supervised NVS~\cite{jiang2025rayzer,huang2025no,huang2025spfsplatv2,mitchel2025xfactor,wang2025less,wang2025recollection} are built as multi-network pipelines, with separate components for camera estimation, scene representation, and rendering.
While effective at small scales, such designs make scaling difficult in practice:
capacity has to be allocated across multiple interacting networks whose behavior is hard to predict and prohibitively expensive to sweep as models grow.
As a result, even when more data or compute is available, scaling remains brittle and inconsistent.

A related challenge is robustness to the videos that are available at scale:
unconstrained videos often contain dynamic scene content, which exposes instabilities in existing methods and prevents direct training on such scalable data sources.
Importantly, our goal is \emph{not} dynamic-scene (4D) NVS, but \emph{static}-scene NVS learned from dynamic video:
dynamic content is never reconstructed, only absorbed as a nuisance factor during training.
Stable learning under these conditions is the prerequisite for accessing the data regime where scaling can be meaningfully studied.

We introduce \textbf{\methodname}, a unified, feed-forward transformer that enables scalable self-supervised novel view synthesis.
Building on the RayZer~\citep{jiang2025rayzer} lineage, we develop a method that unifies camera estimation, scene reconstruction, and rendering in a single backbone, to enable scaling of self-supervised NVS.
This simplification is not merely architectural:
at fixed parameter counts, unification improves both pose estimation and novel view synthesis quality significantly, and enables straightforward, predictable scaling.

To ensure stable training on general video, \methodname uses a minimal explicit dynamic state variable that absorbs time-varying scene content during training.
This variable is treated as a nuisance factor rather than a semantic representation and is not used at inference time -- its role is solely to prevent scene dynamics information from corrupting camera pose representations, enabling stable training on unconstrained videos without changing the static-scene NVS task.

Across three orders of magnitude of training data and four model sizes, \methodname exhibits clean scaling in \emph{both} data and compute simultaneously.
Over the explored regimes, compute-optimal performance is well described by a \emph{single simple power-law fit}.
The insight here is \emph{not} the unsurprising observation that ``more data helps''.
Rather, it is that self-supervised NVS from unconstrained video \emph{becomes} a well-behaved scaling problem once two obstacles are removed:
the brittle optimization of multi-network pipelines, and the corruption of camera representations by dynamic content.
Once training is unified and dynamics are treated as a nuisance factor, self-supervised NVS scales like a standard single-model learning problem -- cleanly and predictably in data, model size, and compute.
We further show that aggregating existing static-scene datasets does not reproduce our scaled model's performance:
the gains arise from \emph{both} increased data availability and a system design that allows scaling to manifest.

Our main contributions are as follows:
\begin{itemize}[leftmargin=*,nosep]
    \item A unified single-network architecture for self-supervised NVS, replacing multi-network pipelines and enabling predictable scaling in model size and compute.
    \item A training formulation that remains stable on unconstrained video of dynamic scenes, treating scene dynamics as a nuisance factor to enable training on large-scale data.
    \item An empirical analysis of scaling behavior across data, model size, and compute, including compute-optimal scaling trends.
\end{itemize}
\section{Related Work}

\paragraph{Feed-Forward Novel View Synthesis}
Per-scene optimization via NeRFs~\cite{mildenhall2020nerf} or 3D Gaussian Splatting~\cite{kerbl20233d} produces high-quality views but requires dense pose captures and per-scene fitting at test time.
Feed-forward models amortize this cost by predicting radiance fields~\cite{hong2023lrm}, Gaussian primitives~\cite{charatan2024pixelsplat,chen2024mvsplat,chen2024mvsplat360,xu2025depthsplat,zhang2024gs}, or latent renderings~\cite{jin2025lvsm,nair2025scaling,sajjadi2022object,sajjadi2022object,safin2023repast,wu2025cat4d,watson2024controlling,watson2022novel,rombach2021geometry,elata2025novel,zhou2025stable,liu2025scaling,yu2024viewcraftertamingvideodiffusion} from posed images, but remain dependent on external pose pipelines~\cite{schoenberger2016sfm,wang2025vggt} at training and often at test time.
Some methods reduce this dependency by leveraging flow, correspondence, or depth cues~\cite{smith2023flowcam,chen2023dbarf,fu2024colmap,hong2024pf3plat}, but retain partial geometric supervision.
Generative approaches for camera-controlled synthesis~\cite{zhou2025stable,liu2025scaling,yu2024viewcraftertamingvideodiffusion,watson2022novel,watson2024controlling,wu2025cat4d,zhang2025world,elata2025novel} often adapt pretrained (video) diffusion models for camera-controllable synthesis~\cite{zhou2025stable,liu2025scaling,yu2024viewcraftertamingvideodiffusion} and can hallucinate content beyond observed regions, but still require posed data for NVS training and take cameras as input rather than learning them.
\methodname removes pose supervision entirely and learns camera representations jointly with view synthesis from raw video.

\paragraph{Self-supervised and Pose-Free NVS}
Foundational work toward removing pose dependence in NVS includes UpSRT~\cite{sajjadi2022scene}, which encodes unposed images into a latent scene representation, decoded using query rays that specify a relative target pose, and the Video Autoencoder~\cite{lai2021video}, which learns to disentangle 3D structure and camera pose from video without pose supervision.
RUST~\cite{sajjadi2023rust} removes train-time pose dependency by learning a latent pose code, but requires partial target views, limiting transferability~\cite{mitchel2025xfactor}; DyST~\cite{seitzer2024dyst} handles dynamics but needs multi-view, multi-dynamics data that is difficult to obtain at scale.
A second line learns explicit camera representations from monocular video.
RayZer~\cite{jiang2025rayzer} trains three separate ViTs end-ot-end on unposed \emph{static}-scene video with only a photometric loss; Pensieve~\cite{wang2025recollection} adds Gaussian splatting and depth losses; Less3Depend~\cite{wang2025less} extends this to sparse images; E-RayZer~\cite{zhao2025erayzer} repurposes the same formulation for self-supervised 3D pretraining.
These methods yield strong results, but are restricted to curated static-scene data whose combined scale~\cite{zhou2018stereo,ling2024dl3dv,liu24uco3d,reizenstein21co3d} is orders of magnitude smaller than available web video (\cref{sec:method_dynamics}), and employ multi-network pipelines that complicate scaling~\cite{ghorbani2022scalingnmt}.
XFactor~\cite{mitchel2025xfactor} further shows that many such systems learn pose shortcuts that fail to transfer across scenes.
Pose-free Gaussian splatting methods~\cite{ye2024no,huang2025no,huang2025spfsplatv2,kang2025selfsplat,li2025vicasplat} address a complementary setting -- sparse unposed images --, but several bootstrap from geometric backbones~\cite{wang2024dust3r,leroy2024grounding} pretrained with 3D supervision.
\methodname addresses all three recurring limitations: static-data ceilings, multi-network complexity, and reliance on pretrained geometric priors.

\paragraph{Large 3D Vision Models}
A growing line of ``large 3D vision models'' learns general 3D understanding by unifying multiple geometry tasks in a single transformer backbone, trading hand-engineered pipelines for scale and data breadth. Systems like VGGT~\citep{wang2025vggt} build upon the success of VGG-SfM~\citep{wang2023vggsfm} but remove most inductive biases, making everything into a single transformer backbone trained for multiple tasks simultaneously. They achieve (near-)state-of-the-art results across many tasks by just training on multiple supervised tasks across a large number of datasets. This has been further improved by works such as MapAnything~\citep{keetha2025mapanything}, which further extended the set of tasks, $\pi^3$~\citep{wang2025pi3}, which removed the dependency on a specific canonical view, and others~\citep{fang2026incvggt,shen2025fastvggt,deng2025vggtlong,feng2025quantized}. Approaches like CUT3R~\citep{wang2025continuous} pursue similar problems from different perspectives, enabling incremental updates and improved efficiency.
\methodname takes inspiration from this direction: just as replacing the individual components of VGG-SfM~\citep{wang2023vggsfm} with a single transformer in VGGT~\citep{wang2025vggt} led to improved scalability and thus improved performance, we aim to make self-supervised NVS scalable using, among other aspects, unification into a single transformer backbone.

\section{Scaling Self-Supervised Novel View Synthesis}\label{sec:method}
\begin{wrapfigure}{r}{\ifeccv.33\linewidth\else.327\linewidth\fi}
    \centering
    \ifeccv\vspace{-2.4em}\else\vspace{-1.1em}\fi
    \includegraphics[width=\linewidth]{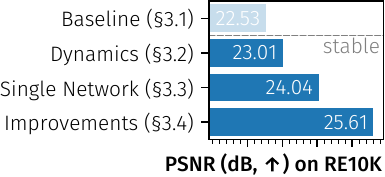}\ifeccv\vspace{-.5em}\else
    \caption{NVS performance across sections, training on general video (here, SA-B).}
    \vspace{-5em}
    \label{fig:improvements}
\end{wrapfigure}
Our goal is to make self-supervised novel view synthesis (NVS) scalable in data, model size, and compute, without introducing task-specific supervision or brittle system design. Starting from a modern feed-forward baseline (§\ref{sec:method_preliminaries}), we identify three bottlenecks that prevent scaling:
\begin{itemize}[leftmargin=2.5em,nosep]
    \item[§\ref{sec:method_dynamics}] Data: existing methods assume static scenes\\for training, severely limiting training data.
    \item[§\ref{sec:method_scalability}] System: multi-network pipelines complicate scaling and optimization.
    \item[§\ref{sec:method_improvements}] Quality: pose shortcuts and coarse patches limit reconstruction quality.
\end{itemize}
We address these with a sequence of targeted modifications, each validated by controlled ablations (see \cref{tab:ablation}).

\subsection{Preliminaries and Baseline}\label{sec:method_preliminaries}
\begin{wrapfigure}{r}{\ifeccv.45\linewidth\else.35\linewidth\fi}
    \centering
    \ifeccv\vspace{-2.6em}\else\vspace{-1.3em}\fi
    \includegraphics[width=\linewidth]{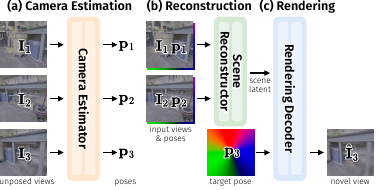}
    \caption{\textbf{Preliminaries: RayZer~\cite{jiang2025rayzer}.} RayZer uses three models responsible for different tasks: \textbf{a}) Camera Estimation, \textbf{b}) Reconstruction, \textbf{c}) Rendering.}
    \ifeccv\vspace{-3em}\else\vspace{-2em}\fi
    \label{fig:architecture_rayzer}
\end{wrapfigure}
We start our exploration with RayZer~\citep{jiang2025rayzer}, a feed-forward NVS method trained in a self-supervised manner on unposed, uncalibrated videos of \emph{static} scenes with camera motion.
Extending upon LVSM~\citep{jin2025lvsm}, RayZer consists of three distinct ViT~\citep{dosovitskiy2021an} subnetworks (\cref{fig:architecture_rayzer}):
a camera estimator $\mathcal{E}_\text{cam}: \{\image_i\} \mapsto \{\pose_i\}$ maps views $\{\image_i\}$ to poses $\{\pose_i\}\;\in SE(3)$ (and camera intrinsics).
Then, the scene reconstructor $\mathcal{E}_\text{scene}: \{(\image_i, \pose_i)\} \mapsto \mathbf{z}$ predicts a latent scene representation $\mathbf{z}$ from input views $\images_\text{input} = \{(\image_i, \pose_i)\}$.
Finally, a rendering decoder $\mathcal{D}_\text{render}: \mathbf{z}, \pose_\text{target} \mapsto \hat{\image}_\text{target}$ predicts target views.
All three networks are trained jointly end-to-end to optimize image-space reconstruction on target views $\images_\text{target}$ held out for the Scene Reconstructor, using poses jointly predicted for all views.
\ifeccv
    Poses are encoded as Plücker rays~\cite{plucker1865xvii}.
\else
    Poses are passed as Plücker maps~\cite{plucker1865xvii}, where pixels encode ray origin and direction.
\fi

\ifeccv\else\clearpage\fi
\paragraph{Baseline (\config{A})}
We use a scale-reduced RayZer-like model ($\sim$140M params) for our baseline (\config{A}) and train on two complementary datasets:
i) Segment Anything-Video~\citep[SA-V,][]{ravi2024sam2}, a diverse open-world video dataset with significant scene dynamics,
and ii) SpatialVid-HQ~\citep[SV-HQ,][]{wang2025spatialvid}, a curated, partially dynamic-scene dataset.
Evaluation is \emph{zero-shot} -- models are evaluated on unseen benchmarks. We measure NVS quality on RealEstate-10K~\citep[RE10K,][]{zhou2018stereo} in the standard pixelSplat~\cite{charatan2024pixelsplat} setting, and camera estimation via transferability~\cite{mitchel2025xfactor} on DL3DV-10k~\cite{ling2024dl3dv}.
\ifeccv\else
The main results are reported in \cref{tab:ablation}; significantly extended results with more NVS metrics and full camera estimation results covering both transferability~\cite{mitchel2025xfactor} and camera token probe results~\cite{jiang2025rayzer} are in \cref{tab:main_ablation}.
\fi

\begin{table}[t]
    \centering
    \caption{\textbf{Ablation Summary.} We progressively address instability (§\ref{sec:method_dynamics}), architectural scalability (§\ref{sec:method_scalability}), and synthesis quality (§\ref{sec:method_improvements}). NVS is zero-shot on RE10K~\cite{zhou2018stereo}, camera estimation via transferability~\cite{mitchel2025xfactor} on DL3DV-10k~\cite{ling2024dl3dv}. Full table in \cref{tab:main_ablation}.}
    \ifeccv\vspace{-0.5em}\fi
    \adjustbox{max width=\linewidth}{
    \newcommand{\wostate}{\raisebox{-.3em}{\smash{\textsc{\smalltableheaderfont\shortstack{w/o\\[-.2em]state}}}}}
    \newcommand{\wstate}{\raisebox{-.3em}{\smash{\textsc{\smalltableheaderfont\shortstack{w/\\[-.2em]state}}}}}
    \begin{tabular}{l@{\hskip .33em}lc c c>{\color{ourgray}}c c cc c c>{\color{ourgray}}c c cc}
        & & & & \multicolumn{5}{c}{\textit{Trained on SA-V~\cite{ravi2024sam2}}} & & \multicolumn{5}{c}{\textit{Trained on SV-HQ~\cite{wang2025spatialvid}}}\\
        \cmidrule{5-9} \cmidrule{11-15}
        \toprule
        \multicolumn{2}{l}{\multirow{2}{*}[-3pt]{Configuration}} & \multirow{2}{*}[-3pt]{\shortstack{Stable\\Training}} & & \multicolumn{2}{c}{NVS PSNR$\uparrow$} & & \multicolumn{2}{c}{Camera Est.} & & \multicolumn{2}{c}{NVS PSNR$\uparrow$} & & \multicolumn{2}{c}{Camera Est.} \\
        \cmidrule{5-6} \cmidrule{8-9} \cmidrule{11-12} \cmidrule{14-15}
        & & & & \wostate & \wstate & & \textsc{\smalltableheaderfont R@10$^\circ\!\!\uparrow$} & \textsc{\smalltableheaderfont t@{0.1$\uparrow$}} & & \wostate & \wstate & & \textsc{\smalltableheaderfont R@10$^\circ\!\!\uparrow$} & \textsc{\smalltableheaderfont t@{0.1$\uparrow$}} \\
        \midrule
        \multicolumn{5}{l}{\textit{§\ref{sec:method_dynamics}: Stable Training on Dynamic Video}} \\
        \textbf{A} & RayZer-like~\citep{jiang2025rayzer} Baseline & $\boldsymbol{\sim}$ & & {\color{ourgray}22.53\rlap{$^*$}} & -- & & {\color{ourgray}59.8\rlap{$^*$}} & {\color{ourgray}6.5\rlap{$^*$}} & & {\color{ourgray}22.69\rlap{$^*$}} & -- & & {\color{ourgray}66.0\rlap{$^*$}} & {\color{ourgray}7.7\rlap{$^*$}} \\
        \textbf{B} & + Dynamic State Prediction                & \cmark & & 13.42\rlap{$^\dagger$}  & 24.01                   & & 56.1 & 7.0 & & 13.48\rlap{$^\dagger$}   & 24.67                   & & 54.4 & 6.0 \\
        \textbf{C} & + State Dropout                           & \cmark & & 23.01                   & 23.76                   & & 62.4 & 8.1 & & 23.02                    & 24.10                   & & 69.2 & 8.2 \\
        \midrule
        \multicolumn{5}{l}{\textit{§\ref{sec:method_scalability}: Scalability through Consolidation}} \\
        \textbf{D} & + Single-network Consolidation            & \cmark & & 24.93                   & 25.33                   & & 68.8 & 16.3 & & 26.98                   & 27.49                   & & 74.1 & 19.7 \\
        \textbf{E} & + Parallel-target Attention               & \cmark & & 24.04                   & 25.12                   & & 70.1 & 15.6 & & 25.91                   & 26.21                   & & 70.9 & 18.2 \\
        \midrule
        \multicolumn{5}{l}{\textit{§\ref{sec:method_improvements}: Improving Synthesis Quality}} \\
        \textbf{F} & + Autoregression over Views (ordered)     & \cmark & & 23.08\rlap{$^\ddagger$} & 24.49\rlap{$^\ddagger$} & & 73.6 & 24.9 & & 23.53\rlap{$^\ddagger$} & 25.78\rlap{$^\ddagger$} & & 76.5 & 25.2 \\
        \textbf{G} & + Random-order Autoregression             & \cmark & & 25.45                   & 26.28                   & & 84.4 & 37.2 & & 27.27                   & 29.57                   & & 86.0 & 39.1 \\
        \textbf{H} & + Local High-resolution Layers            & \cmark & & 25.61                   & 26.87                   & & 85.0 & 40.2 & & 27.78                   & 30.23                   & & 88.7 & 42.4 \\
        \bottomrule
        \multicolumn{15}{c}{\ifeccv\scriptsize\else\footnotesize\fi \shortstack{$^*$Results for \textbf{A} are from selected runs that did not diverge. $^\dagger$Dynamic state modeling without dropout creates inference-time\ifeccv\\[-.33em]\fi dependency on state. \ifeccv\else\\[-.33em]\fi$^\ddagger$Ordered AR does not generalize to standard NVS test settings.}} \\
    \end{tabular}
    }
    \label{tab:ablation}
\end{table}

\subsection{Robust Learning from Dynamic Videos}\label{sec:method_dynamics}
\begin{wrapfigure}{r}{.3\linewidth}
    \centering
    \ifeccv\vspace{-4.15em}\else\vspace{-1.2em}\fi
    \includegraphics[width=\linewidth]{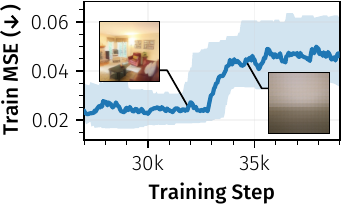}\ifeccv\vspace{-.5em}\fi
    \caption{Training RayZer directly on dynamic videos leads to instabilities and stalled training.}
    \ifeccv\vspace{-2.5em}\else\vspace{-1em}\fi
    \label{fig:instabilities}
\end{wrapfigure}
Scaling self-supervised NVS faces an immediate data bottleneck: truly static-scene videos, as required by current methods~\citep{jiang2025rayzer,wang2025less,wang2025recollection,mitchel2025xfactor}, are a tiny subset of what is available at scale.
However, training RayZer directly on dynamic video leads to gradient spikes and instabilities: the original RayZer~\cite{jiang2025rayzer} diverges consistently when trained on SpatialVid~\cite{wang2025spatialvid} or SA-V~\cite{ravi2024sam2} (cf. \cref{fig:instabilities}; see \cref{sec:rayzer_instabilities} for a detailed analysis).
We frame this as a \emph{representation} problem rather than an optimization problem.
In dynamic video, a target view $j$ is explained from an input view $i$ by two factors: the camera pose change $\pose_i \rightarrow \pose_j$ \emph{and} the dynamic state change $\state_i \rightarrow \state_j$.
Exposing only camera pose as conditioning forces the model to ``hide'' dynamic-state information inside the camera representation, causing representation drift and instabilities.
Importantly, even when training on dynamic video, our target task remains \emph{static}-scene NVS:
the dynamic state is what lets us \emph{learn} this static-scene task from dynamic data without modeling scene motion at inference, making training scalable rather than attempting dynamic-scene reconstruction.

\paragraph{Dynamic State Prediction and Dropout (\config{B, C})}
We address this by predicting a per-view dynamic state embedding $\state_i$ alongside the camera pose:
\begin{equation}
    \mathcal{E}_\text{cam,state}: \{\image_i\} \mapsto \{(\pose_i, \state_i)\},
    \ifeccv\quad\else\qquad\fi
    \mathcal{D}_\text{render,state}: \mathbf{z}, (\pose_\text{target}, \state_\text{target}) \mapsto \hat{\image}_\text{target},
\end{equation}
where $\state_i \in \mathbb{R}^{d_\text{state}}$ lets the model capture time-varying content without needing to interfere with the pose $\pose_i$, and is provided to the renderer as an additional token.
This intentionally minimal change -- no motion fields, temporal losses, or disentanglement -- eliminates training instabilities entirely:
across all our experiments, not a single run that includes this change (\config{B}) has diverged.
Since the target state $\state_\text{target}$ is unknown at inference, we randomly replace it with a zero vector during training~\cite{hinton2012improving}, forcing the model to synthesize plausible views both with and without state conditioning (\config{C}).
This retains the stability gains, while resolving the inference dependency on ground truth state and additionally improving camera estimation (\cref{tab:ablation}, \textbf{A}$\rightarrow$\textbf{C}), consistent with the dynamic state reducing pressure on the pose representation to encode dynamic information.
We stress that the state is deliberately \emph{not} a disentangled 4D representation:
it is a nuisance variable whose only role is to keep time-varying content out of the camera tokens.
We probe what it captures in \cref{sec:dynamic_state_exploration}, where transplanting states across frames shows that it primarily absorbs moving/time-varying scene content while the camera tokens carry pose; at inference on scenes with dynamic content, this manifests as the static scene being rendered from the correct novel pose while moving regions degrade to a temporal average (\cref{sec:dynamic_state_exploration} and limitations in \cref{sec:experiments}).

\subsection{Scalability through Network Consolidation}\label{sec:method_scalability}
With stable training on general video, the next bottleneck is architectural:
scaling multi-network systems -- which current self-supervised NVS methods~\citep{jiang2025rayzer,huang2025no,huang2025spfsplatv2,mitchel2025xfactor,wang2025less,wang2025recollection} rely on -- is highly complex~\citep{ghorbani2022scalingnmt}, as capacity must be distributed across interacting components whose scaling behavior is hard to predict.

\begin{wrapfigure}{r}{.33\linewidth}
    \centering
    \ifeccv\vspace{-2.5em}\else\vspace{-1.1em}\fi
    \includegraphics[width=\linewidth]{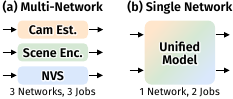}
    \caption{\textbf{Consolidation.} We combine RayZer's three networks (\textbf{a}) into one (\textbf{b}).}
    \ifeccv\vspace{-2em}\else\vspace{-1.5em}\fi
    \label{fig:consolidation_illustration}
\end{wrapfigure}
\paragraph{Single-Network Consolidation (\config{D})}
To reduce scaling decisions to a single network, which can allocate capacity between tasks as needed, and improve performance by sharing features, we unify all three components -- camera/dynamic state estimation, scene reconstruction, and rendering (see \cref{fig:consolidation_illustration}) -- in a single model $\model$.
Besides scaling simplicity, this is motivated by the idea that pose estimation and view synthesis are not separate problems~\citep{wang2025vggt}: they can share features, and training signals can become cleaner when the clear separation between networks is removed.
Our unified model $\mathcal{M}$ operates in two modes (where $\cancelasX{\cdot}$ denotes abscence of an input/output):
\begin{equation}
    \begin{matrix*}[l]
        \!\!\!\model \!:\!\! \left\{
        \vphantom{\rule{0pt}{16pt}}
        \right.\\
        \vphantom{\rule{0pt}{12pt}}
    \end{matrix*}
    \begin{matrix*}[l]
        \{(\image_i, \cancelasX{\pose}_{\phantom{i}}, \cancelasX{\state})\} \mapsto \{(\pose_i, \state_i)\} &&\quad{\color{ourgrayborder}\triangleright\ \text{\ifeccv{Cam.\ Est.}\else{Camera Estimation}\fi}}\ifeccv\!\!\!\!\fi\\
        \underbrace{\{(\image_i, \pose_i, \cancelasX{\state})\}}_\text{input views} \cup \underbrace{(\cancel{\image}, \pose_j, {\color{ourgray}\state_j})}_\text{target pose} \mapsto \underbrace{\hat{\image}_j}_\text{\!\!\!\!\!\!\!\!\!\!\!\!target view\!\!\!\!\!\!\!\!\!\!\!\!} &&\quad{\color{ourgrayborder}\triangleright\ \text{\ifeccv{NVS}\else{Novel View Synthesis}\fi}}\ifeccv\!\!\!\!\fi
    \end{matrix*}
\end{equation}
All heavy computation lies in a single shared backbone, conditioned on token role via adaptive norms~\citep{huang2017arbitrary,nair2025scaling}.
In addition to significantly simplifying scaling decisions, empirically, this unification at fixed parameter count leads to significant gains in both NVS and camera estimation performance (\cref{tab:ablation}, \textbf{C}$\rightarrow$\textbf{D}).

\begin{wrapfigure}{r}{\ifeccv.165\linewidth\else.13\linewidth\fi}
    \ifeccv\else\vspace{-2.6em}\fi
    \ifeccv
        \hspace{-.07\linewidth}\includegraphics[width=1.07\linewidth]{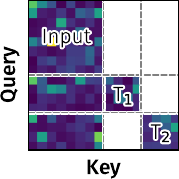}\vspace{-.5em}
    \else
        \hspace{-.0\linewidth}\includegraphics[width=1.0\linewidth]{img/parallel_target_attention.pdf}\vspace{-.5em}
    \fi
    \caption{Our at\-tention mask.}
    \vspace{-3em}
    \label{fig:parallel_attention_mask}
\end{wrapfigure}
\paragraph{Parallel-target Attention (\config{E})}
Naively treating the consolidated model as decoder-only~\cite{jin2025lvsm} reprocesses input views for each target view, which is prohibitively expensive.
We factorize attention such that input tokens only attend to each other, while target tokens attend to themselves and input tokens (see \cref{fig:parallel_attention_mask}).
This enables KV caching during inference and parallel target prediction during training, reducing per-target compute by $\sim\!7\times$ at a minor quality trade-off (\cref{tab:ablation}, \textbf{D}$\rightarrow$\textbf{E}).

\subsection{Improving Synthesis Quality}\label{sec:method_improvements}
With stable training and a single scalable backbone, the remaining issues are quality-related: pose representations can learn shortcuts in video, and large patch sizes sacrifice local details.

\begin{wrapfigure}{r}{\ifeccv.34\linewidth\else.36\linewidth\fi}
    \centering
    \ifeccv\vspace{-2.4em}\else\vspace{-1.2em}\fi
    \includegraphics[width=\linewidth]{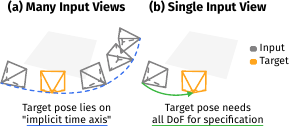}\ifeccv\vspace{-.3em}\fi
    \caption{Many input views (\textbf{a}) allow encoding camera poses via an implicit ``time'' axis; sparse views (\textbf{b}) require true relative camera poses.}
    \ifeccv\vspace{-2em}\else\vspace{-1em}\fi
    \label{fig:ar_nvs_illustration}
\end{wrapfigure}
\paragraph{Autoregressive Pose Learning (\config{F, G})}
When training on video frames, many input views make pose prediction easy to solve by using frame-order shortcuts rather than actual geometry (\cref{fig:ar_nvs_illustration}a).
We find that in practice, this results in predicted poses primarily encoding time rather than the true viewpoint.
In contrast, single- or few-view NVS requires the full pose to be encoded geometrically (\cref{fig:ar_nvs_illustration}b).
We implement that by training autoregressively over views:
given a subset of views, predict another, then condition on the expanded set.
This forces the model to learn to predict poses that are useful for NVS in both sparse and dense settings.
Extending upon our factorized attention pattern (\cref{sec:method_scalability}), we make attention causal over input views and train next-view NVS for $|\mathcal{I}_\text{input}| = 1, 2, ..., |\mathcal{I}_\text{total}| - 1$ input views.
Ordered autoregression (\config{F}) consequently improves camera estimation quality significantly (\cref{tab:ablation}, \textbf{E}$\rightarrow$\textbf{F}), but creates a train-test gap, since standard NVS settings do not condition on and generate frames in temporal order.
Randomizing the autoregression order instead (\config{G}) closes this gap and further improves both camera estimation and NVS quality (\cref{tab:ablation}, \textbf{E}$\rightarrow$\textbf{G}).

\paragraph{Local High-resolution Layers (\config{H})}
Large patch sizes hurt synthesis quality~\cite{wang2025scaling,jin2025lvsm,peebles2023scalable}, but reducing the patch size $p$ scales cost as $\mathcal{O}(p^4)$, making it prohibitively expensive.
Following \citet{crowson2024hourglass}, we add shallow high-resolution local layers (using neighborhood attention~\cite{hassani2023neighborhood}) around the main backbone.
These layers operate intra-frame and provide extra high-frequency capacity at minimal cost (\cref{tab:ablation}, \textbf{G}$\rightarrow$\textbf{H}).

\begin{figure}[t]
    \centering
    \includegraphics[width=\ifeccv.98\linewidth\else.9\linewidth\fi]{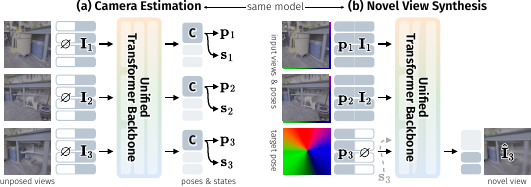}\vspace{-.2em}
    \caption{\textbf{Final Architecture Overview.} \methodname unifies camera estimation (\textbf{a}) and novel view synthesis (\textbf{b}) in a single transformer backbone. Lightweight local intra-frame encoder and decoder layers handle high-resolution processing.}
    \label{fig:architecture_high_level}
\end{figure}

\subsection{Final Architecture}
\begin{wraptable}{r}{.45\linewidth}
    \centering
    \ifeccv\vspace{-3.5em}\else\vspace{-1.4em}\fi
    \caption{\textbf{Model Scales.} We jointly scale depth, width, and head count.}
    \ifeccv\else\vspace{-.2em}\fi
    \adjustbox{max width=\linewidth}{
    \begin{tabular}{lccccc}
        \toprule
        Model & Layers $N$ & Hidden Size $d$ & Heads & Parameters \\
        \midrule
        \methodname-XS & 12 & 384 & 6 & 59M \\
        \methodname-S & 18 & 512 & 8 & 145M \\
        \methodname-B & 24 & 768 & 12 & 422M \\
        \methodname-L & 24 & 1024 & 16 & 743M \\
        \bottomrule
    \end{tabular}
    }
    \ifeccv\vspace{-2.5em}\else\vspace{-1.5em}\fi
    \label{tab:model_scales}
\end{wraptable}
\config{H} is the final \methodname architecture, which we train at four different model scales (\cref{tab:model_scales}):
a single transformer that i) predicts per-frame pose and state tokens $\{(\pose_i,\state_i)\}$ from a set of images $\{\mathbf{I}_i\}$,
ii) synthesizes target views via random-order autoregression with parallel-target attention,
and iii) trains stably on general video.
\methodname-S has the same scale (depth, width) as our ablation models (\config{A-H}), but is trained longer on more data; \methodname-B is scale-matched vs.\ RayZer~\cite{jiang2025rayzer}.
An architecture overview is shown in \cref{fig:architecture_high_level}.

\section{Experiments}\label{sec:experiments}
Our experiments address four major questions:
\begin{itemize}[leftmargin=4em,nosep]
    \item[§\ref{sec:exp_scaling}\phantom{,\ref{sec:exp_prior_art}}] Does \methodname exhibit predictable scaling behavior in data, model size, and compute?
    \item[§\ref{sec:exp_static_vs_dynamic_data}\phantom{,\ref{sec:exp_prior_art}}] Do existing static-scene datasets support the same scaling regime, or is learning from general video essential?
    \item[§\ref{sec:exp_geometry}\phantom{,\ref{sec:exp_prior_art}}] Does the learned camera geometry encode genuine 3D structure, and does it scale alongside synthesis quality?
    \item[§\ref{sec:exp_open_set},\ref{sec:exp_prior_art}] How does open-set self-supervised NVS compare to prior work with supervision \& large-scale pretraining?
\end{itemize}

\begin{figure}[h]
    \centering
    \includegraphics[width=\linewidth]{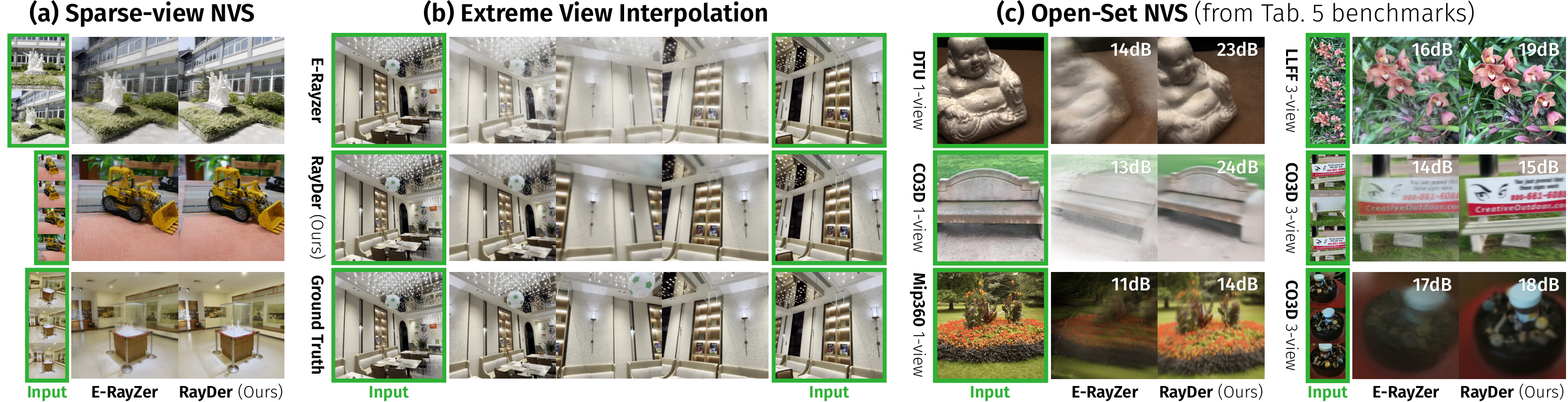}%
    \caption{\textit{Zero-shot} qualitative samples of \methodname compared with E-RayZer~\cite{zhao2025erayzer} in \textbf{(a)} typical (non-dense view) NVS settings, \textbf{(b)} an extreme setting with $\sim$zero context view overlap, and \textbf{(c)} settings evaluated in \cref{tab:open_set_nvs}. Our \methodname model, trained on large-scale non-static-constrained video data, outperforms E-RayZer -- a prior model trained on a multi static dataset mixture -- by a wide margin.}
    \label{fig:qualitative_main}
\end{figure}

\paragraph{Implementation Details}
We train all models on SpatialVid~\cite{wang2025spatialvid} ($\sim$2.7M videos), extracting 8 views per clip with $\sim$0.5s spacing (randomly chosen per epoch), using AdamW~\cite{loshchilov2018decoupled} at batch size 256 and a resolution of $256^2$.
We measure PSNR, LPIPS~\cite{zhang2018unreasonable}, and SSIM~\cite{wang2004image} on synthesized novel views for our evaluations, generally in \emph{zero-shot} settings on unseen datasets and using $256^2$ resolution unless noted otherwise.
For further details, see \cref{sec:app_exploration_details,sec:app_imp_details}.

\paragraph{Train-Test Leakage}
To ensure that our results, especially gains with increased train data scale, stem from improved generalization rather than leakage, we also check for leakage between our train and test sets, in addition to performing all our main evaluations in zero-shot settings.
Specifically, we check each test view from every dataset we evaluate on against the videos used for training our main models \& models used for scaling evaluations (i.e., our copy of SpatialVid~\citep{wang2025spatialvid}), in two stages:
first, we compute pHash and dHash perceptual hashes~\citep{klingerphash,buchnerimagehash} for all frames and flagged any train-test pair with Hamming distance $<8$\,bits as a candidate -- yielding 16{,}381 candidate pairs.
Second, to discard hash collisions on visually unrelated content, we keep only pairs with DINOv3-L~\cite{simeoni2025dinov3} CLS cosine similarity $\geq 0.2$, narrowing the set to 191 candidates.
Manual inspection of all 191 remaining pairs (candidate train view vs.\ full test scene) revealed zero matches -- therefore, our evaluations should be free of the influence of train-test leakage and instead represent genuine generalization.

\subsection{Scaling Behavior across Data, Model Size, and Compute}\label{sec:exp_scaling}
Prior self-supervised NVS methods are fundamentally \emph{data-limited}:
trained on small, curated static-scene datasets that saturate quickly, they cannot meaningfully scale model capacity.
By enabling stable training directly on dynamic video, \methodname allows scaling across multiple orders of magnitude.

\paragraph{Setup}
We train \methodname at four scales (\textbf{XS}/\textbf{S}/\textbf{B}/\textbf{L}; \cref{tab:model_scales}) on three dataset fractions of SpatialVid:
\textbf{1\%} ({$\sim$27k}~videos), \textbf{10\%} ({$\sim$270k}, matching the combined size of common static-scene NVS datasets~\cite{liu24uco3d,reizenstein21co3d,ling2024dl3dv,zhou2018stereo}), and \textbf{100\%} ({$\sim$2.7M}).

\begin{figure}[t]
    \centering
    \includegraphics[width=\linewidth]{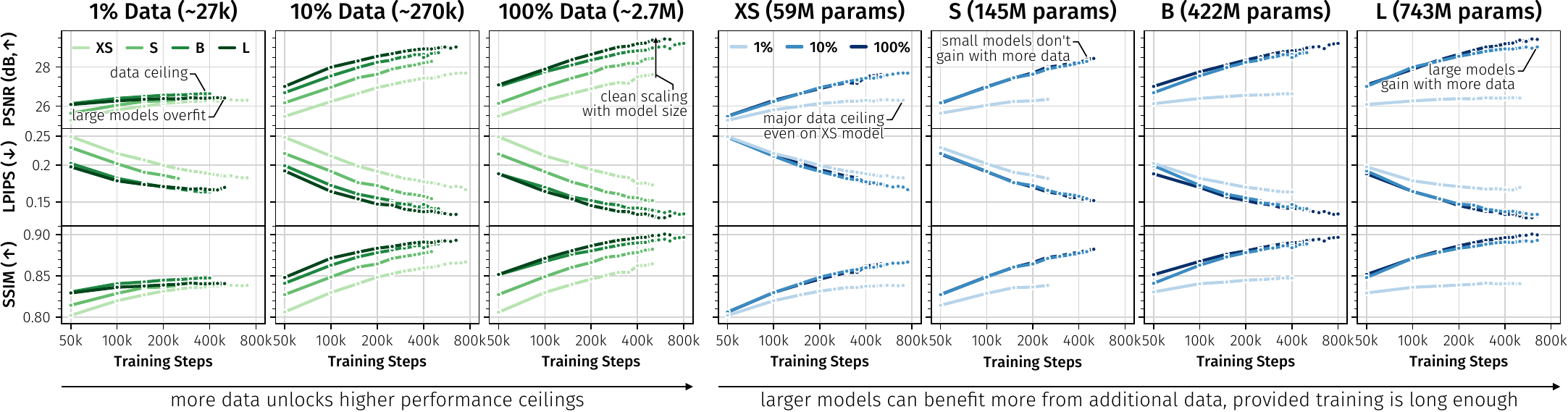}
    \caption{\textbf{Scaling Across Data and Model Size.} We evaluate models trained on SpatialVid ({2.7M} total samples) at different model scales (visualized as shades of {\color{matplotlibgreen}green}) and dataset fractions (shades of {\color{matplotlibblue}blue}), on RE-10k~\cite{zhou2018stereo}. \textit{Left:} Increasing data scale consistently improves performance, as long as model scale is not a limit. At small data scales, large models tend to overfit, resulting in \emph{worse} performance than smaller ones. \textit{Right:} Increasing model scale also consistently improves performance. However, insufficient dataset scale imposes an upper ceiling on achievable test performance.}
    \vspace{-0.5em}
    \label{fig:scaling}
\end{figure}

\Cref{fig:scaling} shows that both data and model scaling consistently lead to improvements, provided neither is a bottleneck.
Increasing data consistently improves performance when model capacity and training are sufficient, while small models saturate early.
Large models overfit on small data, even underperforming against smaller models.
These results highlight \textbf{scaling neither compute nor data alone is sufficient} -- scaling requires \emph{both}.
All models at 1\% data scale converge to a common ceiling, confirming data as the dominant bottleneck at small scale.
Benefits continue beyond 10\%, which matches the combined size of common static-scene NVS datasets, highlighting the need for more scalable data sources.

\begin{figure}[t]
    \centering
    \adjustbox{max width=\linewidth}{
        {
            \includegraphics[height=.4\linewidth,]{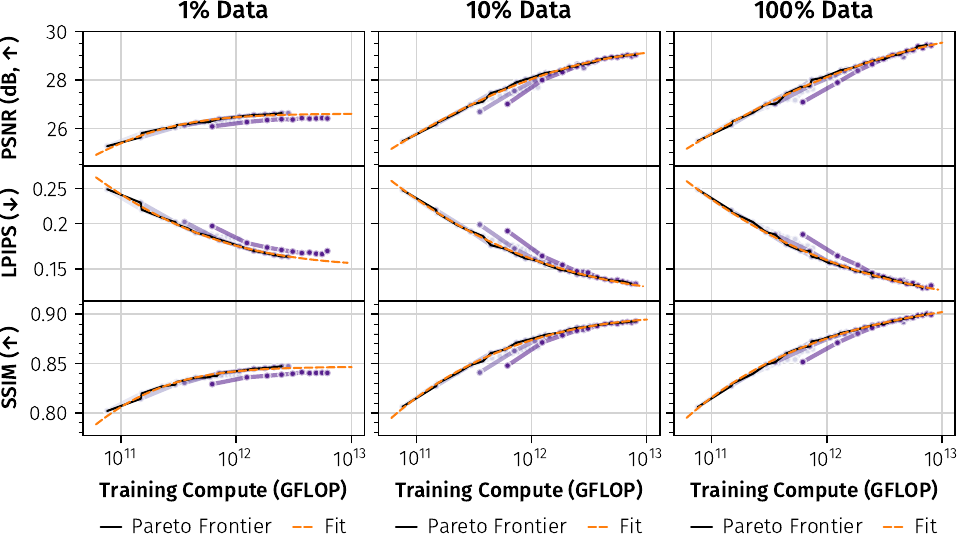}
            \raisebox{.03\linewidth}{\includegraphics[height=.35\linewidth]{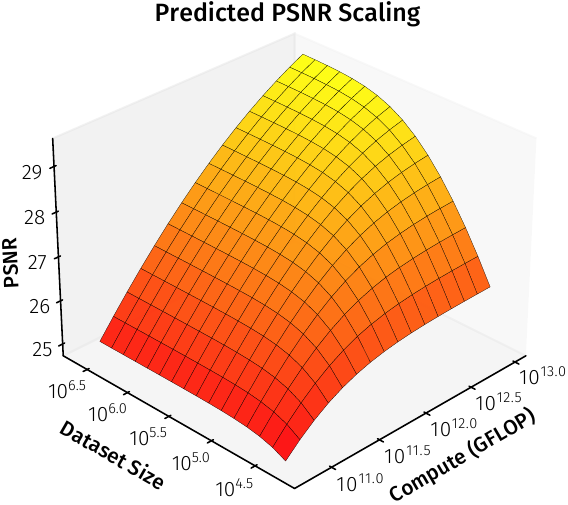}}
        }
    }
    \caption{\textbf{Compute-Optimal Scaling Analysis.} RayDer's compute-optimal performance (i.e., the compute-quality Pareto frontier) on unseen datasets (here, RE10K~\cite{zhou2018stereo}) \emph{across both compute and train dataset size} is well-approximated by a \emph{single} power law.
    \ifeccv\vspace{-1em}\fi
    }
    \label{fig:compute_optimal_scaling}
\end{figure}

\paragraph{Compute-optimal Scaling}\label{sec:exp_compute_optimal_scaling}
Building on LLM scaling analysis~\cite{henighan2020scaling,hoffmann2022an,kaplan2020scaling}, we find that \methodname's compute-optimal Pareto frontier of NVS performance on unseen test sets across training compute $C$ (in GFLOP) and dataset size $D$ (in number of videos) can be modeled as~\citep[cf.][Eq.\ 1.5]{kaplan2020scaling}:
\begin{equation}
    \underbrace{\vphantom{_{a_b}}L(C, D)}_\text{test metric\vphantom{[}} \quad= \underbrace{\vphantom{_{a_b}}L_\infty}_\text{irreducible part~\cite{henighan2020scaling}} + \quad\bigl(\ifeccv\!\!\!\!\!\else\!\!\fi\underbrace{\vphantom{_{a_b}}AC^{-\alpha}}_\text{compute term\vphantom{[}} + \quad\!\!\!\underbrace{\vphantom{_{a_b}}BD^{-\beta}}_\text{data term\vphantom{[}}\ifeccv\else\ \fi\bigr)^\gamma,
    \label{eq:scaling_law}
\end{equation}
with $L \in \{\mathrm{MSE}, \mathrm{LPIPS}, 1 - \mathrm{SSIM}\}$ (MSE corresponds to PSNR and is computed on image data scaled to $[-1, 1]$).
The compute and data terms each capture the error that can only be removed by scaling that aspect.
\ifeccv
    Fitting this power law to our compute-optimal Pareto frontier of models we trained across model and dataset scale (see \cref{fig:compute_optimal_scaling}), we obtain an accurate ($R^2 > 0.99$) description of \methodname's behavior over both compute and data scale, confirming that its scaling is well-behaved.
    All metrics exhibit non-zero irreducible error terms $L_\infty > 0$, reflecting fundamental limits of the setting (e.g., occluded regions).
    Both compute and data terms contribute meaningfully, with the latter formalizing an important empirical insight: increasing compute yields diminishing returns unless sufficient training data is available, and scaling training data beyond curated static-scene datasets requires methods that can train directly on general video.
\else
    Fitting this power law to our compute-optimal Pareto frontier of models we trained across model and dataset scale (see \cref{fig:compute_optimal_scaling}), we obtain an accurate ($R^2 > 0.99$) description of \methodname's behavior over both compute and data scale, confirming that its scaling is well-behaved:
    \begin{align}
        \mathrm{MSE}(C, D) &\approx 0.0033 + \left(200\cdot C^{-0.40} + 2.6\cdot D^{-0.60} \right)^{2.82} &&{\color{ourgrayborder}\triangleright\ R^2 = 0.997}\\
        \mathrm{LPIPS}(C, D) &\approx 0.11 + \left(7000\cdot C^{-0.43} + 14\cdot D^{-0.58} \right)^{1.82} &&{\color{ourgrayborder}\triangleright\ R^2 = 0.997}\\
        1 - \mathrm{SSIM}(C, D) &\approx 0.076 + \left(700\cdot C^{-0.35} + 10\cdot D^{-0.47} \right)^{3.34} &&{\color{ourgrayborder}\triangleright\ R^2 = 0.997}
    \end{align}
    All metrics exhibit non-zero irreducible error terms $L_\infty > 0$, reflecting fundamental limits of the setting (e.g., occluded regions).
    Both compute and data terms contribute meaningfully, with the latter formalizing an important empirical insight: increasing compute yields diminishing returns unless sufficient training data is available, and scaling training data beyond curated static-scene datasets requires methods that can train directly on general video.
\fi
Refitting the same power law on additional, harder zero-shot benchmarks (WildRGBD~\cite{xia2024rgbd}, CO3D~\cite{reizenstein21co3d}) remains similarly accurate (\cref{sec:app_scaling_fit_robustness}), indicating the trend is not an artifact of fitting a single test set.

\begin{figure}[t]
    \centering
    \includegraphics[width=\ifeccv\linewidth\else.8\linewidth\fi]{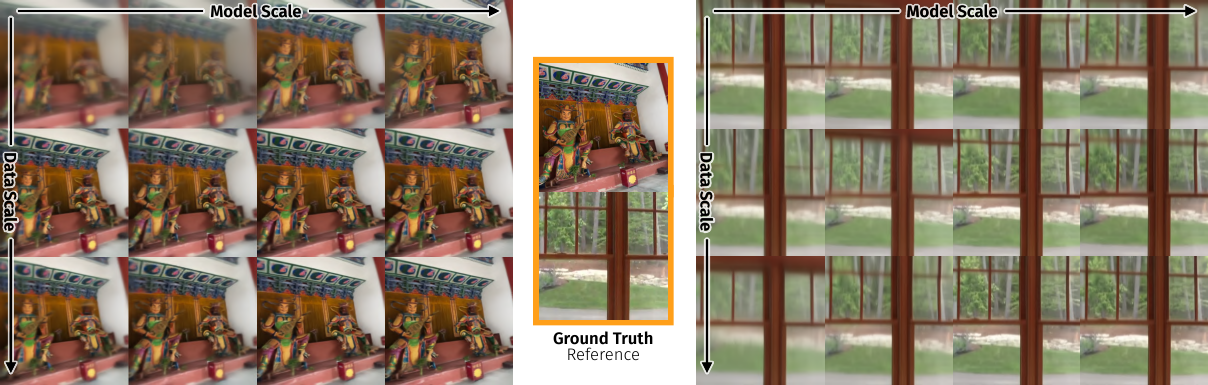}
    \caption{\textbf{Qualitative Scaling.} \methodname's qualitative behavior follows the trends seen in quantitative evals (\cref{fig:scaling}): more data \& compute jointly improve NVS quality.}
    \label{fig:scaling_qualitative}
\end{figure}

\subsection{Static-Scene Data Does Not Enable the Same Scaling Regime}\label{sec:exp_static_vs_dynamic_data}
The scaling analysis in \cref{sec:exp_scaling} establishes that data scale is a dominant factor for continued improvement.
But can existing static-scene datasets, which offer domain-aligned training signals for standard NVS benchmarks, supply sufficient data to sustain this scaling regime?
If so, the added complexity of training on dynamic video would be unnecessary.
To test this, we train two additional \methodname-L models:
a \textbf{static-only} model trained on a mixture of multiple large-scale static-scene NVS datasets (RE10K~\cite{zhou2018stereo}, DL3DV-10K~\cite{ling2024dl3dv}, uCO3D~\cite{liu24uco3d}; $\sim$247k videos total;
denoted as \emph{static mix}), and
a \textbf{mixed} model trained on both SpatialVid~\cite{wang2025spatialvid} (mixed with dynamic content, $\sim$2.7M videos) \emph{plus} the static mix, with equal sampling between both during training.

\Cref{tab:static_vs_dynamic_data} shows that the static-only model significantly underperforms despite using static-scene data closely aligned with the (static-scene) test setting.
The static mix reflects the combined scale of commonly used static-scene NVS datasets and roughly matches the 10\% data fraction in \cref{sec:exp_scaling}, where our scaling analysis already predicts that larger models cannot fully benefit due to limited data.
Despite the domain alignment advantage, the scale deficit dominates: training on more (partially dynamic) video empirically outweighs the benefit of a cleaner training distribution.
Adding static data to the larger video dataset at the same training horizon yields only marginal gains, suggesting the static datasets are largely subsumed by the larger corpus.
These results validate our core thesis: the bottleneck for self-supervised NVS scaling is not the quality of static-scene curation but the \emph{quantity} of data, which requires moving beyond curated static-scene corpora.

\begin{table}[t]
    \centering
    \caption{\textbf{Training Data Comparison.} We train models at scale on three different dataset combinations: \textit{Static Mix:} a combination of multiple public static-scene NVS datasets (RE10K~\cite{zhou2018stereo}, DL3DV-10K~\cite{ling2024dl3dv}, uCO3D~\cite{liu24uco3d}; $\sim$247k videos total), \textit{General:} SpatialVid~\cite{wang2025spatialvid} ($\sim$2.7M videos total), and a combination of the two. During evaluation on unseen datasets, combining even multiple public static-scene datasets underperforms substantially compared to training on general video. Combining both leads to minor additional gains.}
    \adjustbox{max width=\linewidth}{\scalebox{0.8}{
    \begin{tabular}{lcclc c c@{\hskip .5em}c@{\hskip .5em}c}
        \toprule
        \multirow{2}{*}[-2pt]{Model} & \multirow{2}{*}[-2pt]{Steps} & & \multicolumn{1}{c}{\multirow{2}{*}[-2pt]{Training Data}} & \multirow{2}{*}[-2pt]{Scene Dynamics} & & \multicolumn{3}{c}{RE10K NVS} \\
        \cmidrule{7-9}
        & & & & & & \textsc{\smalltableheaderfont PSNR$\uparrow$} & \textsc{\smalltableheaderfont LPIPS$\downarrow$} & \textsc{\smalltableheaderfont SSIM$\uparrow$} \\
        \midrule
        \multirow{3}{*}{\methodname-L} & \multirow{3}{*}{500k} & & Static Mix only ($\sim$250k) & static & & 28.68 & 0.158 & 0.888 \\
        & & & SpatialVid only ($\sim$2.7M) & including dynamic & & \underline{29.38} & \samebf{0.135} & \underline{0.899} \\
        & & & Static Mix + SpatialVid$^\dagger$ & including dynamic & & \samebf{29.42} & \underline{0.136} & \samebf{0.901} \\
        \bottomrule
        \multicolumn{9}{c}{\ifeccv\scriptsize\else\footnotesize\fi $^\dagger$batch composition during training is equally distributed between static mix and SpatialVid.}
    \end{tabular}
    }}
    \label{tab:static_vs_dynamic_data}
\end{table}

\subsection{Learned Camera Geometry: Transferability and Scaling}\label{sec:exp_geometry}
A key concern for self-supervised NVS is whether the learned camera representations encode genuine 3D geometry or merely exploit dataset-specific shortcuts~\cite{mitchel2025xfactor}.
We study the learned poses from two complementary angles:
i) how accurate and \emph{transferable} they are relative to prior work, and
ii) whether their accuracy improves \emph{predictably} as we scale data, model size, and compute -- mirroring the NVS scaling behavior of \cref{sec:exp_scaling}.
We read out the predicted poses in two ways:
a per-scene \emph{probe} that regresses ground-truth poses from frozen camera tokens (following RayZer~\cite{jiang2025rayzer}; see \cref{sec:app_pose_probe_implementation}), and a cross-scene \emph{transfer} protocol that applies a trajectory estimated on one scene to render another (following XFactor~\cite{mitchel2025xfactor}).
Both are evaluated zero-shot on DL3DV-10K~\cite{ling2024dl3dv} and summarized in the tables as rotation/translation accuracies thresholded at $\alpha \in \{10^\circ, 20^\circ, 30^\circ\}$ (R@$\alpha$) and $\alpha \in \{0.1, 0.2, 0.3\}$ (t@$\alpha$).

\paragraph{Pose Transferability}\label{sec:exp_pose_transferability}
We first establish that \methodname's learned poses are genuinely transferable at our final model (\cref{tab:experiment_transferability}), following \citet{mitchel2025xfactor}.
Despite a simpler setup, \methodname matches the specialized XFactor~\cite{mitchel2025xfactor}, which introduces explicit transferability supervision and requires multi-stage training for multi-view NVS, and substantially improves over RayZer~\cite{jiang2025rayzer}, whose poses were shown to lack transferability~\cite{mitchel2025xfactor}.
This suggests that our architectural choices -- particularly autoregressive pose learning -- resolve the transferability limitations of earlier systems without dedicated transferability objectives.

\begin{table}[t]
    \centering
    \caption{\textbf{Pose Transferability}. Evaluation of TPS metric from XFactor~\citep{mitchel2025xfactor} on DL3DV10k~\citep{ling2024dl3dv}. We follow their protocol and measure the accuracy of the transferred trajectory. We find that \methodname, like XFactor, significantly improves pose transferability compared to RayZer, without the need for explicit transferability training.}
    \adjustbox{max width=\linewidth}{\scalebox{0.8}{
    \begin{tabular}{lc@{\hskip .5em}c@{\hskip .5em}c@{\hskip .5em}c@{\hskip .5em}c@{\hskip .5em}c}\toprule
         Model & \textsc{\smalltableheaderfont R@10$^\circ\!\!\uparrow$} & \textsc{\smalltableheaderfont R@20$^\circ\!\!\uparrow$} & \textsc{\smalltableheaderfont R@30$^\circ\!\!\uparrow$} & \textsc{\smalltableheaderfont T@10$^\circ\!\!\uparrow$} & \textsc{\smalltableheaderfont T@20$^\circ\!\!\uparrow$} & \textsc{\smalltableheaderfont T@30$^\circ\!\!\uparrow$}\\
         \midrule
         RayZer~\citep{jiang2025rayzer} & 0.48 & 0.61 & 0.88 & 0.12 & 0.32 & 0.44 \\
         XFactor~\citep{mitchel2025xfactor} &  \samebf{0.93}&  \underline{0.97} &  \samebf{0.99} &  \samebf{0.55}&  \samebf{0.83} & \samebf{0.90} \\
         \methodname-L (Ours) & \underline{0.92} & \samebf{0.98} & \samebf{0.99} & \underline{0.44} & \samebf{0.83} & \samebf{0.90}\\
         \bottomrule
    \end{tabular}
    }}
    \label{tab:experiment_transferability} 
\end{table}

\begin{figure}[t]
    \centering
    \includegraphics[width=\linewidth]{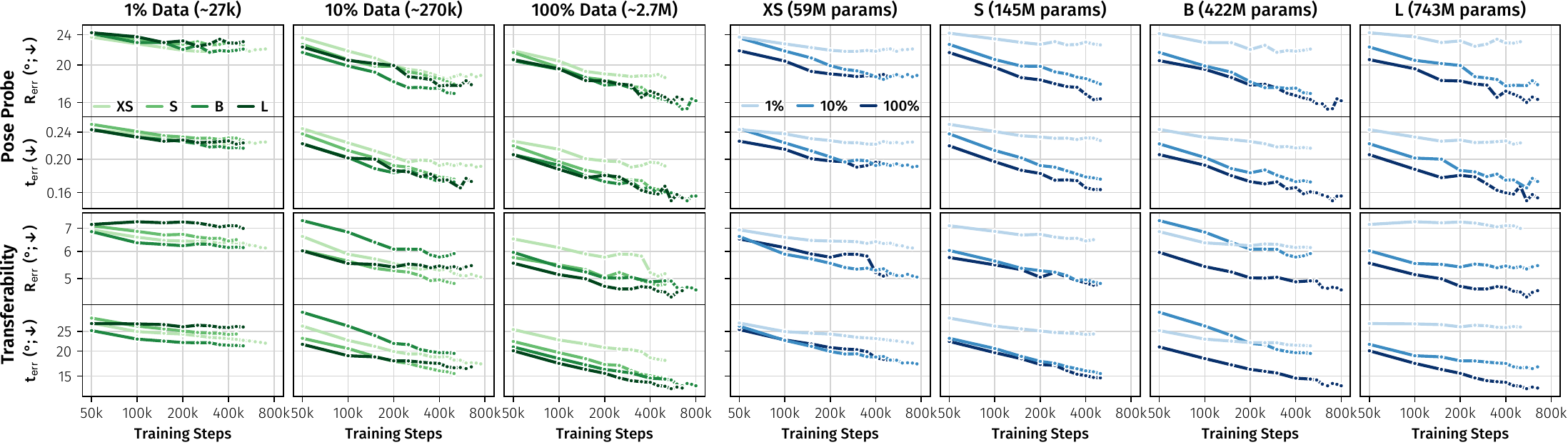}
    \caption{\textbf{Learned Camera Geometry Scales with Data, Model Size, and Compute.} We track the four continuous camera pose errors -- rotation and translation, each read out both via a \emph{probe} on the camera tokens (RayZer~\cite{jiang2025rayzer} protocol; top rows) and via cross-scene \emph{transfer} (XFactor~\cite{mitchel2025xfactor} protocol; bottom rows) -- as a function of training compute, evaluated zero-shot on DL3DV-10K~\cite{ling2024dl3dv}. \textit{Left:} all errors decrease consistently with training \emph{data} scale. \textit{Right:} all errors decrease with \emph{model} scale, with insufficient data again imposing a strong ceiling. Notably, there is \emph{no} significant saturation at scale, indicating that further scaling will likely be beneficial.}
    \vspace{-1em}
    \label{fig:camera_scaling}
\end{figure}

\paragraph{Pose Accuracy Scales with Data, Model Size, and Compute}
Beyond matching prior work at our final scale, the \emph{quality} of the learned geometry scales as orderly as NVS quality (\cref{fig:camera_scaling}):
the continuous rotation and translation errors (the continuous quantities that the thresholded accuracies in \cref{tab:experiment_transferability,tab:main_ablation} discretize) decrease monotonically and predictably with more data and larger models, with too little data again imposing a ceiling that larger models cannot overcome, exactly as for the NVS metrics in \cref{sec:exp_scaling}.

A natural worry is that scaling merely sharpens (dataset-specific) shortcuts rather than improving genuine geometry~\cite{mitchel2025xfactor}.
The per-scene \emph{probe} and cross-scene \emph{transfer} errors, however, improve simultaneously at every scale (\cref{fig:camera_scaling}).
Were scaling solely improving shortcut behavior, transfer would not track the probe -- their tight coupling indicates that scaling improves learning of genuine, transferable 3D geometry.

\subsection{Open-set Novel View Synthesis}\label{sec:exp_open_set}
Most prior self-supervised NVS methods~\citep[cf.,][]{jiang2025rayzer,mitchel2025xfactor,wang2025less} focus on closed-domain evaluation, training and testing on the same datasets.
We train a single \methodname model on generic data and evaluate it zero-shot across a wide range of datasets (LLFF~\citep{mildenhall2019llff}, DTU~\citep{jensen2014large}, CO3D~\citep{liu24uco3d}, WildRGBD~\citep{xia2024rgbd}, Mip-NeRF 360~\citep{barron2022mip}, and Tanks \& Temples~\citep{knapitsch2017tanks}), camera baselines, and numbers of input views, extending the extensive evaluation by \citet{zhou2025stable} in \Cref{tab:open_set_nvs}.
This setting better reflects real-world deployments and avoids dataset-specific tuning.
We note that, unlike the supervised baselines, \methodname uses its own predicted camera poses at inference, not the dataset's ground truth annotations, making this a strictly harder setting.

\begin{table}[t]
    \centering
    \caption{\textbf{Open-set Novel View Synthesis (PSNR$\uparrow$).} We extend the evaluation by \citet{zhou2025stable} and compute PSNR across a large variety of settings (columns). Despite being trained fully self-supervised and without large-scale video diffusion pretraining, RayDer is (near-)state-of-the-art across the majority of datasets and evaluation settings.}
    \ifeccv\vspace{-.7em}\else\vspace{-.4em}\fi
    \ifeccv
        \newcolumntype{C}{>{\hspace{-.5pt}\footnotesize}c<{\hspace{-.5pt}}}
    \else
        \newcolumntype{C}{@{\hskip .15em}c@{\hskip .15em}}
    \fi
    \newcommand{\ti}[1]{\multicolumn{1}{c}{#1}}
    \adjustbox{max width=\linewidth}{\scalebox{0.8}{
      \begin{tabular}{l@{\hskip \ifeccv-.4em\else0em\fi}crCCCCCCCCCC c CCCCCCC}
        & & & \multicolumn{10}{c}{\textit{small-viewpoint}} & {\hskip .2em} & \multicolumn{7}{c}{\textit{large-viewpoint}} \\
        \cmidrule(lr){4-13} \cmidrule(lr){15-21}
        \toprule

        & &  Dataset $\!\!\rightarrow\!\!\!\!\!\!\!$ & \multicolumn{2}{c}{LLFF} & \multicolumn{2}{c}{DTU} & \multicolumn{2}{c}{CO3D} & \multicolumn{2}{c}{WRGBD} & {\normalsize {\hskip -1em}M360{\hskip -1em}} & {\normalsize {\hskip -1em}T\&T{\hskip -1em}} & & {\normalsize {\hskip -1em}CO3D{\hskip -1em}} & \multicolumn{2}{c}{WRGBD} & \multicolumn{2}{c}{M360} &\multicolumn{2}{c}{T\&T}\\
        \cmidrule(lr){4-5} \cmidrule(lr){6-7} \cmidrule(lr){8-9} \cmidrule(lr){10-11} \cmidrule(lr){12-12} \cmidrule(lr){13-13} \cmidrule(lr){15-15} \cmidrule(lr){16-17} \cmidrule(lr){18-19} \cmidrule(lr){20-21}
        & & Split $\!\!\rightarrow\!\!\!\!\!\!\!$ & \multicolumn{2}{c}{R} & \multicolumn{2}{c}{R} & \ti{V} & \ti{R} & \ti{S\textsubscript{e}} & \ti{S\textsubscript{h}} & \ti{R} & \ti{V} & & \ti{R} & \multicolumn{2}{c}{S\textsubscript{h}} & \multicolumn{2}{c}{R} & \multicolumn{2}{c}{S} \\
        \cmidrule(lr){4-5} \cmidrule(lr){6-7} \cmidrule(lr){8-8} \cmidrule(lr){9-9} \cmidrule(lr){10-10} \cmidrule(lr){11-11} \cmidrule(lr){12-12} \cmidrule(lr){13-13} \cmidrule(lr){15-15} \cmidrule(lr){16-17} \cmidrule(lr){18-19} \cmidrule(lr){20-21}
        Model & Params & \smash{\shortstack{{\vspace{-.5em}\shortstack{Self-\\sup.}}}} $|\mathcal{I}_\text{in}|\!\!\rightarrow\!\!\!\!\!\!\!$& \ti{1} & \ti{3} & \ti{1} & \ti{3} & \ti{1} & \ti{3} & \ti{3} & \ti{6} & \ti{6} & \ti{1} & & \ti{1} & \ti{1} & \ti{3} & \ti{1} & \ti{3} & \ti{3} & \ti{6} \\
        \midrule
        
        MVSplat~\citep{chen2024mvsplat} & 12M & {\color{ourred}\xmark}{\hskip 2.8em} & 11.23& 12.50& 13.87& 15.52& 12.52& 13.52& 14.56& 12.54&13.56 &13.22 & & -- & -- & -- & -- & -- & -- & -- \\
        DepthSplat~\citep{xu2025depthsplat} & 354M & {\color{ourred}\xmark}{\hskip 2.8em} & 12.07& 12.62& \underline{14.15}& 16.24 & 13.23 & 13.77 & 15.93 & 14.23 & 14.01 &14.35 & & 10.42 & \phantom{0}9.35 & 13.53 & 10.49 & 12.54 & \phantom{0}9.78 & 10.12 \\
        ViewCrafter$^\dagger$~\citep{yu2024viewcraftertamingvideodiffusion} & 1.4B & {\color{ourred}\xmark}{\hskip 2.8em} & 10.53& 13.52& 12.66 & {16.40} & \underline{18.96} & 14.72 & 16.42 & 12.66 & 14.59 &\underline{18.07} & & 10.11 & \phantom{0}9.12 & 13.45 & \phantom{0}9.79 & 10.34 & \phantom{0}9.88 & 10.32 \\
        SEVA$^\dagger$~\citep{zhou2025stable} & 1.3B & {\color{ourred}\xmark}{\hskip 2.8em} &  {14.03}& {19.48}& {14.47} & \samebf{20.82} & 18.40 & \samebf{19.25} & \underline{19.75} & \samebf{18.91} & \underline{16.70} &15.16 & & \underline{15.30} & \underline{14.37} & \samebf{17.28} & 12.93 & 15.78 & {12.65} & {13.80} \\
        Kaleido$^{\dagger\ddagger}$~\cite{liu2025scaling} & 3.1B & {\color{ourred}\xmark}{\hskip 2.8em} & \underline{15.34} & \underline{20.71} & -- & -- & -- & -- & -- & -- & \samebf{18.03} & -- & & -- & -- & -- & \underline{13.74} & \samebf{16.78} & \underline{13.20} & \samebf{14.61} \\
        
        \midrule
        E-RayZer$^*$ ~\citep{zhao2025erayzer} & 246M & {\color{ourgreen}\cmark}{\hskip 2.7em} & 10.44 & 18.01 & 10.31 & 16.97 & 12.94 & 17.76 & 17.72 & 16.18 & 15.86 & 10.36 & & 12.94 & 10.53 & 14.47 & 9.78 & 15.17 & 12.88 & 13.35 \\
        
        RayDer-L-$576^2$ (Ours) & 743M & {\color{ourgreen}\cmark}{\hskip 2.7em} & \samebf{17.11} & \samebf{21.38} & \samebf{16.01} & \underline{17.92}  & \samebf{21.10} & \underline{19.09} & \samebf{20.07} & \underline{17.23} & {16.25} & \samebf{18.74} & & \samebf{16.84} & \samebf{14.55} & \underline{15.97} & \samebf{14.96} & \underline{15.85} & \samebf{13.59} & \underline{13.81} \\
        
        \bottomrule
        \multicolumn{13}{c}{\ifeccv\scriptsize\else\footnotesize\fi \shortstack{\vphantom{$^\dagger$}Split abbreviations: R: ReconFusion~\cite{wu2024reconfusion}; V: ViewCrafter~\citep{yu2024viewcraftertamingvideodiffusion}; S\textsubscript{\{e,h\}}: SEVA~\citep{zhou2025stable}, easy (e) and hard (h) variants.\\[-.33em]Dataset references: LLFF~\cite{mildenhall2019llff}, DTU~\cite{jensen2014large}, CO3D~\cite{liu24uco3d}, WRGBD~\cite{xia2024rgbd}, M360~\cite{barron2022mip}, T\&T~\cite{knapitsch2017tanks}}} & &
        \multicolumn{7}{c}{\ifeccv\scriptsize\else\footnotesize\fi \shortstack{$^\ddagger$Kaleido evaluates at $512^2$ instead of $576^2$\\[-.33em]$^\dagger$Diffusion-based models. $^*$Multi-dataset Ckpt}}
    \end{tabular}
    }}
    \ifeccv\vspace{-1em}\else\vspace{-.5em}\fi
    \label{tab:open_set_nvs}
\end{table}

Despite being trained fully self-supervised, from scratch in a single stage, \methodname achieves \textbf{state-of-the-art or near-state-of-the-art performance across the majority of settings} at more than an order of magnitude less training compute.
It is competitive with much larger models such as SEVA and Kaleido, which rely on large-scale video diffusion pretraining.
\methodname achievse this while requiring neither pose supervision at train or test time nor pretrained foundation model weights -- a substantially more constrained and scalable setup.
This is unlike E-RayZer~\cite{zhao2026erayzer}, which requires static-scene videos for training and, while combining a large number of static-scene datasets, is still significantly limited in the amount of training data it can use, limiting scaling (see also \cref{sec:exp_static_vs_dynamic_data}).

On lab datasests such as DTU, just like E-RayZer~\cite{zhao2026erayzer}, \methodname underperforms supervised methods.
We find this is primarily due to unreliable pose estimation in regimes (perfectly clean backgrounds with no structure) not present in typical general video training data: \methodname is trained exclusively on unconstrained \emph{real-world} video.
We view this as an expected limitation of the training data distribution rather than a failure of the approach.

\paragraph{Qualitative Results}
\Cref{fig:qualitative_main} compares \methodname against E-RayZer~\cite{zhao2025erayzer} across three challenging regimes -- sparse-view NVS, extreme wide-baseline interpolation, and some settings from \cref{tab:open_set_nvs} -- where \methodname produces markedly sharper and more consistent novel views.
Further samples are in \cref{sec:app_additional_samples}.

\subsection{Closed-Set Static \& Supervised Comparison}\label{sec:exp_prior_art}
We further compare to previous methods in closed-set settings on small-scale static datasets.
In the dense 24-view DL3DV-10K~\citep{ling2024dl3dv} setting introduced by RayZer~\citep{jiang2025rayzer} (\cref{tab:prior_dl3dv_trained}), \methodname is competitive with the state-of-the-art, despite being neither intended nor optimized for this setting -- our other experiments use one to three orders of magnitude more training data.
This demonstrates that our adaptations for large-scale dynamic-scene training do not sacrifice small-scale static-scene capability.

\paragraph*{Can Supervision replace Self-Supervision when training on Dynamic Data?}
An important question is whether supervised mthods could simply use off-the-shelf pose estimators to train on the same large-scale video data.
We test this by training LVSM~\cite{jin2025lvsm} on SpatialVid~\cite{wang2025spatialvid} using pseudo-ground truth camera poses from MegaSaM~\cite{li2025megasam}.
Our self-supervised \methodname outperforms the supervised LVSM by a wide margin (\cref{tab:lvsm_vs_rayder}, +2.9dB PSNR), demonstrating that self-supervised pose learning can be substantially more effective than relying on pseudo-ground truth annotations in this data regime.
This result is practically significant:
obtaining pseudo-GT poses via MegaSaM for SpatialVid cost $\sim$69k~GPU-h~\citep{wang2025spatialvid}, more than an order of magnitude above the $\sim$1{,}2k~GPU-h required to \emph{train} the \methodname-B model in this comparison (\cref{tab:main_training_hparams}), making the supervised path both less effective and less efficient.

\begin{table}[t]
    \centering
    \begin{minipage}{.485\textwidth}
        \centering
        \captionof{table}{\textbf{Static-Dataset Comparison.} We extend the evaluation by \citet{jiang2025rayzer}, training and evaluating on dense-view DL3DV. We train our model with the same settings (transformer size, view count, training steps) as the baselines. Despite the various adaptations to enable training on general video, our model is competitive also in this setting.}
        \adjustbox{max width=\linewidth}{\scalebox{0.8}{
        \begin{tabular}{ll@{}c@{\hskip .5em}c@{\hskip .5em}c@{\hskip .5em}c}
            \toprule
            \multirow{2}{*}[-3pt]{Model} & \multicolumn{2}{c}{Training Data} & \multicolumn{3}{c}{DL3DV (Even~\cite{jiang2025rayzer})} \\
            \cmidrule(lr){2-3} \cmidrule(lr){4-6}
            & \textsc{\smalltableheaderfont Dataset} & \textsc{\smalltableheaderfont GT Pose} & \textsc{\smalltableheaderfont PSNR$\uparrow$} & \textsc{\smalltableheaderfont LPIPS$\downarrow$} & \textsc{\smalltableheaderfont SSIM$\uparrow$} \\
            \midrule
            GS-LRM~\citep{zhang2024gs} & DL3DV~\citep{ling2024dl3dv} & {\color{ourred}\cmark} & 23.49  & 0.252 & 0.712 \\
            LVSM~\citep{jin2025lvsm} & DL3DV~\citep{ling2024dl3dv} & {\color{ourred}\cmark} & 23.69 & 0.242 & 0.723 \\
            RayZer~\citep{jiang2025rayzer} & DL3DV~\citep{ling2024dl3dv} & {\color{ourgreen}\xmark} & \underline{24.36} & \underline{0.209} & \underline{0.757} \\
            \methodname-B (Ours)  & DL3DV~\citep{ling2024dl3dv} & {\color{ourgreen}\xmark} &  \samebf{24.51} & \samebf{0.142} & \samebf{0.758} \\
            \bottomrule
        \end{tabular}
        }}
        \label{tab:prior_dl3dv_trained}
    \end{minipage}%
    {\hskip .03\textwidth}%
    \begin{minipage}{.485\textwidth}
        \centering
        \captionof{table}{\textbf{Supervised Dynamic-Dataset Comparison.} Comparing our \methodname model with LVSM trained on dynamic videos with MegaSaM~\cite{li2025megasam} camera poses at matched settings (transformer size, view count, training steps) results in a major performance gain, despite not using any pose supervision. We also note that obtaining these pseudo-GT camera poses costs an order of magnitude more compute than \emph{training} either model.}
        \adjustbox{max width=\linewidth}{\scalebox{0.8}{
        \begin{tabular}{ll@{}c@{\hskip .5em}c@{\hskip .5em}c@{\hskip .5em}c}
            \toprule
            \multirow{2}{*}[-3pt]{Model} & \multicolumn{2}{c}{Training Data} & \multicolumn{3}{c}{RE10K NVS} \\
            \cmidrule(lr){2-3} \cmidrule(lr){4-6}
            & \textsc{\smalltableheaderfont Dataset} & \textsc{\smalltableheaderfont GT Pose} & \textsc{\smalltableheaderfont PSNR$\uparrow$} & \textsc{\smalltableheaderfont LPIPS$\downarrow$} & \textsc{\smalltableheaderfont SSIM$\uparrow$} \\
            \midrule
            LVSM~\citep{jin2025lvsm} & SpatialVid~\citep{wang2025spatialvid} & {\color{ourred}\cmark} & 25.44 & 0.184& 0.729\\
            \methodname-B (Ours) & SpatialVid~\citep{wang2025spatialvid} & {\color{ourgreen}\xmark} & \samebf{28.35} & \samebf{0.151} & \samebf{0.879} \\
            \bottomrule
        \end{tabular}
        }}\ifeccv\else\vspace{.815em}\fi
        \label{tab:lvsm_vs_rayder}
    \end{minipage}
\end{table}

\subsection{Limitations}

\begin{figure}[t]
    \centering
    \includegraphics[width=.7\linewidth]{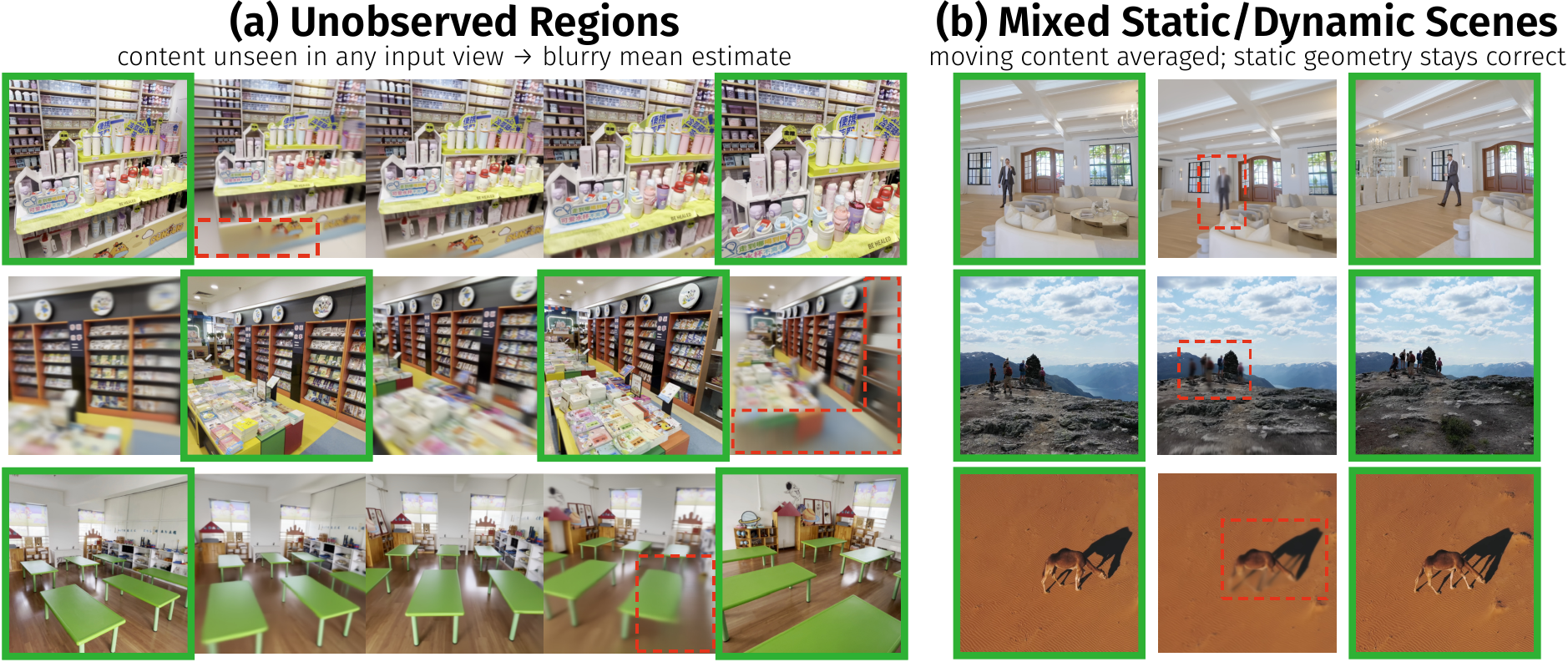}
    \caption{\textbf{Limitations.} Both main failure modes arise from the regression objective collapsing under-constrained content to a low-frequency average, dashed boxes mark affected regions. \textbf{(a)} content unseen in any input view is rendered as a blurry mean estimate. \textbf{(b)} in presence of dynamic content, the static scene is rendered correctly from the novel pose; moving content is averaged.}
    \label{fig:limitations_main}
\end{figure}

\paragraph{Unobserved Regions}
Content not visible in any context view is rendered as a blurry, low-frequency ``mean estimate'' rather than plausible but hallucinated detail (\cref{fig:limitations_main}a; see also \cref{fig:failurecase_warp}), a known consequence of the regression objective also shared by (E-)RayZer~\citep{jiang2025rayzer,zhao2025erayzer}, GS-LRM~\citep{zhang2024gs}, LVSM~\citep{jin2025lvsm}, and others.
The model provides no explicit signal that it is uncertain in these regions.
A generative or uncertainty-aware decoder, compatible with our unified backbone, is a natural direction for future work.

\paragraph{Mixed Static/Dynamic Scenes}
On real video containing both static structure and moving content, \methodname reconstructs the static geometry and renders it from the correct viewpoint, but dynamic content is not rendered faithfully, degrading to a mixture of blur and loose interpolation (\cref{fig:limitations_main}b).
This follows from treating the dynamic state as a nuisance factor (\cref{sec:method_dynamics}) rather than an explicit scene representation; we analyze the resulting entanglement of state and pose in \cref{sec:dynamic_state_exploration}.
An explicit, disentangled treatment~\citep[cf.,][]{seitzer2024dyst} would require multi-view videos of dynamic scenes, reintroducing exactly the dependence on scarce, curated data that our method is designed to avoid.
Extending to full dynamic (4D) NVS while retaining the ability to train on abundant generic video is left to future work.

\section{Conclusion}
Self-supervised novel view synthesis (NVS) has long promised scalability through videos by not requiring ground-truth camera pose annotations, yet existing approaches remained constrained by hard-to-scale multi-network pipelines and restrictive static-scene assumptions.
In this work, we introduced \methodname, a unified feed-forward transformer that consolidates camera estimation, scene reconstruction, and rendering into a single scalable backbone, enabling stable training on unconstrained real-world video while preserving the static-scene NVS objective.
Through explicit dynamic state handling via a nuisance variable, architectural unification, and autoregressive pose learning, \methodname makes scaling of self-supervised NVS clean across data, model size, and compute.
Empirically, this yields clean power-law scaling behavior, strong zero-shot open-set performance competitive with supervised and video diffusion-based systems, and transferable camera pose representations learned entirely without pose supervision.

Beyond the specific architecture, our results suggest a broader perspective: one major limitation of many prior self-supervised NVS methods was not the absence of supervision, but the inability to \emph{use scalable data regimes}.
By enabling stable learning from scratch on generic video and demonstrating clean scaling, \methodname positions self-supervised NVS within the same scaling-driven paradigm that has shaped progress in language and some vision foundation models.

Looking forward, \methodname opens up several directions for future work, including integration with partial supervision and generative modeling, extension toward 4D NVS, and continued scaling toward 3D world foundation models.

\section*{Acknowledgments}
This project has been supported by the Horizon Europe project ELLIOT (GA No.\ 101214398), the project ``GeniusRobot'' (01IS24083) funded by the Federal Ministry of Research, Technology and Space (BMFTR), the BMWE ZIM-project (No.\ KK5785001LO4) ``conIDitional LoRA'', the German Federal Ministry for Economic Affairs and Energy within the project ``NXT GEN AI METHODS - Generative Methoden für Perzeption, Prädiktion und Planung'', and the bidt project KLIMA-MEMES. The authors gratefully acknowledge the Gauss Center for Supercomputing for providing compute through the NIC on JUWELS/JUPITER at JSC and the HPC resources supplied by the NHR@FAU Erlangen.
We thank Olga Grebenkova, Kosta Derpanis, and Tommaso Martorella for feedback, proofreading, and helpful discussions, and Owen Vincent for technical support.

{
    \small
    \bibliographystyle{ieeenat_fullname}

    \bibliography{main}

@String(CVPR= {IEEE Conf. Comput. Vis. Pattern Recog.})

@String(TOG= {ACM Trans. Graph.})

@String(CVPR  = {CVPR})

@String(TOG   = {ACM TOG})

@inproceedings{huang2017arbitrary,
  title={Arbitrary style transfer in real-time with adaptive instance normalization},
  author={Huang, Xun and Belongie, Serge},
  booktitle={Proceedings of the IEEE international conference on computer vision},
  pages={1501--1510},
  year={2017}
}

@inproceedings{
dosovitskiy2021an,
title={An Image is Worth 16x16 Words: Transformers for Image Recognition at Scale},
author={Alexey Dosovitskiy and Lucas Beyer and Alexander Kolesnikov and Dirk Weissenborn and Xiaohua Zhai and Thomas Unterthiner and Mostafa Dehghani and Matthias Minderer and Georg Heigold and Sylvain Gelly and Jakob Uszkoreit and Neil Houlsby},
booktitle={International Conference on Learning Representations},
year={2021},
}

@article{su2021roformer,
  title={RoFormer: enhanced transformer with rotary position embedding. CoRR abs/2104.09864 (2021)},
  author={Su, Jianlin and Lu, Yu and Pan, Shengfeng and Wen, Bo and Liu, Yunfeng},
  journal={arXiv preprint arXiv:2104.09864},
  year={2021}
}

@InProceedings{crowson2024hourglass,
    title = 	 {Scalable High-Resolution Pixel-Space Image Synthesis with Hourglass Diffusion Transformers},
    author =       {Crowson, Katherine and Baumann, Stefan Andreas and Birch, Alex and Abraham, Tanishq Mathew and Kaplan, Daniel Z and Shippole, Enrico},
    booktitle = 	 {Proceedings of the 41st International Conference on Machine Learning},
    pages = 	 {9550--9575},
    year = 	 {2024},
    editor = 	 {Salakhutdinov, Ruslan and Kolter, Zico and Heller, Katherine and Weller, Adrian and Oliver, Nuria and Scarlett, Jonathan and Berkenkamp, Felix},
    volume = 	 {235},
    series = 	 {Proceedings of Machine Learning Research},
    month = 	 {21--27 Jul},
    publisher =    {PMLR}
}

@inproceedings{
loshchilov2018decoupled,
title={Decoupled Weight Decay Regularization},
author={Ilya Loshchilov and Frank Hutter},
booktitle={International Conference on Learning Representations},
year={2019},
}

@article{zhang2019root,
  title={Root mean square layer normalization},
  author={Zhang, Biao and Sennrich, Rico},
  journal={Advances in Neural Information Processing Systems},
  volume={32},
  year={2019}
}

@inproceedings{zhang2018unreasonable,
  title={The unreasonable effectiveness of deep features as a perceptual metric},
  author={Zhang, Richard and Isola, Phillip and Efros, Alexei A and Shechtman, Eli and Wang, Oliver},
  booktitle={Proceedings of the IEEE conference on computer vision and pattern recognition},
  pages={586--595},
  year={2018}
}

@inproceedings{jin2025lvsm,
title={LVSM: A Large View Synthesis Model with Minimal 3D Inductive Bias},
author={Haian Jin and Hanwen Jiang and Hao Tan and Kai Zhang and Sai Bi and Tianyuan Zhang and Fujun Luan and Noah Snavely and Zexiang Xu},
booktitle={The Thirteenth International Conference on Learning Representations},
year={2025},
}

@inproceedings{sajjadi2022scene,
  title={Scene representation transformer: Geometry-free novel view synthesis through set-latent scene representations},
  author={Sajjadi, Mehdi SM and Meyer, Henning and Pot, Etienne and Bergmann, Urs and Greff, Klaus and Radwan, Noha and Vora, Suhani and Lu{\v{c}}i{\'c}, Mario and Duckworth, Daniel and Dosovitskiy, Alexey and others},
  booktitle={Proceedings of the IEEE/CVF Conference on Computer Vision and Pattern Recognition},
  pages={6229--6238},
  year={2022}
}

@article{sajjadi2022object,
  title={Object scene representation transformer},
  author={Sajjadi, Mehdi SM and Duckworth, Daniel and Mahendran, Aravindh and Van Steenkiste, Sjoerd and Pavetic, Filip and Lucic, Mario and Guibas, Leonidas J and Greff, Klaus and Kipf, Thomas},
  journal={Advances in neural information processing systems},
  volume={35},
  pages={9512--9524},
  year={2022}
}

@inproceedings{sajjadi2023rust,
  title={Rust: Latent neural scene representations from unposed imagery},
  author={Sajjadi, Mehdi SM and Mahendran, Aravindh and Kipf, Thomas and Pot, Etienne and Duckworth, Daniel and Lu{\v{c}}i{\'c}, Mario and Greff, Klaus},
  booktitle={Proceedings of the IEEE/CVF Conference on Computer Vision and Pattern Recognition},
  pages={17297--17306},
  year={2023}
}

@article{jiang2025rayzer,
    title={RayZer: A Self-supervised Large View Synthesis Model},
    author={Jiang, Hanwen and Tan, Hao and Wang, Peng and Jin, Haian and Zhao, Yue and Bi, Sai and Zhang, Kai and Luan, Fujun and Sunkavalli, Kalyan and Huang, Qixing and Pavlakos, Georgios},
    booktitle={arXiv preprint arXiv:2505.00702},
    year={2025},
}

@inproceedings{seitzer2024dyst,
title={Dy{ST}: Towards Dynamic Neural Scene Representations on Real-World Videos},
author={Maximilian Seitzer and Sjoerd van Steenkiste and Thomas Kipf and Klaus Greff and Mehdi S. M. Sajjadi},
booktitle={The Twelfth International Conference on Learning Representations},
year={2024},
}

@article{huang2025no,
  title={No Pose at All: Self-Supervised Pose-Free 3D Gaussian Splatting from Sparse Views},
  author={Huang, Ranran and Mikolajczyk, Krystian},
  journal={arXiv preprint arXiv:2508.01171},
  year={2025}
}

@inproceedings{kang2025selfsplat,
  title={SelfSplat: Pose-free and 3D prior-free generalizable 3D Gaussian splatting},
  author={Kang, Gyeongjin and Yoo, Jisang and Park, Jihyeon and Nam, Seungtae and Im, Hyeonsoo and Shin, Sangheon and Kim, Sangpil and Park, Eunbyung},
  booktitle={Proceedings of the Computer Vision and Pattern Recognition Conference},
  pages={22012--22022},
  year={2025}
}

@inproceedings{lai2021video,
  title={Video autoencoder: self-supervised disentanglement of static 3d structure and motion},
  author={Lai, Zihang and Liu, Sifei and Efros, Alexei A and Wang, Xiaolong},
  booktitle={Proceedings of the IEEE/CVF International Conference on Computer Vision},
  pages={9730--9740},
  year={2021}
}

@inproceedings{safin2023repast,
    title	= {RePAST: Relative Pose Attention Scene Representation Transformer},author	= {Aleksandr Safin and Daniel Duckworth and Mehdi S. M. Sajjadi},year	= {2023}
}

@inproceedings{wu2025cat4d,
  title={Cat4d: Create anything in 4d with multi-view video diffusion models},
  author={Wu, Rundi and Gao, Ruiqi and Poole, Ben and Trevithick, Alex and Zheng, Changxi and Barron, Jonathan T and Holynski, Aleksander},
  booktitle={Proceedings of the Computer Vision and Pattern Recognition Conference},
  pages={26057--26068},
  year={2025}
}

@article{watson2024controlling,
  title={Controlling space and time with diffusion models},
  author={Watson, Daniel and Saxena, Saurabh and Li, Lala and Tagliasacchi, Andrea and Fleet, David J},
  journal={arXiv preprint arXiv:2407.07860},
  year={2024}
}

@article{zhou2025stable,
  title={Stable virtual camera: Generative view synthesis with diffusion models},
  author={Zhou, Jensen Jinghao and Gao, Hang and Voleti, Vikram and Vasishta, Aaryaman and Yao, Chun-Han and Boss, Mark and Torr, Philip and Rupprecht, Christian and Jampani, Varun},
  journal={arXiv preprint arXiv:2503.14489},
  year={2025}
}

@inproceedings{zhang2025world,
  title={World-consistent video diffusion with explicit 3d modeling},
  author={Zhang, Qihang and Zhai, Shuangfei and Martin, Miguel Angel Bautista and Miao, Kevin and Toshev, Alexander and Susskind, Joshua and Gu, Jiatao},
  booktitle={Proceedings of the Computer Vision and Pattern Recognition Conference},
  pages={21685--21695},
  year={2025}
}

@inproceedings{elata2025novel,
  title={Novel view synthesis with pixel-space diffusion models},
  author={Elata, Noam and Kawar, Bahjat and Ostrovsky-Berman, Yaron and Farber, Miriam and Sokolovsky, Ron},
  booktitle={Proceedings of the Computer Vision and Pattern Recognition Conference},
  pages={26756--26766},
  year={2025}
}

@article{li2025vicasplat,
  title={Vicasplat: A single run is all you need for 3d gaussian splatting and camera estimation from unposed video frames},
  author={Li, Zhiqi and Dong, Chengrui and Chen, Yiming and Huang, Zhangchi and Liu, Peidong},
  journal={arXiv preprint arXiv:2503.10286},
  year={2025}
}

@inproceedings{wang2024dust3r,
  title={Dust3r: Geometric 3d vision made easy},
  author={Wang, Shuzhe and Leroy, Vincent and Cabon, Yohann and Chidlovskii, Boris and Revaud, Jerome},
  booktitle={Proceedings of the IEEE/CVF Conference on Computer Vision and Pattern Recognition},
  pages={20697--20709},
  year={2024}
}

@inproceedings{leroy2024grounding,
  title={Grounding image matching in 3d with mast3r},
  author={Leroy, Vincent and Cabon, Yohann and Revaud, J{\'e}r{\^o}me},
  booktitle={European Conference on Computer Vision},
  pages={71--91},
  year={2024},
  organization={Springer}
}

@article{wang2025recollection,
  title={Recollection from Pensieve: Novel View Synthesis via Learning from Uncalibrated Videos},
  author={Wang, Ruoyu and Ma, Yi and Gao, Shenghua},
  journal={arXiv preprint arXiv:2505.13440},
  year={2025}
}

@inproceedings{zhang2024gs,
  title={Gs-lrm: Large reconstruction model for 3d gaussian splatting},
  author={Zhang, Kai and Bi, Sai and Tan, Hao and Xiangli, Yuanbo and Zhao, Nanxuan and Sunkavalli, Kalyan and Xu, Zexiang},
  booktitle={European Conference on Computer Vision},
  pages={1--19},
  year={2024},
  organization={Springer}
}

@article{hong2023lrm,
  title={Lrm: Large reconstruction model for single image to 3d},
  author={Hong, Yicong and Zhang, Kai and Gu, Jiuxiang and Bi, Sai and Zhou, Yang and Liu, Difan and Liu, Feng and Sunkavalli, Kalyan and Bui, Trung and Tan, Hao},
  journal={arXiv preprint arXiv:2311.04400},
  year={2023}
}

@inproceedings{zhou2019continuity,
  title={On the continuity of rotation representations in neural networks},
  author={Zhou, Yi and Barnes, Connelly and Lu, Jingwan and Yang, Jimei and Li, Hao},
  booktitle={Proceedings of the IEEE/CVF conference on computer vision and pattern recognition},
  pages={5745--5753},
  year={2019}
}

@inproceedings{usenko2018double,
  title={The double sphere camera model},
  author={Usenko, Vladyslav and Demmel, Nikolaus and Cremers, Daniel},
  booktitle={2018 International Conference on 3D Vision (3DV)},
  pages={552--560},
  year={2018},
  organization={IEEE}
}

@article{wang2025less,
  title={The Less You Depend, The More You Learn: Synthesizing Novel Views from Sparse, Unposed Images without Any 3D Knowledge},
  author={Wang, Haoru and Ye, Kai and Li, Yangyan and Chen, Wenzheng and Chen, Baoquan},
  journal={arXiv preprint arXiv:2506.09885},
  year={2025}
}

@InProceedings{wang2023vggsfm,
  author    = {Jianyuan Wang and Nikita Karaev and Christian Rupprecht and David Novotny},
  title     = {VGGSfM: Visual Geometry Grounded Deep Structure From Motion},
  year      = {2023}
}

@article{wang2025pi3,
  title={$\pi^3$: Scalable Permutation-Equivariant Visual Geometry Learning},
  author={Wang, Yifan and Zhou, Jianjun and Zhu, Haoyi and Chang, Wenzheng and Zhou, Yang and Li, Zizun and Chen, Junyi and Pang, Jiangmiao and Shen, Chunhua and He, Tong},
  journal={arXiv preprint arXiv:2507.13347},
  year={2025}
}

@article{plucker1865xvii,
  title={Xvii. on a new geometry of space},
  author={Plucker, Julius},
  journal={Philosophical Transactions of the Royal Society of London},
  number={155},
  pages={725--791},
  year={1865},
  publisher={The Royal Society London}
}

@article{mitchel2025xfactor,
  title={True Self-Supervised Novel View Synthesis is Transferable},
  author={Mitchel, Thomas W and Ryu, Hyunwoo and Sitzmann, Vincent},
  journal={arXiv preprint arXiv:2510.13063},
  year={2025}
}

@article{hinton2012improving,
  title={Improving neural networks by preventing co-adaptation of feature detectors},
  author={Hinton, Geoffrey E and Srivastava, Nitish and Krizhevsky, Alex and Sutskever, Ilya and Salakhutdinov, Ruslan R},
  journal={arXiv preprint arXiv:1207.0580},
  year={2012}
}

@inproceedings{nair2025scaling,
  title={Scaling Transformer-Based Novel View Synthesis with Models Token Disentanglement and Synthetic Data},
  author={Nair, Nithin Gopalakrishnan and Kaza, Srinivas and Luo, Xuan and Patel, Vishal M and Lombardi, Stephen and Park, Jungyeon},
  booktitle={Proceedings of the IEEE/CVF International Conference on Computer Vision},
  pages={28567--28576},
  year={2025}
}

@inproceedings{peebles2023scalable,
  title={Scalable diffusion models with transformers},
  author={Peebles, William and Xie, Saining},
  booktitle={Proceedings of the IEEE/CVF international conference on computer vision},
  pages={4195--4205},
  year={2023}
}

@article{wang2025scaling,
  title={Scaling laws in patchification: An image is worth 50,176 tokens and more},
  author={Wang, Feng and Yu, Yaodong and Wei, Guoyizhe and Shao, Wei and Zhou, Yuyin and Yuille, Alan and Xie, Cihang},
  journal={arXiv preprint arXiv:2502.03738},
  year={2025}
}

@article{ravi2024sam2,
  title={Sam 2: Segment anything in images and videos},
  author={Ravi, Nikhila and Gabeur, Valentin and Hu, Yuan-Ting and Hu, Ronghang and Ryali, Chaitanya and Ma, Tengyu and Khedr, Haitham and R{\"a}dle, Roman and Rolland, Chloe and Gustafson, Laura and others},
  journal={arXiv preprint arXiv:2408.00714},
  year={2024}
}

@misc{youtube2025press,
	author = {},
	title = {{Y}ou{T}ube for {P}ress --- blog.youtube},
	howpublished = {\url{https://blog.youtube/press/}},
	year = {},
	note = {[Accessed 09-11-2025]},
}

@article{wang2025spatialvid,
  title={Spatialvid: A large-scale video dataset with spatial annotations},
  author={Wang, Jiahao and Yuan, Yufeng and Zheng, Rujie and Lin, Youtian and Gao, Jian and Chen, Lin-Zhuo and Bao, Yajie and Zhang, Yi and Zeng, Chang and Zhou, Yanxi and others},
  journal={arXiv preprint arXiv:2509.09676},
  year={2025}
}

@inproceedings{schoenberger2016sfm,
    author={Sch\"{o}nberger, Johannes Lutz and Frahm, Jan-Michael},
    title={Structure-from-Motion Revisited},
    booktitle={Conference on Computer Vision and Pattern Recognition (CVPR)},
    year={2016},
}

@inproceedings{chen2024mvsplat,
  title={Mvsplat: Efficient 3d gaussian splatting from sparse multi-view images},
  author={Chen, Yuedong and Xu, Haofei and Zheng, Chuanxia and Zhuang, Bohan and Pollefeys, Marc and Geiger, Andreas and Cham, Tat-Jen and Cai, Jianfei},
  booktitle={European Conference on Computer Vision},
  pages={370--386},
  year={2024},
  organization={Springer}
}

@article{kerbl20233d,
  title={3D Gaussian splatting for real-time radiance field rendering.},
  author={Kerbl, Bernhard and Kopanas, Georgios and Leimk{\"u}hler, Thomas and Drettakis, George},
  journal={ACM Trans. Graph.},
  volume={42},
  number={4},
  pages={139--1},
  year={2023}
}

@inproceedings{mildenhall2020nerf,
  title={NeRF: Representing Scenes as Neural Radiance Fields for View Synthesis},
  author={Mildenhall, Ben and Srinivasan, Pratul P and Tancik, Matthew and Barron, Jonathan T and Ramamoorthi, Ravi and Ng, Ren},
  booktitle={European Conference on Computer Vision},
  pages={405--421},
  year={2020},
  organization={Springer}
}

@inproceedings{charatan2024pixelsplat,
  title={pixelsplat: 3d gaussian splats from image pairs for scalable generalizable 3d reconstruction},
  author={Charatan, David and Li, Sizhe Lester and Tagliasacchi, Andrea and Sitzmann, Vincent},
  booktitle={Proceedings of the IEEE/CVF conference on computer vision and pattern recognition},
  pages={19457--19467},
  year={2024}
}

@article{watson2022novel,
  title={Novel view synthesis with diffusion models},
  author={Watson, Daniel and Chan, William and Martin-Brualla, Ricardo and Ho, Jonathan and Tagliasacchi, Andrea and Norouzi, Mohammad},
  journal={arXiv preprint arXiv:2210.04628},
  year={2022}
}

@article{hong2024pf3plat,
  title={Pf3plat: Pose-free feed-forward 3d gaussian splatting},
  author={Hong, Sunghwan and Jung, Jaewoo and Shin, Heeseong and Han, Jisang and Yang, Jiaolong and Luo, Chong and Kim, Seungryong},
  journal={arXiv preprint arXiv:2410.22128},
  year={2024}
}

@article{smith2023flowcam,
  title={Flowcam: Training generalizable 3d radiance fields without camera poses via pixel-aligned scene flow},
  author={Smith, Cameron and Du, Yilun and Tewari, Ayush and Sitzmann, Vincent},
  journal={arXiv preprint arXiv:2306.00180},
  year={2023}
}

@inproceedings{chen2023dbarf,
  title={Dbarf: Deep bundle-adjusting generalizable neural radiance fields},
  author={Chen, Yu and Lee, Gim Hee},
  booktitle={Proceedings of the IEEE/CVF Conference on Computer Vision and Pattern Recognition},
  pages={24--34},
  year={2023}
}

@inproceedings{fu2024colmap,
  title={Colmap-free 3d gaussian splatting},
  author={Fu, Yang and Liu, Sifei and Kulkarni, Amey and Kautz, Jan and Efros, Alexei A and Wang, Xiaolong},
  booktitle={Proceedings of the IEEE/CVF Conference on Computer Vision and Pattern Recognition},
  pages={20796--20805},
  year={2024}
}

@article{huang2025spfsplatv2,
  title={SPFSplatV2: Efficient Self-Supervised Pose-Free 3D Gaussian Splatting from Sparse Views},
  author={Huang, Ranran and Mikolajczyk, Krystian},
  journal={arXiv preprint arXiv:2509.17246},
  year={2025}
}

@article{ye2024no,
  title={No pose, no problem: Surprisingly simple 3d gaussian splats from sparse unposed images},
  author={Ye, Botao and Liu, Sifei and Xu, Haofei and Li, Xueting and Pollefeys, Marc and Yang, Ming-Hsuan and Peng, Songyou},
  journal={arXiv preprint arXiv:2410.24207},
  year={2024}
}

@article{keetha2025mapanything,
  title={MapAnything: Universal feed-forward metric 3D reconstruction},
  author={Keetha, Nikhil and M{\"u}ller, Norman and Sch{\"o}nberger, Johannes and Porzi, Lorenzo and Zhang, Yuchen and Fischer, Tobias and Knapitsch, Arno and Zauss, Duncan and Weber, Ethan and Antunes, Nelson and others},
  journal={arXiv preprint arXiv:2509.13414},
  year={2025}
}

@inproceedings{wang2025vggt,
  title={Vggt: Visual geometry grounded transformer},
  author={Wang, Jianyuan and Chen, Minghao and Karaev, Nikita and Vedaldi, Andrea and Rupprecht, Christian and Novotny, David},
  booktitle={Proceedings of the Computer Vision and Pattern Recognition Conference},
  pages={5294--5306},
  year={2025}
}

@article{shen2025fastvggt,
  title={Fastvggt: Training-free acceleration of visual geometry transformer},
  author={Shen, You and Zhang, Zhipeng and Qu, Yansong and Cao, Liujuan},
  journal={arXiv preprint arXiv:2509.02560},
  year={2025}
}

@article{deng2025vggtlong,
  title={VGGT-Long: Chunk it, Loop it, Align it--Pushing VGGT's Limits on Kilometer-scale Long RGB Sequences},
  author={Deng, Kai and Ti, Zexin and Xu, Jiawei and Yang, Jian and Xie, Jin},
  journal={arXiv preprint arXiv:2507.16443},
  year={2025}
}

@article{feng2025quantized,
  title={Quantized Visual Geometry Grounded Transformer},
  author={Feng, Weilun and Qin, Haotong and Wu, Mingqiang and Yang, Chuanguang and Li, Yuqi and Li, Xiangqi and An, Zhulin and Huang, Libo and Zhang, Yulun and Magno, Michele and others},
  journal={arXiv preprint arXiv:2509.21302},
  year={2025}
}

@inproceedings{wang2025continuous,
  title={Continuous 3d perception model with persistent state},
  author={Wang, Qianqian and Zhang, Yifei and Holynski, Aleksander and Efros, Alexei A and Kanazawa, Angjoo},
  booktitle={Proceedings of the Computer Vision and Pattern Recognition Conference},
  pages={10510--10522},
  year={2025}
}

@inproceedings{rombach2021geometry,
  title={Geometry-free view synthesis: Transformers and no 3d priors},
  author={Rombach, Robin and Esser, Patrick and Ommer, Bj{\"o}rn},
  booktitle={Proceedings of the IEEE/CVF International Conference on Computer Vision},
  pages={14356--14366},
  year={2021}
}

@inproceedings{ling2024dl3dv,
  title={Dl3dv-10k: A large-scale scene dataset for deep learning-based 3d vision},
  author={Ling, Lu and Sheng, Yichen and Tu, Zhi and Zhao, Wentian and Xin, Cheng and Wan, Kun and Yu, Lantao and Guo, Qianyu and Yu, Zixun and Lu, Yawen and others},
  booktitle={Proceedings of the IEEE/CVF Conference on Computer Vision and Pattern Recognition},
  pages={22160--22169},
  year={2024}
}

@inproceedings{hassani2023neighborhood,
  title={Neighborhood attention transformer},
  author={Hassani, Ali and Walton, Steven and Li, Jiachen and Li, Shen and Shi, Humphrey},
  booktitle={Proceedings of the IEEE/CVF conference on computer vision and pattern recognition},
  pages={6185--6194},
  year={2023}
}

@misc{zhao2025erayzer,
      title={E-RayZer: Self-supervised 3D Reconstruction as Spatial Visual Pre-training}, 
      author={Qitao Zhao and Hao Tan and Qianqian Wang and Sai Bi and Kai Zhang and Kalyan Sunkavalli and Shubham Tulsiani and Hanwen Jiang},
      year={2025},
      eprint={2512.10950},
      archivePrefix={arXiv},
      primaryClass={cs.CV},
}

@inproceedings{li2025megasam,
  title={MegaSaM: Accurate, fast and robust structure and motion from casual dynamic videos},
  author={Li, Zhengqi and Tucker, Richard and Cole, Forrester and Wang, Qianqian and Jin, Linyi and Ye, Vickie and Kanazawa, Angjoo and Holynski, Aleksander and Snavely, Noah},
  booktitle={Proceedings of the Computer Vision and Pattern Recognition Conference},
  pages={10486--10496},
  year={2025}
}

@article{zhou2018stereo,
  title={Stereo magnification: Learning view synthesis using multiplane images},
  author={Zhou, Tinghui and Tucker, Richard and Flynn, John and Fyffe, Graham and Snavely, Noah},
  journal={arXiv preprint arXiv:1805.09817},
  year={2018}
}

@misc{wang2025OpenRayzer,
	author = {Wang, Haoru},
	title = {{O}pen-{R}ayzer: a open-source {S}elf-{R}eimplemented {V}ersion of the paper "{R}ay{Z}er: {A} {S}elf-supervised {L}arge {V}iew {S}ynthesis {M}odel" --- github.com},
	howpublished = {\url{https://github.com/ou524u/Open-Rayzer}},
	year = {2025},
}

@misc{buchnerimagehash,
	author = {Buchner, Johannes},
	title = {{I}mage{H}ash: A Python Perceptual Image Hashing Module --- github.com},
	howpublished = {\url{https://github.com/JohannesBuchner/imagehash}},
	year = {2025},
}

@misc{klingerphash,
  author       = {Klinger, Evan and Starkweather, David},
  title        = {{pHash}: The Open Source Perceptual Hash Library},
  year         = {2010},
  howpublished = {\url{https://www.phash.org/}}
}

@misc{yu2024viewcraftertamingvideodiffusion,
      title={ViewCrafter: Taming Video Diffusion Models for High-fidelity Novel View Synthesis}, 
      author={Wangbo Yu and Jinbo Xing and Li Yuan and Wenbo Hu and Xiaoyu Li and Zhipeng Huang and Xiangjun Gao and Tien-Tsin Wong and Ying Shan and Yonghong Tian},
      year={2024},
      eprint={2409.02048},
      archivePrefix={arXiv},
      primaryClass={cs.CV},
}

@inproceedings{xu2025depthsplat,
  title={Depthsplat: Connecting gaussian splatting and depth},
  author={Xu, Haofei and Peng, Songyou and Wang, Fangjinhua and Blum, Hermann and Barath, Daniel and Geiger, Andreas and Pollefeys, Marc},
  booktitle={Proceedings of the Computer Vision and Pattern Recognition Conference},
  pages={16453--16463},
  year={2025}
}

@article{kaplan2020scaling,
  title={Scaling laws for neural language models},
  author={Kaplan, Jared and McCandlish, Sam and Henighan, Tom and Brown, Tom B and Chess, Benjamin and Child, Rewon and Gray, Scott and Radford, Alec and Wu, Jeffrey and Amodei, Dario},
  journal={arXiv preprint arXiv:2001.08361},
  year={2020}
}

@article{henighan2020scaling,
  title={Scaling laws for autoregressive generative modeling},
  author={Henighan, Tom and Kaplan, Jared and Katz, Mor and Chen, Mark and Hesse, Christopher and Jackson, Jacob and Jun, Heewoo and Brown, Tom B and Dhariwal, Prafulla and Gray, Scott and others},
  journal={arXiv preprint arXiv:2010.14701},
  year={2020}
}

@inproceedings{wu2024reconfusion,
  title={Reconfusion: 3d reconstruction with diffusion priors},
  author={Wu, Rundi and Mildenhall, Ben and Henzler, Philipp and Park, Keunhong and Gao, Ruiqi and Watson, Daniel and Srinivasan, Pratul P and Verbin, Dor and Barron, Jonathan T and Poole, Ben and others},
  booktitle={Proceedings of the IEEE/CVF conference on computer vision and pattern recognition},
  pages={21551--21561},
  year={2024}
}

@article{hu2024minicpm,
  title={Minicpm: Unveiling the potential of small language models with scalable training strategies},
  author={Hu, Shengding and Tu, Yuge and Han, Xu and He, Chaoqun and Cui, Ganqu and Long, Xiang and Zheng, Zhi and Fang, Yewei and Huang, Yuxiang and Zhao, Weilin and others},
  journal={arXiv preprint arXiv:2404.06395},
  year={2024}
}

@inproceedings{
ghorbani2022scalingnmt,
title={Scaling Laws for Neural Machine Translation},
author={Behrooz Ghorbani and Orhan Firat and Markus Freitag and Ankur Bapna and Maxim Krikun and Xavier Garcia and Ciprian Chelba and Colin Cherry},
booktitle={International Conference on Learning Representations},
year={2022},
}

@inproceedings{liu24uco3d,
                Author = {Liu, Xingchen and Tayal, Piyush and Wang, Jianyuan
                          and Zarzar, Jesus and Monnier, Tom and Tertikas, Konstantinos
                          and Duan, Jiali and Toisoul, Antoine and Zhang, Jason Y.
                          and Neverova, Natalia and Vedaldi, Andrea
                          and Shapovalov, Roman and Novotny, David},
                Booktitle = {arXiv},
                Title = {UnCommon Objects in 3D},
                Year = {2024},
            }

@article{liu2025scaling,
  title={Scaling Sequence-to-Sequence Generative Neural Rendering},
  author={Liu, Shikun and Ng, Kam Woh and Jang, Wonbong and Guo, Jiadong and Han, Junlin and Liu, Haozhe and Douratsos, Yiannis and P{\'e}rez, Juan C and Zhou, Zijian and Phung, Chi and others},
  journal={arXiv preprint arXiv:2510.04236},
  year={2025}
}

@inproceedings{jensen2014large,
  title={Large scale multi-view stereopsis evaluation},
  author={Jensen, Rasmus and Dahl, Anders and Vogiatzis, George and Tola, Engil and Aan{\ae}s, Henrik},
  booktitle={2014 IEEE Conference on Computer Vision and Pattern Recognition},
  pages={406--413},
  year={2014},
  organization={IEEE}
}

@article{mildenhall2019llff,
  title={Local Light Field Fusion: Practical View Synthesis with Prescriptive Sampling Guidelines},
  author={Ben Mildenhall and Pratul P. Srinivasan and Rodrigo Ortiz-Cayon and Nima Khademi Kalantari and Ravi Ramamoorthi and Ren Ng and Abhishek Kar},
  journal={ACM Transactions on Graphics (TOG)},
  year={2019},
}

@inproceedings{reizenstein21co3d,
	Author = {Reizenstein, Jeremy and Shapovalov, Roman and Henzler, Philipp and Sbordone, Luca and Labatut, Patrick and Novotny, David},
	Booktitle = {International Conference on Computer Vision},
	Title = {Common Objects in 3D: Large-Scale Learning and Evaluation of Real-life 3D Category Reconstruction},
	Year = {2021},
}

@inproceedings{xia2024rgbd,
  title={Rgbd objects in the wild: Scaling real-world 3d object learning from rgb-d videos},
  author={Xia, Hongchi and Fu, Yang and Liu, Sifei and Wang, Xiaolong},
  booktitle={Proceedings of the IEEE/CVF Conference on Computer Vision and Pattern Recognition},
  pages={22378--22389},
  year={2024}
}

@inproceedings{barron2022mip,
  title={Mip-nerf 360: Unbounded anti-aliased neural radiance fields},
  author={Barron, Jonathan T and Mildenhall, Ben and Verbin, Dor and Srinivasan, Pratul P and Hedman, Peter},
  booktitle={Proceedings of the IEEE/CVF conference on computer vision and pattern recognition},
  pages={5470--5479},
  year={2022}
}

@article{knapitsch2017tanks,
    author    = {Arno Knapitsch and Jaesik Park and Qian-Yi Zhou and Vladlen Koltun},
    title     = {Tanks and Temples: Benchmarking Large-Scale Scene Reconstruction},
    journal   = {ACM Transactions on Graphics},
    volume    = {36},
    number    = {4},
    year      = {2017},
}

@inproceedings{gao2022dynamic,
    title={Monocular Dynamic View Synthesis: A Reality Check},
    author={Gao, Hang and Li, Ruilong and Tulsiani, Shubham and Russell, Bryan and Kanazawa, Angjoo},
    booktitle={NeurIPS},
    year={2022},
}

@misc{touvron2023llama2openfoundation,
      title={Llama 2: Open Foundation and Fine-Tuned Chat Models}, 
      author={Hugo Touvron and Louis Martin and Kevin Stone and Peter Albert and Amjad Almahairi and Yasmine Babaei and Nikolay Bashlykov and Soumya Batra and Prajjwal Bhargava and Shruti Bhosale and Dan Bikel and Lukas Blecher and Cristian Canton Ferrer and Moya Chen and Guillem Cucurull and David Esiobu and Jude Fernandes and Jeremy Fu and Wenyin Fu and Brian Fuller and Cynthia Gao and Vedanuj Goswami and Naman Goyal and Anthony Hartshorn and Saghar Hosseini and Rui Hou and Hakan Inan and Marcin Kardas and Viktor Kerkez and Madian Khabsa and Isabel Kloumann and Artem Korenev and Punit Singh Koura and Marie-Anne Lachaux and Thibaut Lavril and Jenya Lee and Diana Liskovich and Yinghai Lu and Yuning Mao and Xavier Martinet and Todor Mihaylov and Pushkar Mishra and Igor Molybog and Yixin Nie and Andrew Poulton and Jeremy Reizenstein and Rashi Rungta and Kalyan Saladi and Alan Schelten and Ruan Silva and Eric Michael Smith and Ranjan Subramanian and Xiaoqing Ellen Tan and Binh Tang and Ross Taylor and Adina Williams and Jian Xiang Kuan and Puxin Xu and Zheng Yan and Iliyan Zarov and Yuchen Zhang and Angela Fan and Melanie Kambadur and Sharan Narang and Aurelien Rodriguez and Robert Stojnic and Sergey Edunov and Thomas Scialom},
      year={2023},
      eprint={2307.09288},
      archivePrefix={arXiv},
      primaryClass={cs.CL},
}

@article{wang2004image,
  title={Image quality assessment: from error visibility to structural similarity},
  author={Wang, Zhou and Bovik, Alan C and Sheikh, Hamid R and Simoncelli, Eero P},
  journal={IEEE transactions on image processing},
  volume={13},
  number={4},
  pages={600--612},
  year={2004},
  publisher={IEEE}
}

@inproceedings{
fang2026incvggt,
title={Inc{VGGT}: Incremental {VGGT} for Memory-Bounded Long-Range 3D Reconstruction},
author={Keyu Fang and Changchun Zhou and Yuzhe Fu and Hai Helen Li and Yiran Chen},
booktitle={The Fourteenth International Conference on Learning Representations},
year={2026},
}

@inproceedings{
hoffmann2022an,
title={An empirical analysis of compute-optimal large language model training},
author={Jordan Hoffmann and Sebastian Borgeaud and Arthur Mensch and Elena Buchatskaya and Trevor Cai and Eliza Rutherford and Diego de las Casas and Lisa Anne Hendricks and Johannes Welbl and Aidan Clark and Tom Hennigan and Eric Noland and Katherine Millican and George van den Driessche and Bogdan Damoc and Aurelia Guy and Simon Osindero and Karen Simonyan and Erich Elsen and Oriol Vinyals and Jack William Rae and Laurent Sifre},
booktitle={Advances in Neural Information Processing Systems},
editor={Alice H. Oh and Alekh Agarwal and Danielle Belgrave and Kyunghyun Cho},
year={2022},
}

@article{chen2024mvsplat360,
  title={Mvsplat360: Feed-forward 360 scene synthesis from sparse views},
  author={Chen, Yuedong and Zheng, Chuanxia and Xu, Haofei and Zhuang, Bohan and Vedaldi, Andrea and Cham, Tat-Jen and Cai, Jianfei},
  journal={Advances in Neural Information Processing Systems},
  volume={37},
  pages={107064--107086},
  year={2024}
}

@inproceedings{zhao2026erayzer,
  title     = {E-RayZer: Self-supervised 3D Reconstruction as Spatial Visual Pre-training}, 
  author    = {Qitao Zhao and Hao Tan and Qianqian Wang and Sai Bi and Kai Zhang and Kalyan Sunkavalli and Shubham Tulsiani and Hanwen Jiang},
  booktitle = {CVPR},
  year      = {2026} 
}

@inproceedings{perazzi2016benchmark,
  title={A benchmark dataset and evaluation methodology for video object segmentation},
  author={Perazzi, Federico and Pont-Tuset, Jordi and McWilliams, Brian and Van Gool, Luc and Gross, Markus and Sorkine-Hornung, Alexander},
  booktitle={Proceedings of the IEEE conference on computer vision and pattern recognition},
  pages={724--732},
  year={2016}
}

@misc{simeoni2025dinov3,
  title={{DINOv3}},
  author={Sim{\'e}oni, Oriane and Vo, Huy V. and Seitzer, Maximilian and Baldassarre, Federico and Oquab, Maxime and Jose, Cijo and Khalidov, Vasil and Szafraniec, Marc and Yi, Seungeun and Ramamonjisoa, Micha{\"e}l and Massa, Francisco and Haziza, Daniel and Wehrstedt, Luca and Wang, Jianyuan and Darcet, Timoth{\'e}e and Moutakanni, Th{\'e}o and Sentana, Leonel and Roberts, Claire and Vedaldi, Andrea and Tolan, Jamie and Brandt, John and Couprie, Camille and Mairal, Julien and J{\'e}gou, Herv{\'e} and Labatut, Patrick and Bojanowski, Piotr},
  year={2025},
  eprint={2508.10104},
  archivePrefix={arXiv},
  primaryClass={cs.CV},
  url={https://arxiv.org/abs/2508.10104},
}
}

\clearpage
\ifeccv
    \setcounter{page}{1}
    \crefalias{section}{appsec}
    \crefalias{figure}{appfig}
    \crefalias{table}{apptab}
    \crefalias{equation}{appeq}
\fi

\setcounter{figure}{0}
\setcounter{table}{0}
\setcounter{equation}{0}
\setcounter{section}{0}
\renewcommand\thesection{\Alph{section}}
\renewcommand\thefigure{\Alph{section}.\arabic{figure}}
\renewcommand\thetable{\Alph{section}.\arabic{table}}
\renewcommand\theequation{\Alph{section}.\arabic{equation}}

\section{\todo{Extended Exploration Details}}\label{sec:app_exploration_details}
We show a full overview of our main exploration's results from \cref{sec:method} with extended metrics in \cref{tab:main_ablation}.
For config \textbf{A}, we repeated the training with different seeds until we got a run that did not diverge.
We attempted to train RayZer models in our setting (view count, training data), but those trainings consistently diverged, even after multiple attempts.

\paragraph{Training Details}\label{sec:app_training_details}
Models are trained on 8 frames (6 in, 2 out) at a resolution of $256^2$ extracted from the respective source videos at 2 fps.
We perform our exploration on two datasets in parallel:
i) Segment Anything-Video~\citep[SA-V,][]{ravi2024sam2}: a publicly available, highly diverse dataset that includes both dynamic cameras and highly dynamic scene content.
ii) SpatialVid-HQ~\citep[SV-HQ,][]{wang2025spatialvid}: a high-quality, curated dataset, which contains a mixture of dynamic and (mostly) static scenes.
This ensures that our findings generalize across different settings -- both truly open-set, highly dynamic videos, and more curated, yet not fully static-scene videos. Notably, SA-V also contains a significant fraction of videos with (almost) static cameras, which likely makes training more challenging.

We train these models for 200k steps at batch size 256 on 32 Nvidia H200 using AdamW~\citep{loshchilov2018decoupled} with a learning rate of $10^{-4}$, $(\beta_1, \beta_2) = (0.9, 0.95)$, and weight decay $0.01$, with a linear warmup over 1k steps and a constant learning rate after.
Unlike RayZer~\cite{jiang2025rayzer}, we do \emph{not} use any curricula (e.g., RayZer uses dataset-specific frame intervals that are scaled according to a predefined schedule throughout training).
This is an important step to reduce the scaling complexity, as proper scaling for such curricula with data complexity, data scale, training time, and model size is unclear.

Unlike previous works, we focus on \emph{zero-shot} evaluation on unseen datasets, reflecting real-world deployments.
We evaluate NVS performance on RealEstate-10k~\citep[RE10K,][]{zhou2018stereo} in the pixelSplat~\citep{charatan2024pixelsplat} setting, following standard feedforward NVS parameters. Camera poses are predicted by the model itself following standard practice for self-supervised NVS.
The accuracy of the predicted camera poses is evaluated using pointwise probes applied to camera tokens on DL3DV-10k~\citep{ling2024dl3dv}, following \citet{jiang2025rayzer}. Details of our probing setup are presented in \Cref{sec:app_pose_probe_implementation}.

\begin{table}[ht]
    \centering
    \caption{\textbf{Main Exploration Overview.} We show a full overview over all ablations conducted as a part of \cref{sec:method} with full evaluation results. Novel view synthesis performance is measured on RealEstate-10k~\citep{zhou2018stereo} in the standard pixelSplat~\citep{charatan2024pixelsplat} setting. For camera estimation, we follow RayZer and evaluate (zero-shot) on DL3DV-10k~\citep{ling2024dl3dv}. Neither dataset is a part of the training distribution, thus, all evaluations are zero-shot. We measure camera estimation performance using both probes on camera tokens following RayZer~\cite{jiang2025rayzer} and transferability following X-Factor~\cite{mitchel2025xfactor}. We show ablation results for models trained on both Segment Anything-Video~\citep[SA-V,][]{ravi2024sam2} (high dynamics) in \textbf{(a)} and for models trained on the curated, medium-dynamics dataset SpatialVid-HQ~\citep{wang2025spatialvid} in \textbf{(b)}. We provide additional baselines using the official RayZer~\citep{jiang2025rayzer} and LVSM~\citep{jin2025lvsm} codebases with default hyperparameters (resulting in significantly larger models) trained in the same setting.}
    \adjustbox{max width=\linewidth}{
    \begin{tabular}{l@{\hskip 1em}c@{\hskip .33em}l>{\color{ourgray}}rc@{\hskip .2em}c@{\hskip .2em}c>{\color{ourgray}}c@{\hskip .2em}>{\color{ourgray}}c@{\hskip .2em}>{\color{ourgray}}cc@{\hskip .2em}c@{\hskip .2em}c@{\hskip .2em}c@{\hskip .2em}c@{\hskip .2em}cc@{\hskip .2em}c@{\hskip .2em}c@{\hskip .2em}c@{\hskip .2em}c@{\hskip .2em}c}
        \multicolumn{20}{l}{\textbf{(a)} \textit{Trained on SA-V~\citep{ravi2024sam2} (no camera annotations)}}\\
        \toprule
        \multicolumn{4}{l}{\multirow{2}{*}[-3pt]{Configuration}} & \multicolumn{3}{c}{NVS w/o State} & \multicolumn{3}{c}{\color{ourgray}NVS w/ State} & \multicolumn{6}{c}{Camera Estimation (Probe; $\uparrow$)} & \multicolumn{6}{c}{Camera Estimation (Transfer; $\uparrow$)} \\
        \cmidrule(lr){5-7} \cmidrule(lr){8-10} \cmidrule(lr){11-16} \cmidrule(lr){17-22}
        & & & & \textsc{\smalltableheaderfont PSNR$\uparrow$} & \textsc{\smalltableheaderfont LPIPS$\downarrow$} & \textsc{\smalltableheaderfont SSIM$\uparrow$} & \textsc{\smalltableheaderfont PSNR$\uparrow$} & \textsc{\smalltableheaderfont LPIPS$\downarrow$} & \textsc{\smalltableheaderfont SSIM$\uparrow$} & \textsc{\smalltableheaderfont R@10$^\circ\!\!$} & \textsc{\smalltableheaderfont R@20$^\circ\!\!$} & \textsc{\smalltableheaderfont R@30$^\circ\!\!$} & \textsc{\smalltableheaderfont t@0.1} & \textsc{\smalltableheaderfont t@0.2} & \textsc{\smalltableheaderfont t@0.3} & \textsc{\smalltableheaderfont R@10$^\circ\!\!$} & \textsc{\smalltableheaderfont R@20$^\circ\!\!$} & \textsc{\smalltableheaderfont R@30$^\circ\!\!$} & \textsc{\smalltableheaderfont t@0.1} & \textsc{\smalltableheaderfont t@0.2} & \textsc{\smalltableheaderfont t@0.3} \\
        \midrule
        & & LVSM~\citep{jin2025lvsm} & (cannot train w/o poses) & -- & -- & -- & -- & -- & -- & -- & -- & -- & -- & -- & -- & -- & -- & -- & -- & -- & -- \\
        & & RayZer~\citep{jiang2025rayzer} & (depth $3 \times 8$, width 768) & \multicolumn{3}{c}{diverges consistently} & -- & -- & -- & -- & -- & -- & -- & -- & -- & -- & -- & -- & -- & -- & -- \\
        \midrule
        & \textbf{A} & RayZer-like~\citep{jiang2025rayzer} baseline & (depth $3 \times 6$, width 512)  & \color{ourgray} 22.53\rlap{$^*$} & \color{ourgray} 0.353\rlap{$^*$} & \color{ourgray} 0.708\rlap{$^*$} & -- & -- & -- & \color{ourgray} 39.0\rlap{$^*$} & \color{ourgray} 54.7\rlap{$^*$} & \color{ourgray} 72.8\rlap{$^*$} & \color{ourgray} 15.3\rlap{$^*$} & \color{ourgray} 37.3\rlap{$^*$} & \color{ourgray} 56.5\rlap{$^*$} & \color{ourgray} 59.8\rlap{$^*$} & \color{ourgray} 84.5\rlap{$^*$} & \color{ourgray} 90.1\rlap{$^*$} & \color{ourgray} 06.5\rlap{$^*$} & \color{ourgray} 21.9\rlap{$^*$} & \color{ourgray} 33.9\rlap{$^*$} \\
        \midrule
        \multirow{2}{0.0em}{\rotatebox{90}{{\scriptsize S.~\ref{sec:method_dynamics}}}} & \textbf{B} & \multicolumn{2}{@{}l}{+ Dynamic State Prediction} & 13.42 & 0.632 & 0.509 & 24.01 & 0.347 & 0.746 & 14.2 & 16.0 & 28.3 & 12.1 & 14.7 & 31.5 & 56.1 & 78.1 & 89.9 & 07.0 & 19.2 & 32.7 \\
        & \textbf{C} & \multicolumn{2}{@{}l}{+ Dynamic State Dropout} & 23.01 & 0.324 & 0.720 & 23.76 & 0.337 & 0.719 & 42.5 & 61.0 & 80.1 & 19.0 & 33.9 & 59.2 & 62.4 & 76.5 & 93.5 & 08.1 & 23.9 & 31.4 \\
        \midrule
        \multirow{2}{0.0em}{\rotatebox{90}{{\scriptsize S.~\ref{sec:method_scalability}}}} & \textbf{D} & + Single Network & (depth 18, width 512) & 24.93 & 0.226 & 0.793 & 25.33 & 0.285 & 0.833 & 60.3 & 78.8 & 92.7 & 34.3 & 49.4 & 79.0 & 68.8 & 79.1 & 92.0 & 16.3 & 23.5 & 38.3 \\
        & \textbf{E} & + {Parallel Targets} & ({$-81\%$} FLOPS/novel view) & 24.04 & 0.299 & 0.788 & 25.12 & 0.288 & 0.815 & 60.1 & 77.9 & 93.1 & 34.7 & 50.1 & 78.8 & 70.1 & 84.4 & 88.6 & 15.6 & 27.2 & 38.9 \\
        \midrule
        \multirow{3}{0.0em}{\rotatebox{90}{{\scriptsize \cref{sec:method_improvements}}}} & \textbf{F} & \multicolumn{2}{@{}l}{+ Autoregression over Views} & 23.08 & 0.326 & 0.719 & 24.49 & 0.305& 0.772 & 59.1 & 85.3 & 96.6 & 44.8 & 51.5 & 76.4 & 73.6 & 84.8 & 88.2 & 24.9 & 47.6 & 69.2 \\
        & \textbf{G} & \multicolumn{2}{@{}l}{+ Random-order Autoregression} & 25.45 & 0.237 & 0.817 & 26.28 & 0.217 & 0.871 & 62.6 & 83.2 & 97.9 & 32.7 & 57.3 & 78.2 & 84.4 & 93.4 & 94.8 & 37.2 & 70.4 & 83.1 \\
        & \textbf{H} & \multicolumn{2}{@{}l}{+ Local High-resolution Layers} & 25.61 & 0.226 & 0.823 & 26.87 & 0.209 & 0.867 & 61.9 & 87.8 & 97.2 & 38.0 & 58.8 & 79.0 & 85.0 & 94.5 & 96.8 & 40.2 & 72.4 & 86.3 \\
        \bottomrule
        & \multicolumn{15}{l}{\footnotesize $^*$Results for \textbf{A} are from selected runs that did not diverge.} \\
        \\

        \multicolumn{20}{l}{\textbf{(b)} \textit{Trained on SpatialVid-HQ~\citep{wang2025spatialvid} (mixed low \& high dynamics; includes MegaSaM~\citep{li2025megasam} camera annotations used for supervising LVSM training)}}\\
        \toprule
        \multicolumn{4}{l}{\multirow{2}{*}[-3pt]{Configuration}} & \multicolumn{3}{c}{NVS w/o State} & \multicolumn{3}{c}{\color{ourgray}NVS w/ State} & \multicolumn{6}{c}{Camera Estimation (Probe; $\uparrow$)} & \multicolumn{6}{c}{Camera Estimation (Transfer; $\uparrow$)} \\
        \cmidrule(lr){5-7} \cmidrule(lr){8-10} \cmidrule(lr){11-16} \cmidrule(lr){17-22}
        & & & & \textsc{\smalltableheaderfont PSNR$\uparrow$} & \textsc{\smalltableheaderfont LPIPS$\downarrow$} & \textsc{\smalltableheaderfont SSIM$\uparrow$} & \textsc{\smalltableheaderfont PSNR$\uparrow$} & \textsc{\smalltableheaderfont LPIPS$\downarrow$} & \textsc{\smalltableheaderfont SSIM$\uparrow$} & \textsc{\smalltableheaderfont R@10$^\circ\!\!$} & \textsc{\smalltableheaderfont R@20$^\circ\!\!$} & \textsc{\smalltableheaderfont R@30$^\circ\!\!$} & \textsc{\smalltableheaderfont t@0.1} & \textsc{\smalltableheaderfont t@0.2} & \textsc{\smalltableheaderfont t@0.3} & \textsc{\smalltableheaderfont R@10$^\circ\!\!$} & \textsc{\smalltableheaderfont R@20$^\circ\!\!$} & \textsc{\smalltableheaderfont R@30$^\circ\!\!$} & \textsc{\smalltableheaderfont t@0.1} & \textsc{\smalltableheaderfont t@0.2} & \textsc{\smalltableheaderfont t@0.3} \\
        \midrule
        & & LVSM~\citep{jin2025lvsm} & (depth 24, width 768) & 24.21 & 0.217 & 0.787 & -- & -- & -- & -- & -- & -- & -- & -- & -- & -- & -- & -- & -- & -- & -- \\
        & & RayZer~\citep{jiang2025rayzer} & (depth $3 \times 8$, width 768) & \multicolumn{3}{c}{diverges consistently} & -- & -- & -- & -- & -- & -- & -- & -- & -- & -- & -- & -- & -- & -- & -- \\
        \midrule
        & \textbf{A} & RayZer-like~\citep{jiang2025rayzer} baseline & (depth $3 \times 6$, width 512) & \color{ourgray} 22.69\rlap{$^*$}  & \color{ourgray} 0.362\rlap{$^*$}  & \color{ourgray} 0.711\rlap{$^*$} & -- & -- & -- & \color{ourgray} 42.1\rlap{$^*$} & \color{ourgray} 59.9\rlap{$^*$} & \color{ourgray} 79.4\rlap{$^*$} & \color{ourgray} 16.5\rlap{$^*$} & \color{ourgray} 36.6\rlap{$^*$} & \color{ourgray} 57.2\rlap{$^*$}  & \color{ourgray} 66.0\rlap{$^*$}  & \color{ourgray} 84.3\rlap{$^*$}  & \color{ourgray} 92.8\rlap{$^*$}  & \color{ourgray} 07.7\rlap{$^*$}  & \color{ourgray} 20.5\rlap{$^*$}  & \color{ourgray} 34.1\rlap{$^*$}  \\
        \midrule
        \multirow{2}{0.0em}{\rotatebox{90}{{\scriptsize S.~\ref{sec:method_dynamics}}}} & \textbf{B} & \multicolumn{2}{@{}l}{+ Dynamic State Prediction} & 13.48 & 0.624 & 0.512 & 24.67 & 0.313 & 0.762 & 14.4 & 19.2 & 24.4 & 10.1 & 19.3 & 33.1 & 54.4 & 79.0 & 88.4 & 06.0 & 19.0 & 32.5 \\
        & \textbf{C} & \multicolumn{2}{@{}l}{+ Dynamic State Dropout} & 23.02 & 0.349 & 0.724 & 24.10 & 0.328 & 0.745 & 43.5 & 64.0 & 82.8 & 19.7 & 37.1 & 57.5 & 69.2 & 83.9 & 93.7 & 08.6 & 24.2 & 33.1 \\
        \midrule
        \multirow{2}{0.0em}{\rotatebox{90}{{\scriptsize S.~\ref{sec:method_scalability}}}} & \textbf{D} & + Single Network & (depth 18, width 512) & 26.98 & 0.195 & 0.849 & 27.49 & 0.189 & 0.854 & 66.7 & 87.6 & 96.5 & 32.0 & 56.2 & 71.8 & 74.1 & 86.2 & 92.6 & 19.7 & 22.7 & 39.9 \\
        & \textbf{E} & + {Parallel Targets} & ({$-81\%$} FLOPS/novel view) & 25.91 & 0.225 & 0.824 & 26.21 & 0.213 & 0.828 & 62.8 & 85.9 & 97.0 & 30.1 & 59.5 & 70.7 & 70.9 & 84.7 & 88.3 & 18.2 & 24.4 & 34.7 \\
        \midrule
        \multirow{3}{0.0em}{\rotatebox{90}{{\scriptsize \cref{sec:method_improvements}}}} & \textbf{F} & \multicolumn{2}{@{}l}{+ Autoregression over Views} & 23.53 & 0.301 & 0.752 & 25.78 & 0.267 & 0.790 & 64.6 & 92.9 & 98.1 & 41.2 &  55.7 & 75.5 & 76.5 & 86.1 & 89.4 & 25.2 & 49.6 & 67.1 \\
        & \textbf{G} & \multicolumn{2}{@{}l}{+ Random-order Autoregression} & 27.27 & 0.189 & 0.855 & 29.57 & 0.148 & 0.892 & 67.5 & 90.1 & 97.9 & 39.3 & 62.2 & 77.0  & 86.0 & 94.2 & 97.1 & 39.1 & 71.7 & 82.6 \\
        & \textbf{H} & \multicolumn{2}{@{}l}{+ Local High-resolution Layers} & 27.78 & 0.168 & 0.868 & 30.23 & 0.142 & 0.897 & 68.1 & 90.9 & 97.0 & 38.0 & 62.5 & 77.6 & 88.7 & 96.9 & 98.4 & 42.4 & 76.2 & 87.4 \\
        \bottomrule
        & \multicolumn{15}{l}{\footnotesize $^*$Results for \textbf{A} are from selected runs that did not diverge.}
    \end{tabular}
    }
    \label{tab:main_ablation}
\end{table}
\clearpage

\section{Implementation Details}\label{sec:app_imp_details}
\paragraph{Hyperparameters}
We show relevant hyperparameters for all trained model variations in \Cref{tab:exploration_hparams,tab:main_training_hparams}.

\begin{table}[H]
    \centering
    \caption{\textbf{Main Exploration Hyperparameters.} Details for our models presented in \Cref{sec:method}. Hyperparameters are identical between variants trained on SA-V~\citep{ravi2024sam2} and SV-HQ~\citep{wang2025spatialvid}.}
    \adjustbox{max width=\linewidth}{\scalebox{\ifeccv0.65\else0.7\fi}{
    \begin{tabular}{lccccccccccccccccc}
        \toprule
        Variant & \config{A} & \config{B} & \config{C} & \config{D} & \config{E} & \config{F} & \config{G} & \config{H} \\
        \midrule
        Trainable Parameters & 134M & 134M & 134M & 139M & 139M & 139M & 139M & 145M \\
        Resolution & $256^2$ & $256^2$ & $256^2$ & $256^2$ & $256^2$ & $256^2$ & $256^2$ & $256^2$ \\
        \midrule
        Training Steps & 200k & 200k & 200k & 200k & 200k & 200k & 200k & 200k \\
        Batch Size & 256 & 256 & 256 & 256 & 256 & 256 & 256 & 256 \\
        Precision & bf16 MP & bf16 MP & bf16 MP & bf16 MP & bf16 MP & bf16 MP & bf16 MP & bf16 MP \\
        Training Hardware & 32 H200 & 32 H200 & 32 H200 & 32 H200 & 32 H200 & 32 H200 & 32 H200 & 32 H200 \\
        \midrule
        Width & 512 & 512 & 512 & 512 & 512 & 512 & 512 & 512 \\
        Depth ([$\mathcal{E}_\text{cam}$, $\mathcal{E}_\text{scene}$, $\mathcal{D}_\text{render}$] or $\mathcal{M}$) & [6, 6, 6] & [6, 6, 6] & [6, 6, 6] & 18 & 18 & 18 & 18 & 18 \\
        Local Layers & -- & -- & -- & -- & -- & -- & -- & $[2, 2] \cdot 2$ \\
        Local Layer Width & -- & -- & -- & -- & -- & -- & -- & $[128, 256] \cdot 2$ \\
        Attention Head Dim & 64 & 64 & 64 & 64 & 64 & 64 & 64 & 64 \\
        Neighborhood~\cite{hassani2023neighborhood} Kernel Size & -- & -- & -- & -- & -- & -- & -- & $7^2$ \\
        Patch Size & $16^2$ & $16^2$ & $16^2$ & $16^2$ & $16^2$ & $16^2$ & $16^2$ & $4^2$ \\
        Positional Encoding & RoPE~\cite{su2021roformer} & RoPE~\cite{su2021roformer} & RoPE~\cite{su2021roformer} & RoPE~\cite{su2021roformer} & RoPE~\cite{su2021roformer} & RoPE~\cite{su2021roformer} & RoPE~\cite{su2021roformer} & RoPE~\cite{su2021roformer} \\
        \midrule
        Dynamic State Dim & -- & 256 & 256 & 256 & 256 & 256 & 256 & 256 \\
        Dynamic State Dropout Rate & -- & 0 & 0.5 & 0.5 & 0.5 & 0.5 & 0.5 & 0.5 \\
        \midrule
        Train Dataset & SA-V/SV-HQ & SA-V/SV-HQ & SA-V/SV-HQ & SA-V/SV-HQ & SA-V/SV-HQ & SA-V/SV-HQ & SA-V/SV-HQ & SA-V/SV-HQ \\
        Avg.\ Frame Extraction Rate & 2fps & 2fps & 2fps & 2fps & 2fps & 2fps & 2fps & 2fps \\
        Input Views & 6 & 6 & 6 & 6 & 6 & 1..7 & 1..7 & 1..7 \\
        Output Views & 2 & 2 & 2 & 2 & 2 & 7 & 7 & 7 \\
        Frame Order & -- & -- & -- & -- & -- & ordered & random & random \\
        $\lambda_{\mathrm{perc}}$ & 0 & 0 & 0 & 0 & 0 & 0 & 0 & 0 \\
        \midrule
        Optimizer & AdamW~\cite{loshchilov2018decoupled} & AdamW~\cite{loshchilov2018decoupled} & AdamW~\cite{loshchilov2018decoupled} & AdamW~\cite{loshchilov2018decoupled} & AdamW~\cite{loshchilov2018decoupled} & AdamW~\cite{loshchilov2018decoupled} & AdamW~\cite{loshchilov2018decoupled} & AdamW~\cite{loshchilov2018decoupled} \\
        Learning Rate & $10^{-4}$ & $10^{-4}$ & $10^{-4}$ & $10^{-4}$ & $10^{-4}$ & $10^{-4}$ & $10^{-4}$ & $10^{-4}$ \\
        Learning Rate Warmup & 1k & 1k & 1k & 1k & 1k & 1k & 1k & 1k \\
        Learning Rate Schedule & constant & constant & constant & constant & constant & constant & constant & constant \\
        Betas $(\beta_1, \beta_2)$ & $(0.9, 0.95)$ & $(0.9, 0.95)$ & $(0.9, 0.95)$ & $(0.9, 0.95)$ & $(0.9, 0.95)$ & $(0.9, 0.95)$ & $(0.9, 0.95)$ & $(0.9, 0.95)$ \\
        Weight Decay & 0.01 & 0.01 & 0.01 & 0.01 & 0.01 & 0.01 & 0.01 & 0.01 \\
        \bottomrule
    \end{tabular}
    }}
    \label{tab:exploration_hparams}
\end{table}

\begin{table}[H]
    \centering
    \caption{\textbf{Main Training Hyperparameters.} Details for our models presented in \cref{sec:experiments}. \textsc{Scaling}-\textbf{S} is derived from \config{H} and has identical hyperparameters except for the training dataset. The ``RayZer setting'' model has the same depth, width, and overall view count as the models by \citet{jiang2025rayzer} and is only trained on DL3DV~\citep{ling2024dl3dv} to match their setting.}
    \adjustbox{max width=\linewidth}{\scalebox{\ifeccv0.65\else0.7\fi}{
    \begin{tabular}{lccccccHHH}
        & \multicolumn{4}{c}{\textit{Scaling}} & \textit{RayZer Setting (\cref{tab:prior_dl3dv_trained})} & \multicolumn{1}{c}{\textit{Final Models}} \\
        \cmidrule(lr){2-5} \cmidrule(lr){6-6} \cmidrule(lr){7-7}
        \toprule
        Variant & \textsc{Scaling}-\textbf{XS} & \textsc{Scaling}-\textbf{S} & \textsc{Scaling}-\textbf{B} & \textsc{Scaling}-\textbf{L} & \textsc{RayDer}-\textbf{B}-DL3DV & \textsc{RayDer}-\textbf{L}-$576^2$ \\
        \midrule
        Trainable Parameters & 59M & 145M & 422M & 743M & 422M & 743M \\
        Resolution & $256^2$ & $256^2$ & $256^2$ & $256^2$ & $256^2$ & $576^2$ \\
        \midrule
        Training Steps & various & various & various & various & 50k & $\underbracket[0.187ex]{500\text{k}}_\text{\textsc{Scaling}-\textbf{L}}\!\!\! + \!\underbracket[0.187ex]{100\text{k}}_{576^{\smash{2}}\text{ tune}}\! + \underbracket[0.187ex]{50\text{k}}_{\text{decay}}$ \\
        Batch Size & 256 & 256 & 256 & 256 & 256 & 256 & 256 & 256 & 256 \\
        Precision & bf16 MP & bf16 MP & bf16 MP & bf16 MP & bf16 MP & bf16 MP & bf16 MP & bf16 MP & bf16 MP \\
        Training Hardware & 32 (G)H200 & 32 (G)H200 & 64 (G)H200 & 128 (G)H200 & 64 (G)H200 & 128 (G)H200 \\
        Step Time (incl.\ comm.) & 0.23s$^\ast$ & 0.31s$^\ast$ & 0.34s$^\ast$ & 0.32s$^\ast$ & 1.22s$^\ast$ & 1.98s$^\ast$ \\
        TFLOP/step (BS 1) & 5.93 & 11.56 & 27.79 & 48.37 & 101.79 & 304.08 \\
        \midrule
        Width & 384 & 512 & 768 & 1024 & 768 & 1024 \\
        Depth & 12 & 18 & 24 & 24 & 24 & 24 \\
        Local Layers & $[2, 2] \cdot 2$ & $[2, 2] \cdot 2$ & $[2, 2] \cdot 2$ & $[2, 2] \cdot 2$ & $[2, 2] \cdot 2$ & $[2, 2] \cdot 2$ & $[2, 2] \cdot 2$ & $[2, 2] \cdot 2$ & $[2, 2] \cdot 2$ \\
        Local Layer Width & $[128, 256] \cdot 2$ & $[128, 256] \cdot 2$ & $[128, 256] \cdot 2$ & $[128, 256] \cdot 2$ & $[128, 256] \cdot 2$ & $[128, 256] \cdot 2$ & $[128, 256] \cdot 2$ & $[128, 256] \cdot 2$ & $[128, 256] \cdot 2$ \\
        Attention Head Dim & 64 & 64 & 64 & 64 & 64 & 64 & 64 & 64 & 64 \\
        Neighborhood~\cite{hassani2023neighborhood} Kernel Size & $7^2$ & $7^2$ & $7^2$ & $7^2$ & $7^2$ & $7^2$ & $7^2$ & $7^2$ & $7^2$ \\
        Patch Size & $4^2$ & $4^2$ & $4^2$ & $4^2$ & $4^2$ & $4^2$ & $4^2$ & $4^2$ & $4^2$ \\
        Positional Encoding & RoPE~\cite{su2021roformer} & RoPE~\cite{su2021roformer} & RoPE~\cite{su2021roformer} & RoPE~\cite{su2021roformer} & RoPE~\cite{su2021roformer} & RoPE~\cite{su2021roformer} & RoPE~\cite{su2021roformer} & RoPE~\cite{su2021roformer} & RoPE~\cite{su2021roformer} \\
        \midrule
        Dynamic State Dim & 256 & 256 & 256 & 256 & 256 & 256 & 256 & 256 & 256 \\
        Dynamic State Dropout Rate & 0.5 & 0.5 & 0.5 & 0.5 & 0.5 & 0.5 & 0.5 & 0.5 & 0.5 \\
        \midrule
        Train Dataset & SV~\cite{wang2025spatialvid} & SV~\cite{wang2025spatialvid} & SV~\cite{wang2025spatialvid} & SV~\cite{wang2025spatialvid} & DL3DV-10K~\cite{ling2024dl3dv} & SV~\cite{wang2025spatialvid} \\
        Avg.\ Frame Extraction Rate & 2fps & 2fps & 2fps & 2fps & 1.5fps & 2fps & 2fps & 2fps & 2fps \\
        Input Views & 1..7 & 1..7 & 1..7 & 1..7 & 1..23 & 1..7 \\
        Output Views & 7 & 7 & 7 & 7 & 23 & 7 \\
        Frame Order & random & random & random & random & random & random \\
        $\lambda_{\mathrm{perc}}$ & 0 & 0 & 0 & 0 & 0.2 & 0 \\
        \midrule
        Optimizer & AdamW~\cite{loshchilov2018decoupled} & AdamW~\cite{loshchilov2018decoupled} & AdamW~\cite{loshchilov2018decoupled} & AdamW~\cite{loshchilov2018decoupled} & AdamW~\cite{loshchilov2018decoupled} & AdamW~\cite{loshchilov2018decoupled} & AdamW~\cite{loshchilov2018decoupled} & AdamW~\cite{loshchilov2018decoupled} \\
        Weight Decay & 0.01 & 0.01 & 0.01 & 0.01 & 0.01 & 0.01 & 0.01 & 0.01 \\
        Betas $(\beta_1, \beta_2)$ & $(0.9, 0.95)$ & $(0.9, 0.95)$ & $(0.9, 0.95)$ & $(0.9, 0.95)$ & $(0.9, 0.95)$ & $(0.9, 0.95)$ & $(0.9, 0.95)$ & $(0.9, 0.95)$ \\
        Learning Rate & $10^{-4}$ & $10^{-4}$ & $10^{-4}$ & $10^{-4}$ & $10^{-4}$ & $10^{-4}$ & $10^{-4}$ & $10^{-4}$ \\
        Learning Rate Warmup & 1k & 1k & 1k & 1k & 1k & 1k & 1k & 1k \\
        Learning Rate Schedule & constant & constant & constant & constant & cosine decay & WSD~\citep{hu2024minicpm} \\
        \bottomrule
        \multicolumn{7}{c}{\small \textit{$^\ast$training speed measured on 4$\times$ Nvidia H200 nodes with GPUs power-limited to 500W and NDR200 interconnect; other H200 setups can be faster}}
    \end{tabular}
    }}
    \label{tab:main_training_hparams}
\end{table}

\subsection{Architecture Details}

\paragraph{Transformer Block Setup}
We adopt the general Llama 2-style~\cite{touvron2023llama2openfoundation} transformer block setup from HDiT~\cite{crowson2024hourglass}, but incorporate a VGGT-style~\citep{wang2025vggt} intra-frame and global attention factorization (inspired by Open-RayZer~\cite{wang2025OpenRayzer}).

\paragraph{Choice of Canonical View}
Unlike RayZer~\cite{jiang2025rayzer}, we do not rely on a canonical view.
Instead, camera poses are predicted pointwise -- i.e., not with a relative MLP head that predicts based on two camera tokens, but a head that directly predicts an absolute pose from a single token, similar to $\pi^3$~\cite{wang2025pi3}.
We found during early explorations that this leads to slightly more stable training behavior without any evident drawbacks.
The same setup has independently been adopted by Open-RayZer~\cite{wang2025OpenRayzer}.

\paragraph{Output Heads}
We simplify RayZer's multi-layer MLP output heads to an RMSNorm~\cite{zhang2019root} followed by a single linear layer.
Separate heads are used for camera pose prediction, intrinsics prediction, and dynamic state prediction.
We observed no degradation in performance from this simplification in early explorations.

\paragraph{Attention between Views}
Inspired by Open-RayZer~\cite{wang2025OpenRayzer}, we adopt a VGGT-style~\cite{wang2025vggt} attention setup where each transformer layer has two attention layers -- one for \emph{intra-view} attention, one for \emph{global} attention.
We use axial RoPE~\cite{su2021roformer,crowson2024hourglass} for intra-view attention and no positional encoding for global attention.
Our final attention masking across views during novel view synthesis is defined as follows:
Let $\{t_{i,j}\}_j$ be the set of tokens corresponding to view $I_i$, which, in turn is either an input view ($I_i \in \mathcal{I}_\mathrm{input}$, with $\mathcal{I}_\mathrm{input}$ being an \emph{ordered} set) or a target view ($I_i \in \mathcal{I}_\mathrm{target}$).
For a token $t_{i,j}$, whether it can attend to another token $t_{i',k}$ is decided by the following rule \emph{at inference time}:
\begin{equation}
    \begin{cases}
        \mathrm{yes,}\ifeccv\else\quad\fi i = i' &\quad{\color{ourgrayborder}\triangleright\ \text{same view}} \\
        \mathrm{yes,}\ifeccv\else\quad\fi i' < i \land (I_i \in \mathcal{I}_\mathrm{input} \land I_{i'} \in \mathcal{I}_\mathrm{input}) &\quad{\color{ourgrayborder}\triangleright\ \text{both input view \& causal}} \\
        \mathrm{yes,}\ifeccv\else\quad\fi I_i \in \mathcal{I}_\mathrm{target} \land I_{i'} \in \mathcal{I}_\mathrm{input} &\quad{\color{ourgrayborder}\triangleright\ \text{target view attends to all inputs}} \\
        \mathrm{\rlap{no,}\phantom{yes,}}\ifeccv\else\quad\fi \text{otherwise}. \\
    \end{cases}
\end{equation}
At \emph{train time}, $\mathcal{I}_\mathrm{input}$ and $\mathcal{I}_\mathrm{target}$ have significant overlap.
Specifically, for the (randomly) ordered set of all views $\mathcal{I} = \{I_1, I_2, \ldots, I_K\}$, we use $\mathcal{I}_\mathrm{input} = \mathcal{I} \setminus \{I_K\}, \mathcal{I}_\mathrm{target} = \mathcal{I} \setminus \{I_1\}$.
Importantly, views that are both input and target views will be present as tokens \emph{twice} in the sequence.
Here, we will abuse notation somewhat for simplicity: when comparing view indices $i, i'$, we will be referring to the set of all views $\mathcal{I}$, regardless of whether the tokens belong to an input or target view; when comparing set containment (e.g., $I_i \in \mathcal{I}_\mathrm{input}$), the indices are ``role-aware''.
For a token $t_{i,j}$, whether it can attend to another token $t_{i',k}$ is then decided by the following \emph{train-time} rule (differences to inference-time \emph{emphasized}):

\ifeccv
    \noindent \adjustbox{max width=\linewidth}{
    \begin{equation}
        \begin{cases}
            \mathrm{yes,} i = i' \land ((I_i \in \mathcal{I}_\mathrm{input}) = (I_{i'} \in \mathcal{I}_\mathrm{input})) &\quad{\color{ourgrayborder}\triangleright\ \text{same view, \emph{same role}}} \\
            \mathrm{yes,} i' < i \land (I_i \in \mathcal{I}_\mathrm{input} \land I_{i'} \in \mathcal{I}_\mathrm{input}) &\quad{\color{ourgrayborder}\triangleright\ \text{both input view \& causal}} \\
            \mathrm{yes,} I_i \in \mathcal{I}_\mathrm{target} \land I_{i'} \in \mathcal{I}_\mathrm{input} \land i' < i &\quad{\color{ourgrayborder}\triangleright\ \text{target view attends to all inputs \emph{before it}}} \\
            \mathrm{\rlap{no,}\phantom{yes,}} \text{otherwise}. \\
        \end{cases}
    \end{equation}
    }
\else
    \begin{equation}
        \begin{cases}
            \mathrm{yes}, \quad i = i' \land ((I_i \in \mathcal{I}_\mathrm{input}) = (I_{i'} \in \mathcal{I}_\mathrm{input})) &\quad{\color{ourgrayborder}\triangleright\ \text{same view, \emph{same role}}} \\
            \mathrm{yes}, \quad i' < i \land (I_i \in \mathcal{I}_\mathrm{input} \land I_{i'} \in \mathcal{I}_\mathrm{input}) &\quad{\color{ourgrayborder}\triangleright\ \text{both input view \& causal}} \\
            \mathrm{yes}, \quad I_i \in \mathcal{I}_\mathrm{target} \land I_{i'} \in \mathcal{I}_\mathrm{input} \land i' < i &\quad{\color{ourgrayborder}\triangleright\ \text{target view attends to all inputs \emph{before it}}} \\
            \mathrm{\rlap{no,}\phantom{yes,}} \quad \text{otherwise}. \\
        \end{cases}
    \end{equation}
\fi
This ensures that training is fully autoregressive/causal, with target views completely independent during both training and inference.
During training, some target views see only a subset of input views ot obtain a better training signal (\cref{sec:method_improvements}); during inference time, every target view sees all input views to maximize NVS quality.

\paragraph{Local High-resolution Layers}
We follow HDiT~\cite{crowson2024hourglass} and add low-width, shallow transformer blocks with neighborhood attention~\cite{hassani2023neighborhood} to the outside of the main transformer, with skips around the backbone.
In our setup, neighborhood attention is performed \emph{exclusively} intra-frame, as neighborhoods are not well-defined for general multi-view setups.
During camera estimation, only the encoder side, alongside the main block, is used, whereas all blocks are utilized during novel view synthesis.
In preliminary explorations, we found that scaling these additional layers alongside the main block is not necessary -- neither in width nor in depth.
This is consistent with \citet{crowson2024hourglass} consistently using two layers per resolution in the local layers across all configurations.

\paragraph{Conditioning on Token Roles}
Following \citet{nair2025scaling}, we condition the transformer on the role of each token. We extend their setup from differentiating between input and output view tokens to differentiating between roles across two axes:
\begin{equation*}
    \{\text{Camera Estimation}, \text{NVS}\} \times \{\text{View Token}, \text{Camera Token}, \text{State Token}\}.
\end{equation*}
Conditioning is done via RMSNorms~\cite{zhang2019root} with adaptive~\citet{huang2017arbitrary} scale on the input of each block; we do not use post-modulation.
This adds a substantial number of trainable parameters to the backbone, but only results in negligible increases in computational cost.

\paragraph{Training Supervision}
Following RayZer~\citep{jiang2025rayzer}, we train \methodname end-to-end with a pixel-space reconstruction loss.
Given a set of input views
\begin{equation}
\mathcal{I} = \{ I_i \in \mathbb{R}^{H \times W \times 3} \mid i = 1, \dots, K \},
\end{equation}
we randomly partition \(\mathcal{I}\) into two disjoint subsets\footnote{In practice, during training in our final setup, we do \emph{not} have a strict separation between context/input and target sets. However, with respect to each \emph{individual} target view, such sets can implicitly be constructed, to accurately describe the overall training behavior. We will use this implicit notation setup for this paragraph to be more consistent with \citet{jiang2025rayzer}.}, a context set \(\mathcal{I}_{\mathcal A}\) and a target set \(\mathcal{I}_{\mathcal B}\), such that
\begin{equation}
\mathcal{I}_{\mathcal A} \cup \mathcal{I}_{\mathcal B} = \mathcal{I},
\qquad
\mathcal{I}_{\mathcal A} \cap \mathcal{I}_{\mathcal B} = \emptyset.
\end{equation}
Conditioned on the context set \(\mathcal{I}_{\mathcal A}\), the model predicts the corresponding held-out target views
\begin{equation}
\hat{\mathcal{I}}_{\mathcal B} = \{ \hat{I}_j \mid I_j \in \mathcal{I}_{\mathcal B} \}
\end{equation}
The full training objective is then given by loss over the target set $\mathcal{I}_{\mathcal B}$ and the corresponding predictions $\hat{\mathcal{I}}_{\mathcal B}$
\begin{equation}
\mathcal{L} = \frac{1}{|\mathcal{I}_{\mathcal B}|} \sum_{I_j \in \mathcal{I}_{\mathcal B}} \left[ \mathrm{MSE}(I_j, \hat{I}_j) + \lambda_{\mathrm{perc}} \, \mathrm{Percep}(I_j, \hat{I}_j) \right].
\label{eq:trainloss}
\end{equation}
Here, $\mathrm{MSE}(\cdot,\cdot)$ denotes the pixel-wise mean squared error, $\mathrm{Percep}(\cdot,\cdot)$ denotes the optional perceptual loss \cite{zhang2018unreasonable}, and $\lambda_{\mathrm{perc}} \geq 0$ is the corresponding weighting factor. The partition \((\mathcal{I}_{\mathcal A}, \mathcal{I}_{\mathcal B})\) is randomly resampled during training.

\paragraph{Ray Encoding}
We follow RayZer and use Plücker~\citep{plucker1865xvii} ray maps that we concatenate to the input alongside RGB pixels (if provided).

\paragraph{Learning Rate Scheduling}
Unlike RayZer~\cite{jiang2025rayzer}, which uses a cosine decay schedule, we follow the Warmup Stable Decay (WSD) schedule proposed by \citet{hu2024minicpm}.
Notably, this schedule, which consists of three stages -- a linear warmup, a constant stage, and exponential decay -- does not need to commit to a fixed training length ahead of time, which is crucial for our scaling experiments.
For those, we omit the decay stage, as this enables very significant compute savings, and as we found that decaying leads to similar gains on our benchmarks across similar training horizons -- i.e., according to our early explorations, comparisons for our models for the purposes of our main exploration and scaling experiments are fair even without decay.

For the exponential decay stage, we experimented with halving the learning rate every \{1k, 5k, 10k\} steps.
We found that 5k and 10k performed similarly, while 1k performed significantly worse, and chose 10k to err on the safe side.

\paragraph{Extrinsics Parametrization}
We parameterize the camera extrinsics as a 6D twist $\xi=(\omega,v)\in\mathbb{R}^6$ and map it to a rigid transform $(R,t)\in SE(3)$ via the exponential map: $R=\exp(\hat{\omega})$ using Rodrigues' formula (with small-angle Taylor fallbacks for numerical stability), and $t=J(\omega)\,v$ where $J(\omega)$ is the $SO(3)$ left Jacobian (also using a Taylor series near $\|\omega\|\approx 0$ for improved stability). This yields an unconstrained, fully differentiable parameterization with a minimal number of parameters that guarantees valid rotations, unlike the $SO(3) \times \mathbb{R}^3$ parametrization~\cite{zhou2019continuity} used by RayZer~\cite{jiang2025rayzer}, which has singularities. In experiments using their choice of extrinsics parametrization, we have \emph{not} observed any instabilities that were directly attributable to the choice of parametrization. We adopted our choice of parametrization since it is more compact and less \emph{likely} to exhibit instabilities \emph{in this use case}, not out of necessity. Importantly, this does \emph{not} apply to the $SO(3)$ \emph{regression} setting that the parametrization by \citet{zhou2019continuity} was originally developed for: there, the singularities are not a major concern, and instead, the periodicity of our 6D twist parametrization choice would become problematic (ambiguous targets). Since the poses are passed to a renderer, which itself is perfectly invariant to these ambiguities (all periodic values map to the same $SE(3)$ pose and thus to the same Plücker ray maps).

\paragraph{Intrinsics Parametrization}
For pinhole camera models, we parametrize the focal length $f$ as
\begin{equation}
    f_x = f_y = f = \exp(\theta_f) + \epsilon_f,
\end{equation}
where $\theta_f$ is a parameter predicted by the model, and $\epsilon_f = 10^{-6}$ ensures that the focal length can not go to 0. The focal length is defined with respect to a normalized camera coordinate system $(u, v) \inapprox [-1, 1]^2$, where we scale by the image's height to ensure it always falls in $[-1, 1]$, scaling the coordinate system consistently. We enable a learnable bias for the prediction of $\theta_f$ and initialize both it and the weight matrix of the layer predicting it to zeros, resulting in initial predicted focal lengths $f = 1$, close to the typical mean value of approximately $2$ that we observe the model learns to predict on our training data.

Unlike RayZer, we predict per-view intrinsics as we noticed during inspection that a fraction of the training data includes zooming over the course of the video. However, this did not have any noticeable impact on training stability, and typical NVS benchmarks do not include such variations.

In some cases, we observe divergence of the model's predicted intrinsics during training, where the predicted focal length becomes either approximately zero ($< 10^{-5}$) or very large ($\gg 10^{10}$). Specifically, we observe these divergences to be more likely to happen when training at low global batch sizes (e.g., 16). Adding a sufficiently long learning rate warmup period seems to address this problem. We specifically find a linear warmup from 0 to the peak learning rate over 1000 steps to suffice for preventing this divergence in our test cases. Notably, while this reduces NVS quality somewhat, this does not cause the model to collapse. Intuitively, we attribute this to NVS still being relatively well-defined even without intrinsics prediction in the majority of cases, since intrinsics can be approximately inferred from the reference views provided during rendering, as they are often consistent across views. RayZer~\cite{jiang2025rayzer} similarly uses a warmup, albeit longer at 3k steps.

\subsection{Further Details}

\subsubsection{Scaling Power Laws}\label{sec:app_scaling_power_laws}
We also explore fitting power laws to capture our model's scaling behavior, fitting on \emph{eval} metrics on unseen datasets (typically RE10K~\cite{zhou2018stereo} after training on SpatialVid~\cite{wang2025spatialvid}).
We generally first determine the pareto frontier (per training dataset size $D$) of the target metric over training compute $C$ (quantified in GFLOP, with data points starting at 50k steps of training), and then fit the target function to the pareto frontier.

\paragraph{What Metric to Fit}
NVS performance is typically quantified using PSNR, LPIPS, and SSIM.
Power laws are typically fit on metrics where \textit{lower = better}, and which are not already log-scaled.
We therefore fit the scaling laws not necessarily on the target metrics directly, but on:
\begin{itemize}\ifeccv[leftmargin=*]\fi
    \item PSNR $\rightarrow$ MSE
    \item LPIPS $\rightarrow$ LPIPS
    \item SSIM $\rightarrow$ 1 $-$ SSIM
\end{itemize}
We find that fitting standard power laws to these generally works well for our models.
When visualizing fitted functions, we transform the predicted values back to the original metric.

\paragraph{What Function to Fit}
When fitting \emph{for one specific amount of data} over compute $C$, we find the following standard power law to consistently lead to good fits:
\begin{equation}
    L(C) = L_\infty + AC^{-\alpha},
\end{equation}
where $L$ is the target metric, $L_\infty$ is the irreducible part~\citep{henighan2020scaling}, and $A, \alpha$ are the coefficients.
We also explored fitting $L = L_\infty + A(C + C_0)^{-\alpha}$, but found this to be unnecessary, with $C_0 \approx 0$ consistently across all three metrics.
This is valuable, 

When fitting one shared function also \emph{across dataset size $D$}, we found the following formulation inspired by Eq.\ 1.5 by \citet{kaplan2020scaling} leads to good fits:
\begin{equation}
    L(C, D) = L_\infty + \bigl(AC^{-\alpha} + BD^{-\beta}\bigr)^\gamma.
\end{equation}
As in the setting where we only fit to a single training dataset scale, we also evaluate a more complex version
\begin{equation*}
    L(C, D) = L_\infty + \bigl(A(C + C_0)^{-\alpha} + B(D + D_0)^{-\beta}\bigr)^\gamma,
\end{equation*}
but find that the constants $C_0, D_0$ are unnecessary to achieve a good fit (and tend toward zero even when optimized), so we omit them.

Fitting this power law to our compute-optimal Pareto frontier of models we trained across model and dataset scale (see \cref{fig:compute_optimal_scaling}, left), we get (rounded to two significant digits):
\begin{align}
    \mathrm{MSE}(C, D) &\approx 0.0033 + \left(200\cdot C^{-0.40} + 2.6\cdot D^{-0.60} \right)^{2.82} &&{\color{ourgrayborder}\triangleright\ R^2 = 0.997}\\
    \mathrm{LPIPS}(C, D) &\approx 0.11 + \left(7000\cdot C^{-0.43} + 14\cdot D^{-0.58} \right)^{1.82} &&{\color{ourgrayborder}\triangleright\ R^2 = 0.997}\\
    1 - \mathrm{SSIM}(C, D) &\approx 0.076 + \left(700\cdot C^{-0.35} + 10\cdot D^{-0.47} \right)^{3.34} &&{\color{ourgrayborder}\triangleright\ R^2 = 0.997}
\end{align}

We also explored the following other options:
\begin{align}
    \label{eq:app_scaling_joint_simple} L(C, D) &= L_\infty + A(C + C_0)^{-\alpha} + B(D + D_0)^{-\beta},\\
    L(C, D) &= L_\infty + A(C + C_0)^{-\alpha} + B(D + D_0)^{-\beta} + \underbrace{E(C + C_0)^{-\alpha_2}(D + D_0)^{-\beta_2}}_\text{multiplicative cross term},\\
    L(C, D) &= L_\infty + B(D + D_0)^{-\beta} + \underbrace{A(C + C_0)^{-\alpha}(D + D_0)^{-\delta}}_\text{dataset-modulated compute term},
\end{align}
which led to bad fits (except \Cref{eq:app_scaling_joint_simple} for LPIPS specifically, while still getting a bad fit for the other metrics), and
\begin{align}
    L(C, D) = L_\infty + BD^{-\beta} + \underbrace{AC^{-\alpha}\left(\frac{D}{D+D_1}\right)^\kappa}_\text{dataset-gated compute term},
\end{align}
which led to decent but less optimal fits.
It also predicted decreases in performance with additional data for very low-compute training, which we were unable to reproduce.

\paragraph{Scaling-Fit Robustness across Benchmarks}\label{sec:app_scaling_fit_robustness}
To verify that the compute-data power law of \cref{eq:scaling_law} captures a property of the model and data rather than a benchmark-specific fitting artifact, we refit it -- with identical functional form -- on additional, deliberately harder and noisier zero-shot evaluation sets beyond RE10K.
\Cref{tab:extra_scaling_law_fits} reports the resulting goodness of fit.
The fit remains accurate across all benchmarks and metrics, supporting that the observed scaling trend is not specific to a single test distribution.

\begin{table}[ht]
    \centering
    \caption{\textbf{Scaling Law Fit on Additional Datasets.} Despite being significantly smaller compard to RE10K, goodness-of-fit is also high for other eval sets. Split abbreviations refer to the ones from \cref{tab:open_set_nvs}.}
    \adjustbox{max width=\linewidth}{\scalebox{.85}{
    \begin{tabular}{lcccc}
        \toprule
        Eval.~Dataset & Split & $R^2$ (PSNR) & $R^2$ (LPIPS) & $R^2$ (SSIM) \\
        \midrule
        RE10K~\citep{zhou2018stereo} & pixelSplat, 2-view & 0.997 & 0.997 & 0.997 \\
        LLFF~\citep{mildenhall2019llff} & R, 3-view & 0.985 & 0.986 & 0.988 \\
        Co3Dv2~\citep{reizenstein21co3d} & R, 3-view & 0.971 & 0.970 & 0.962 \\
        WildRGBD~\citep{xia2024rgbd} & S\textsubscript{e}, 3-view & 0.993 & 0.992 & 0.994 \\
        WildRGBD~\citep{xia2024rgbd} & S\textsubscript{h}, 3-view & 0.948 & 0.993 & 0.971 \\
        \bottomrule
    \end{tabular}
    }}
    \label{tab:extra_scaling_law_fits}
\end{table}

\subsubsection{Training Data Details}
\paragraph{Datasets}
We directly use the original videos of SpatialVid~\citep{wang2025spatialvid} and its HQ subset, and SA-V~\citep{ravi2024sam2}, with the only further preprocessing being (randomized) sharding and the preprocessing detailed the following preprocessing paragraph.
We specifically chose SA-V due to it being deliberately recorded (with a focus on diversity of content), undergoing a review process, and having a clearly defined license. This makes it a good candidate for explorations on truly open-set data that is also likely to enable direct fair comparisons in the future.

For the 1\% and 10\% subsets of SpatialVid in our data scaling analysis, we define a specific randomly chosen subset of our training shards that is identical across all runs.
The ratio of HQ shards vs.\ non-HQ shards reflects that of the full SpatialVid dataset.
The 1\% subset is chosen such that it is a subset of the 10\% subset (which in turn is a subset of the 100\% subset).

\paragraph{Preprocessing}
Before training, we unify codecs and slightly reduce the frame rate, converting to high-bitrate H.264 at 6 fps.
While minimally reducing the amount of available training data variation due to the reduced frame rate, we found this to be crucial to enable efficient training without being bottlenecked by data loading, as the common approach of extracting the whole training dataset into individual frame images to enable fast data loading chosen by many previous NVS methods is not tractable at the data quantities explored in this work.

\paragraph{Video Frame Sampling}
During training, we sample frames with an average fps of 2, for which we randomly select chunks from source videos.
To increase data variation, we randomly perturb the exact sampling times, choosing a random time uniformly in a local range.
Let the frame interval be denoted as $\Delta t$ and the uniform location of the $i$-th frame be denoted as $t_i$,
Then, the perturbed frame location is drawn from $t_i' \sim \mathcal{U}(t_i - \frac{1}{2}\Delta t, t_i + \frac{1}{2}\Delta t)$.
If a video snippet is too short even for our 2fps 8 frame setting (i.e., shorter than 4s), we discard it during training.

\subsubsection{Pose Probe}\label{sec:app_pose_probe_implementation}
Similar to RayZer \cite{wang2025OpenRayzer} and XFactor \cite{mitchel2025xfactor}, we train a probe from froze camera-estimator features to poses, in order to assess the quality of the predicted camera poses.
We take a frame-distance of 1 and 24 frames (i.e. 24 consecutive frames from DL3dV10k). We choose 24 frames since RayZer was trained on this number of views, and distance 1 to ensure that the pose-estimation does not fail due to a too large baseline. For evaluation we use the middle frame to align the trajectories. We follow the evaluation protocol from XFactor, but instead of fitting a 3-layer MLP, we use a 2-layer MLP with hidden dimension 128 in order not to saturate the metrics too quickly as in XFactor. Similarly, we train the probe for 10000 iterations with AdamW with a learning rate of $1e-4$.  We align the poses at the midpoint, i.e. $t_{i} = t_i - t_{mid}$ and $R_{i} = R_iR_{mid}^T$. We normalize the camera poses to range $[-1,1]$:  $t_{i} = t_i / ( \max(\|t_i\|) +\varepsilon)$ In order to measure performance, we follow XFactor and RayZer, where we measure rotation- and translation-accuracy t@$\alpha$, R@$\alpha$ where $\alpha \in \{ 10,20,30\}$ degrees.
Per frame $i$, we are learning a mapping which learns the relative transforms to obtain the GT trajectory. For the given camera-to-world transform $[R_i|t_i], R_i \in SO(3)$ we define the relative transform $R_{ij} = R_jR_i^\top, t_{ij} = t_i - t_j$. We learn $f_\theta: f_i \mapsto \xi_i = (\omega_i, v_i)\in \mathbb{R}^6$ for camera-estimator features $f_i$. We map these to $SE(3)$ using Rodrigues' formula. For the training we use a geodesic loss.

\subsection{Other Things we Tried}
    In this section, we briefly discuss some things we tried but ultimately did not include in the main paper.
    These explorations were mostly performed in early stages of the project before the final version of the model was fixed, and are included in hopes of being useful to other people considering exploring related aspects in the future.

    \paragraph{Pose Scene Normalisation}
    We briefly explored normalizing the scene such that camera poses can not collapse to singular points or expand significantly.
    However, we observed no significant gains from this in our setup, with already unstable configurations collapsing anyway and stable configurations generally having non-divergent poses.

    \paragraph{Alternative Intrinsics Parametrizations}
    We explored various parametrizations for intrinsics, including:
    \begin{itemize}\ifeccv[leftmargin=*]\fi
        \item Pinhole with $f_x=f_y=f$, $c=\text{image center}$
        \item Pinhole with $f_x, f_y$, $c=\text{image center}$
        \item Pinhole with $f_x, f_y, c_x, x_y$ (used for RayZer~\citep{jiang2025rayzer})
        \item Double-Sphere~\cite{usenko2018double} camera model
    \end{itemize}

    For the Double-Sphere~\cite{usenko2018double} model, we parametrize the additional parameters $\xi, \alpha$ as
    \begin{equation}
        \xi = \tanh(\theta_\xi), \qquad \alpha = \sigma(\theta_\alpha - \epsilon_\alpha).
    \end{equation}
    This ensures that the ranges $\xi \in [-1, 1]$ and $\alpha \in [0, 1]$ are enforced. $\xi = 0, \alpha = 0$ recovers a pinhole camera, so we ensure that a zero init leads to $\xi = 0$ and $\alpha \approx 0$ by choosing an $\epsilon_\alpha > 1$. 
    
    We found no significant performance differences between these variants, so ultimately chose the simplest parametrization, which has the added benefit of the fewest degrees of freedom for the camera pose prediction.

    \paragraph{Alternative Extrinsics Parametrizations}
    Similarly, we explored different parametrizations for extrinsics, including the Zhou 6D parametrization~\citet{zhou2019continuity} used by RayZer~\cite{jiang2025rayzer}, the SVD parametrization used by $\pi^3$~\citet{wang2025pi3} (and later adopted by Open-RayZer~\citep{wang2025OpenRayzer}), and the $\mathfrak{se}(3)$ parametrization we ultimately chose.
    We found no significant performance differences between them, so we simply chose the most compact $\mathfrak{se}(3)$ parametrization.
    It is worth noting that this is different from common observations in pose \emph{regression} with direct supervision: the cyclical nature of the $\mathfrak{se}(3)$ parametrization becomes a problem there due to ambiguous targets. However, in end-to-end optimization supervised by a downstream loss, this is not a problem.

\section{\todo{Further Explorations}}

\subsection{Effect of ``Nuisance'' Dynamic State during NVS}\label{sec:dynamic_state_exploration}

We visualize the effect of our dynamic state $\state$ modeling in a standard dynamic video setting.
Crucially, this does \emph{not} represent our general inference setting -- we intend for this state to only be used during training (to obtain stable training behavior on general video) and subsequently discarded during inference.
Here, we analyze what happens when the state is used during inference, in two different settings:
i) dynamic camera, dynamic scene; and ii) static camera, dynamic scene.

\begin{figure}[ht]
    \centering
    \includegraphics[width=0.48\linewidth]{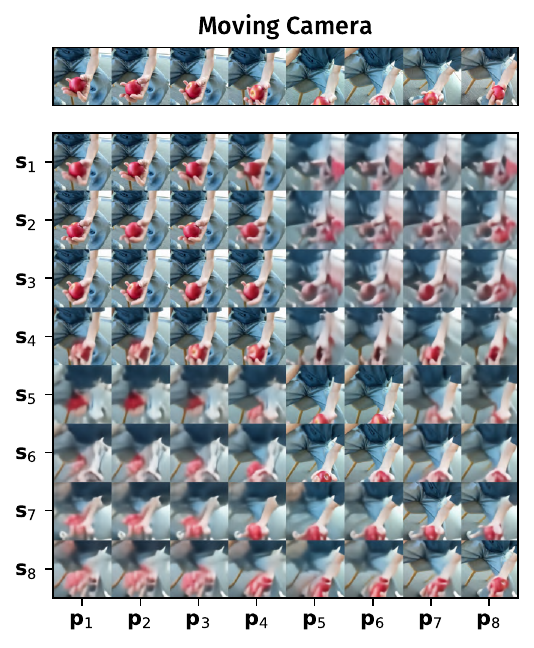}
    \includegraphics[width=0.48\linewidth]{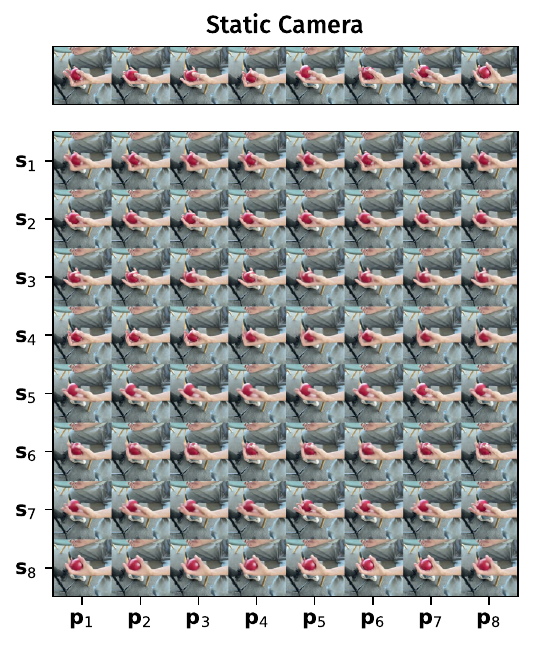}
    \caption{\textbf{Dynamic State Transplantation Across Time.} Starting from a video of 8 frames (top), we predict frame-wise poses and dynamic states, and then render views for each combination of state and pose.}
    \label{fig:dynamic_state_mismatch}
\end{figure}

Starting from a video, we use \methodname-\textbf{L} to extract a set of poses and dynamic states $\{(\pose_i, \state_i)\}_i$ and then render views for each combination $(\pose_i, \state_j)$, using the views $\{1, \ldots, N\} \setminus \{i\}$ as context.
We show the results of all combinations in \Cref{fig:dynamic_state_mismatch}.
The left side shows this setup starting from a video with a moving camera from DyCheck~\citep{gao2022dynamic}.
As expected, the diagonal, where both state and pose match, is typically the sharpest frame reconstruction.
When poses vary greatly between the view from which the state was extracted and with which the state is being rendered, the state seems to mostly supersede the pose in practice.
This shows that, as expected, given the fact that we do not require access to multi-view video training data to explicitly disentangle the dynamic state w.r.t.\ pixel-space image content and dynamic content, $\state$ models not a \emph{pure} dynamic state.
Instead, it serves to fulfill its primary role of \emph{stabilizing training in the presence of dynamics in the training videos}.

As expected, matching camera pose and state embedding produces the most accurate reconstruction.
In the case of a static camera and dynamic scene, poses, as expected, do not play a relevant role, and effectively the same video is reconstructable from all poses by just iterating through the state embeddings.
In the mismatched case, we observe more complex behavior: the state embedding seems to partially compete with the pose, resulting in blurry synthesized frames with mismatched geometry.
We interpret the qualitative results as the ``state'' embedding encoding information that helps obtain better reconstructions, but which are \emph{not} disentangled (unlike embeddings from methods like DyST~\cite{seitzer2024dyst}, which explicitly disentangle them by using multi-view video during training) and rather just encode the residual in the original image space.
We therefore consider this additional variable not an true representation of the scene's state itself but rather a nuisance variable whose sole role is improving training stability, and refer to it a such in the main paper.

\subsection{Further Failure Cases/Limitations}

\begin{figure}[t!]
    \centering

    \begin{subfigure}[t]{\linewidth}
        \centering
        \includegraphics[width=0.9\linewidth]{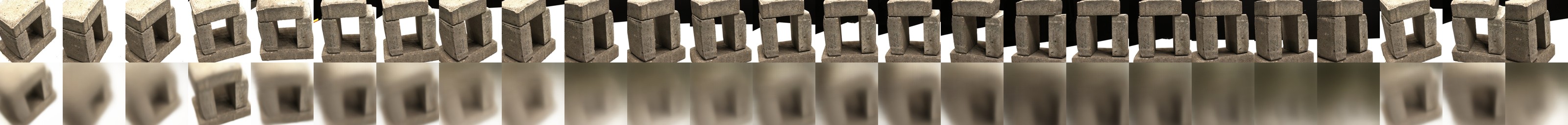}
        \label{fig:failurecase_dtu}
    \end{subfigure}

    \vspace{0.75em}

    \begin{subfigure}[t]{\linewidth}
        \centering
        \includegraphics[width=0.9\linewidth]{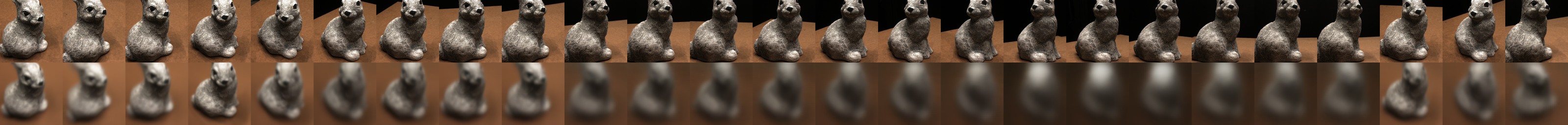}
        \label{fig:failurecase_dtu2}
    \end{subfigure}

    \caption{\textbf{Failure cases in DTU.} \methodname trained on SpatialVid often fails on DTU due to the evaluation setting differing too greatly from the training setting. We can observe that the camera estimation stage fails particularly for larger transforms between views.}
    \label{fig:failurecases_dtu}
\end{figure}

We have found that \methodname fails to produce good predictions on DTU in the setting of SEVA \cite{zhou2025stable}, as shown in \cref{fig:failurecase_dtu} and in \cref{tab:open_set_nvs}. The cause may be that the model is trained on open-set data of real world scenes, while the scenes in DTU are object centric with black and white backgrounds, placing them out of the training distribution. This explanation is consistent with the other experiments in \cref{tab:open_set_nvs}, where the rest of the eval settings are closer to the training setting.

\begin{figure}[t!]
    \centering
    \includegraphics[width=1\linewidth]{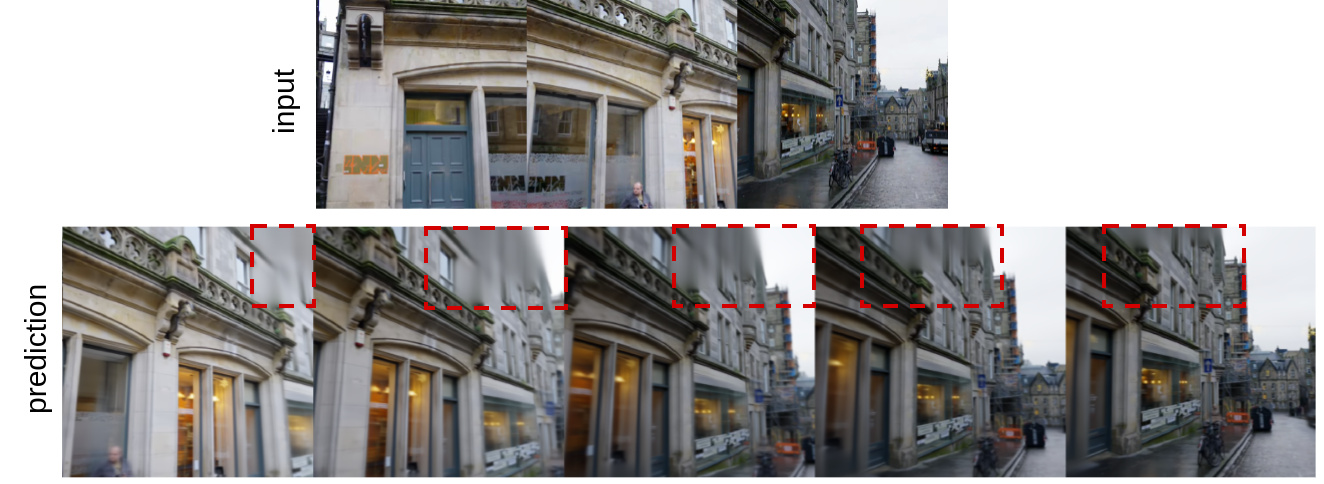}
    \caption{\textbf{Unobserved Regions.} \methodname predicts blurry ``averaged'' patches in regions unobserved in any of the context views.}
    \label{fig:failurecase_warp}
\end{figure}

Furthermore, we have observed that \methodname produces blurred artifacts for parts of novel views which are not observed in any of the given views, which is a result of the regression objective. This effect can be observed in \cref{fig:failurecase_warp}. Note that there is a clearly observable boundary which separates the observed and unobserved parts of the scene. This failure case has also been described in RayZer \cite{jiang2025rayzer}, as well as GS-LRM \cite{zhang2024gs} and LVSM \cite{jin2025lvsm}, all of which are trained with the same objective.
Additionally, just like RayZer, we have blurryness for fine details and objects close to the camera.

\subsection{Stability of RayZer trained on Dynamic Videos}\label{sec:rayzer_instabilities}

    In our attempts to train RayZer~\cite{jiang2025rayzer} models on video datasets such as SpatialVid~\cite{wang2025spatialvid} or SA-V~\cite{ravi2024sam2} using the official code, we found training to be very unstable.
    Specifically, models seem to converge during (very) early training and then diverge or stall abruptly.
    When this happens seems to be highly influenced by the choice of batch size and view distance, although we were unable to find stable configurations that enable learning true NVS.
    Generally, we train multiple variants of RayZer on SpatialVid and SA-V, with minimal changes compared to the official default configuration for DL3DV-10k.
    We train for 200k iterations, where the learning rate schedule and view selection schedule are adjusted accordingly.
    We explore two main variations:
    \begin{enumerate}
        \item First, we explore a variation with 2 input views and 3 target views for later evaluation in the PixelSplat \cite{charatan2024pixelsplat} setting on RealEstate10k.
        We adapt the default config used in the original RayZer for DL3DV-10k, adjust the view selection and learning rate schedule accordingly, and train for 200k iterations.
        More precisely, we scale the view selection such that the mean time passed between frames is the same in DL3DV-10k and the video datasets.
        This training recipe has worked the best for our experiments, and slight deviations from this recipe lead to even more degraded performance.
        \item Secondly, we explore a variant in the setting described in our main exploration, namely: 6 input views and 2 output views, and sampling at 2 fps.
        Note that these are the exact settings we train \methodname on.
    \end{enumerate}
    We train all models using the official implementation\footnote{\url{https://github.com/hwjiang1510/RayZer}}. The official trainer already uses gradient clipping and training step skipping when the gradient norm is too large, to improve stability during training.\footnote{We also identified bugs in the training code, such as gradients never being synchronized during DDP training without gradient accumulation, which we fixed. We explored training models both with and without these fixes enabled, and encountered similar instabilities across all configurations. All final numbers reported use the corrected implementation, which generally led to the best results.}
    In both settings described above, alongside a multitude of variations we explored, we observe divergences and stalled training at some point in the training process.
    Divergences are typically characterized by a sudden spike in the gradient norm and a subsequent sharp drop in training PSNR, marking a drop in the learned representation.
    A representative visualization is shown in \cref{fig:instabilities}.
    The resulting predictions resemble the mean of the input images.
    Stalled training, on the other hand, refers to the vast majority of training steps being skipped entirely once gradient norms exceed a set amount, for which we follow the original RayZer configuration, resulting in training progress stopping.
    Note that we have found disabling the skipping mechanism, while preventing the stalling, also leads to degeneracies during training, including divergences.

    In additional runs using the first setting, trained on SpatialVid, we further vary the batch size and view distance.
    We observe that the learned camera space converges to a degenerate solution during training, where $SE(3)$ interpolation between views results in a rotation of the camera between views.
    Generally, smaller batch sizes during training lead to faster divergences.
    
    In our main table, we use the runs with the highest evaluatiuon PSNR we were able to obtain before stalling/divergence.

\begin{figure}[h!]
    \centering
    \begin{minipage}{0.48\linewidth}
        \centering
        \includegraphics[width=\linewidth]{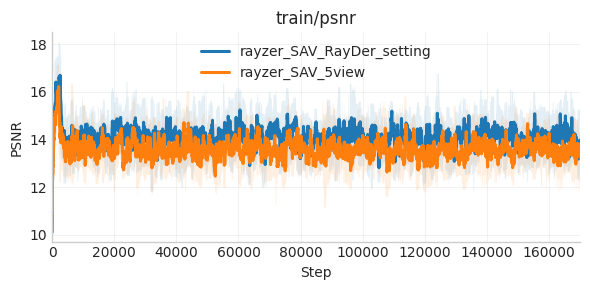}
    \end{minipage}
    \hfill
    \begin{minipage}{0.48\linewidth}
        \centering
        \includegraphics[width=\linewidth]{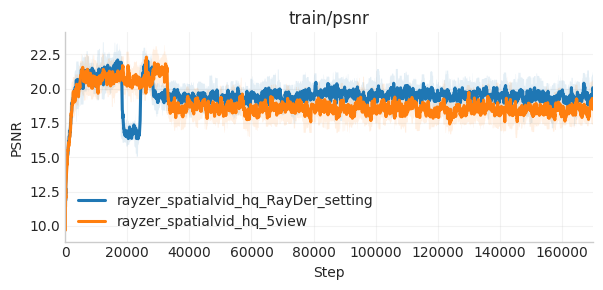}
    \end{minipage}

    \vspace{0.5em}

    \begin{minipage}{0.48\linewidth}
        \centering
        \includegraphics[width=\linewidth]{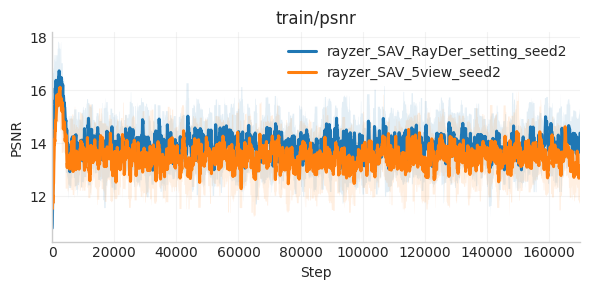}
    \end{minipage}
    \hfill
    \begin{minipage}{0.48\linewidth}
        \centering
        \includegraphics[width=\linewidth]{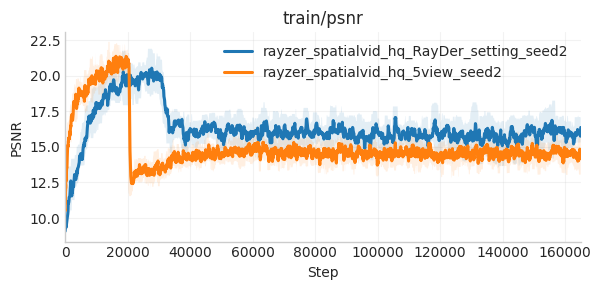}
    \end{minipage}

    \caption{Training runs (PSNR) of RayZer on video datasets. At some point during training, RayZer's PSNR drops sharply. This behaviour is representative across seeds.}
    \vspace{-1em}
    \label{fig:rayzer_failure_psnr}
\end{figure}

\section{Additional Evaluations}
We provide additional SSIM and LPIPS results for the open-set novel view synthesis evaluation in \cref{tab:open_set_nvs}. 
Importantly, \methodname-L-$576^2$ is trained using only the MSE reconstruction loss and does not use perceptual supervision, i.e., $\lambda_{\text{perc}} = 0$ in \cref{eq:trainloss}. 
The results are reported in \cref{tab:open_set_nvs_ssim} and \cref{tab:open_set_nvs_lpips}. 
Despite this purely reconstruction-based objective, \methodname achieves near state-of-the-art SSIM across several datasets. 
The LPIPS results are correspondingly weaker, which is consistent with the absence of perceptual loss during training.

\begin{table}[t]
    \centering
    \caption{\textbf{Open-set Novel View Synthesis (LPIPS$\downarrow$).} We extend the evaluation by \citet{zhou2025stable} and compute LPIPS across a large variety of settings (columns). Note that the \methodname model evaluated here was not trained with any perceptual loss.}
    \ifeccv\vspace{-.7em}\fi
    \ifeccv
        \newcolumntype{C}{>{\hspace{-.5pt}\footnotesize}c<{\hspace{-.5pt}}}
    \else
        \newcolumntype{C}{@{\hskip .15em}c@{\hskip .15em}}
    \fi
    \newcommand{\ti}[1]{\multicolumn{1}{c}{#1}}
    \adjustbox{max width=\linewidth}{\scalebox{0.8}{
      \begin{tabular}{l@{\hskip \ifeccv-.4em\else0em\fi}crCCCCCCCCCC c CCCCCCC}
        & & & \multicolumn{10}{c}{\textit{small-viewpoint}} & {\hskip .2em} & \multicolumn{7}{c}{\textit{large-viewpoint}} \\
        \cmidrule(lr){4-13} \cmidrule(lr){15-21}
        \toprule

        & &  Dataset $\!\!\rightarrow\!\!\!\!\!\!\!$ & \multicolumn{2}{c}{LLFF} & \multicolumn{2}{c}{DTU} & \multicolumn{2}{c}{CO3D} & \multicolumn{2}{c}{WRGBD} & {\normalsize {\hskip -1em}M360{\hskip -1em}} & {\normalsize {\hskip -1em}T\&T{\hskip -1em}} & & {\normalsize {\hskip -1em}CO3D{\hskip -1em}} & \multicolumn{2}{c}{WRGBD} & \multicolumn{2}{c}{M360} &\multicolumn{2}{c}{T\&T}\\
        \cmidrule(lr){4-5} \cmidrule(lr){6-7} \cmidrule(lr){8-9} \cmidrule(lr){10-11} \cmidrule(lr){12-12} \cmidrule(lr){13-13} \cmidrule(lr){15-15} \cmidrule(lr){16-17} \cmidrule(lr){18-19} \cmidrule(lr){20-21}
        & & Split $\!\!\rightarrow\!\!\!\!\!\!\!$ & \multicolumn{2}{c}{R} & \multicolumn{2}{c}{R} & \ti{V} & \ti{R} & \ti{S\textsubscript{e}} & \ti{S\textsubscript{h}} & \ti{R} & \ti{V} & & \ti{R} & \multicolumn{2}{c}{S\textsubscript{h}} & \multicolumn{2}{c}{R} & \multicolumn{2}{c}{S} \\
        \cmidrule(lr){4-5} \cmidrule(lr){6-7} \cmidrule(lr){8-8} \cmidrule(lr){9-9} \cmidrule(lr){10-10} \cmidrule(lr){11-11} \cmidrule(lr){12-12} \cmidrule(lr){13-13} \cmidrule(lr){15-15} \cmidrule(lr){16-17} \cmidrule(lr){18-19} \cmidrule(lr){20-21}
        Model & Params & \smash{\shortstack{{\vspace{-.5em}\shortstack{Self-\\sup.}}}} $|\mathcal{I}_\text{in}|\!\!\rightarrow\!\!\!\!\!\!\!$& \ti{1} & \ti{3} & \ti{1} & \ti{3} & \ti{1} & \ti{3} & \ti{3} & \ti{6} & \ti{6} & \ti{1} & & \ti{1} & \ti{1} & \ti{3} & \ti{1} & \ti{3} & \ti{3} & \ti{6} \\
        \midrule
        
        MVSplat~\citep{chen2024mvsplat} & 12M & {\color{ourred}\xmark}{\hskip 2.8em} & 0.542  & 0.497 & 0.386  & 0.310  & 0.634 & 0.614 & 0.504 & 0.643  & 0.556 & 0.519 & & -- & -- & -- & -- & -- & -- & -- \\
        DepthSplat~\citep{xu2025depthsplat} & 354M & {\color{ourred}\xmark}{\hskip 2.8em} & 0.530  & 0.465 & \underline{0.369} & 0.304  & 0.618 & 0.603&  0.499 & 0.530 & 0.534 & 0.462  & & 0.756 & 0.732 & 0.588 &  0.691 & 0.491 & 0.706 & 0.611 \\
        ViewCrafter$^\dagger$~\citep{yu2024viewcraftertamingvideodiffusion} & 1.4B & {\color{ourred}\xmark}{\hskip 2.8em} & 0.620  & 0.435 & 0.485 & \underline{0.272}  & \underline{0.324} & \underline{0.513} &  \underline{0.324} & 0.639  & 0.464 & \samebf{0.283} & & 0.789 & 0.775 & {0.603} & 0.723 & 0.540 & 0.671 & 0.604 \\
        SEVA$^\dagger$~\citep{zhou2025stable} & 1.3B & {\color{ourred}\xmark}{\hskip 2.8em} & \underline{0.389}  & \underline{0.181} & \samebf{0.316} & \samebf{0.158}  & \samebf{0.318} & \samebf{0.278} &  \samebf{0.215} & \samebf{0.237} & \underline{0.319} & \underline{0.354} & & \samebf{0.445} & \samebf{0.423} & \samebf{0.289} &  \underline{0.573} & \underline{0.364} & \samebf{0.463} & \underline{0.387} \\
        Kaleido$^{\dagger\ddagger}$~\cite{liu2025scaling} & 3.1B & {\color{ourred}\xmark}{\hskip 2.8em} & \samebf{0.301} & \samebf{0.123} & -- & -- & -- & -- & -- & -- & \samebf{0.286} & -- & & -- & -- & -- & \samebf{0.530} & \samebf{0.344} & \underline{0.465} & \samebf{0.363} \\
        
        \midrule
        E-RayZer$^*$ ~\citep{zhao2025erayzer} & 246M & {\color{ourgreen}\cmark}{\hskip 2.7em} & 0.505 & 0.438 & 0.540 & 0.393 & 0.528 & 0.529 & 0.439 & \underline{0.528} & 0.678 & 0.585 & & 0.626 & 0.653 & \underline{0.588} & 0.738 & 0.699 & 0.688 & 0.678 \\
        RayDer-L-$576^2$ (Ours) & 743M & {\color{ourgreen}\cmark}{\hskip 2.7em} & 0.586 & 0.352 & 0.508 & 0.461  & 0.494 & 0.565 & 0.468 & \underline{0.588} & 0.743 & 0.528 & & \underline{0.623} & \underline{0.647} & 0.625 & 0.766 & 0.752 & 0.746 & 0.732 \\
        \bottomrule
        \multicolumn{13}{c}{\ifeccv\scriptsize\else\footnotesize\fi \shortstack{\vphantom{$^\dagger$}Split abbreviations: R: ReconFusion~\cite{wu2024reconfusion}; V: ViewCrafter~\citep{yu2024viewcraftertamingvideodiffusion}; S\textsubscript{\{e,h\}}: SEVA~\citep{zhou2025stable}, easy (e) and hard (h) variants.\\[-.33em]Dataset references: LLFF~\cite{mildenhall2019llff}, DTU~\cite{jensen2014large}, CO3D~\cite{liu24uco3d}, WRGBD~\cite{xia2024rgbd}, M360~\cite{barron2022mip}, T\&T~\cite{knapitsch2017tanks}}} & &
        \multicolumn{7}{c}{\ifeccv\scriptsize\else\footnotesize\fi \shortstack{$^\ddagger$Kaleido evaluates at $512^2$ instead of $576^2$\\[-.33em]$^\dagger$Diffusion-based models. $^*$Multi-dataset Ckpt}}
    \end{tabular}
    }}
    \vspace{-1em}
    \label{tab:open_set_nvs_lpips}
\end{table}

\begin{table}[t]
    \centering
    \caption{\textbf{Open-set Novel View Synthesis (SSIM$\uparrow$).} We extend the evaluation by \citet{zhou2025stable} and compute SSIM across a large variety of settings (columns). Despite being trained fully self-supervised and without large-scale video diffusion pretraining, RayDer is (near-)state-of-the-art across the majority of datasets and evaluation settings.}
    \ifeccv\vspace{-.7em}\fi
    \ifeccv
        \newcolumntype{C}{>{\hspace{-.5pt}\footnotesize}c<{\hspace{-.5pt}}}
    \else
        \newcolumntype{C}{@{\hskip .15em}c@{\hskip .15em}}
    \fi
    \newcommand{\ti}[1]{\multicolumn{1}{c}{#1}}
    \adjustbox{max width=\linewidth}{\scalebox{0.8}{
      \begin{tabular}{l@{\hskip \ifeccv-.4em\else0em\fi}crCCCCCCCCCC c CCCCCCC}
        & & & \multicolumn{10}{c}{\textit{small-viewpoint}} & {\hskip .2em} & \multicolumn{7}{c}{\textit{large-viewpoint}} \\
        \cmidrule(lr){4-13} \cmidrule(lr){15-21}
        \toprule

        & &  Dataset $\!\!\rightarrow\!\!\!\!\!\!\!$ & \multicolumn{2}{c}{LLFF} & \multicolumn{2}{c}{DTU} & \multicolumn{2}{c}{CO3D} & \multicolumn{2}{c}{WRGBD} & {\normalsize {\hskip -1em}M360{\hskip -1em}} & {\normalsize {\hskip -1em}T\&T{\hskip -1em}} & & {\normalsize {\hskip -1em}CO3D{\hskip -1em}} & \multicolumn{2}{c}{WRGBD} & \multicolumn{2}{c}{M360} &\multicolumn{2}{c}{T\&T}\\
        \cmidrule(lr){4-5} \cmidrule(lr){6-7} \cmidrule(lr){8-9} \cmidrule(lr){10-11} \cmidrule(lr){12-12} \cmidrule(lr){13-13} \cmidrule(lr){15-15} \cmidrule(lr){16-17} \cmidrule(lr){18-19} \cmidrule(lr){20-21}
        & & Split $\!\!\rightarrow\!\!\!\!\!\!\!$ & \multicolumn{2}{c}{R} & \multicolumn{2}{c}{R} & \ti{V} & \ti{R} & \ti{S\textsubscript{e}} & \ti{S\textsubscript{h}} & \ti{R} & \ti{V} & & \ti{R} & \multicolumn{2}{c}{S\textsubscript{h}} & \multicolumn{2}{c}{R} & \multicolumn{2}{c}{S} \\
        \cmidrule(lr){4-5} \cmidrule(lr){6-7} \cmidrule(lr){8-8} \cmidrule(lr){9-9} \cmidrule(lr){10-10} \cmidrule(lr){11-11} \cmidrule(lr){12-12} \cmidrule(lr){13-13} \cmidrule(lr){15-15} \cmidrule(lr){16-17} \cmidrule(lr){18-19} \cmidrule(lr){20-21}
        Model & Params & \smash{\shortstack{{\vspace{-.5em}\shortstack{Self-\\sup.}}}} $|\mathcal{I}_\text{in}|\!\!\rightarrow\!\!\!\!\!\!\!$& \ti{1} & \ti{3} & \ti{1} & \ti{3} & \ti{1} & \ti{3} & \ti{3} & \ti{6} & \ti{6} & \ti{1} & & \ti{1} & \ti{1} & \ti{3} & \ti{1} & \ti{3} & \ti{3} & \ti{6} \\
        \midrule
        
        MVSplat~\citep{chen2024mvsplat} & 12M & {\color{ourred}\xmark}{\hskip 2.8em} & 0.283 & 0.358 &  0.576 & 0.624& 0.403 & 0.370 &  0.405 & 0.368 & 0.312  &  0.394 & & -- & -- & -- & -- & -- & -- & -- \\
        DepthSplat~\citep{xu2025depthsplat} & 354M & {\color{ourred}\xmark}{\hskip 2.8em} & 0.299 & 0.396 &  \underline{0.601} & 0.638 & 0.429 & 0.402& 0.436 & 0.417 & {0.324} &  0.413 & & 0.385 &  0.234 & 0.335 & 0.206 & 0.291  & 0.315 &0.326 \\
        ViewCrafter$^\dagger$~\citep{yu2024viewcraftertamingvideodiffusion} & 1.4B & {\color{ourred}\xmark}{\hskip 2.8em} & 0.146 & 0.454 & 0.542 &\underline{0.671} & \underline{0.641} & 0.483 & 0.465 & 0.376 & 0.354 & \underline{0.563} & & 0.277 & 0.225 & 0.321 & 0.199 & 0.264 & 0.328 & 0.337 \\
        SEVA$^\dagger$~\citep{zhou2025stable} & 1.3B & {\color{ourred}\xmark}{\hskip 2.8em} & \underline{0.384} &0.602& 0.585 & 0.647 & 0.585 & \underline{0.647} & \underline{0.670} & \samebf{0.646} & \underline{0.395} & 0.437 & &  0.536 &  0.505 & \samebf{0.603} & \underline{0.282} & \samebf{0.377} & 0.385 & 0.427 \\
        Kaleido$^{\dagger\ddagger}$~\cite{liu2025scaling} & 3.1B & {\color{ourred}\xmark}{\hskip 2.8em} & 0.375 & \samebf{0.659} & -- & -- & -- & -- & -- & -- &\samebf{0.433} & -- & & -- & -- & -- &  0.248 & \underline{0.361} & 0.368 & 0.429 \\
        
        \midrule
        E-RayZer$^*$ ~\citep{zhao2025erayzer} & 246M & {\color{ourgreen}\cmark}{\hskip 2.7em} & 0.287 & 0.519 &  0.492 & 0.669 & 0.569 & 0.629 & 0.648 & 0.572 & 0.347 & 0.431 & & \underline{0.560} & \underline{0.508} & 0.542 & 0.273 & 0.336 & \underline{0.423} & \underline{0.433} \\
        RayDer-L-$576^2$ (Ours) & 743M & {\color{ourgreen}\cmark}{\hskip 2.7em} & \samebf{0.469} & \underline{0.650} & \samebf{0.654} & \samebf{0.702} & \samebf{0.668} & \samebf{0.657} & \samebf{0.672} & \underline{0.601} & 0.339 & \samebf{0.578} & & \samebf{0.625} & \samebf{0.558} & \underline{0.577} & \samebf{0.339} & 0.353 & \samebf{0.447} & \samebf{0.450} \\
        \bottomrule
        \multicolumn{13}{c}{\ifeccv\scriptsize\else\footnotesize\fi \shortstack{\vphantom{$^\dagger$}Split abbreviations: R: ReconFusion~\cite{wu2024reconfusion}; V: ViewCrafter~\citep{yu2024viewcraftertamingvideodiffusion}; S\textsubscript{\{e,h\}}: SEVA~\citep{zhou2025stable}, easy (e) and hard (h) variants.\\[-.33em]Dataset references: LLFF~\cite{mildenhall2019llff}, DTU~\cite{jensen2014large}, CO3D~\cite{liu24uco3d}, WRGBD~\cite{xia2024rgbd}, M360~\cite{barron2022mip}, T\&T~\cite{knapitsch2017tanks}}} & &
        \multicolumn{7}{c}{\ifeccv\scriptsize\else\footnotesize\fi \shortstack{$^\ddagger$Kaleido evaluates at $512^2$ instead of $576^2$\\[-.33em]$^\dagger$Diffusion-based models. $^*$Multi-dataset Ckpt}}
    \end{tabular}
    }}
    \vspace{-1em}
    \label{tab:open_set_nvs_ssim}
\end{table}

\clearpage

\section{Additional Samples}\label{sec:app_additional_samples}
We show additional qualitative examples in \cref{fig:app_qual_wildrgbd,fig:app_qual_dl3dvbench,fig:app_qual_r10k,fig:app_qual_wildrgbd_2,fig:app_qual_wildrgbd_3,fig:app_qual_llff}.
We also provide video visualizations of interpolated fly-throughs through various scenes from the DL3DV-10K~\cite{ling2024dl3dv} Eval set as videos in the supplementary material, including comparisons with E-RayZer~\cite{zhao2025erayzer} -- the primary prior self-supervised NVS method trained on a mixture of static-scene datasets.

\newlength{\labw}
\setlength{\labw}{10mm}

\newcommand{\sampleblock}[1]{%
  \adjustbox{max width=\linewidth}{
  \begin{tikzpicture}
    \node[inner sep=0pt, anchor=north west] (img)
      at (\labw,0)
      {\includegraphics[width=\linewidth-\labw]{#1}};
    \node[anchor=east, align=right]
      at ([xshift=3mm]$(img.north west)!0.25!(img.south west)$)
      {\rotatebox{90}{\shorttabulara{\scriptsize \textbf{Targets}}}};
    \node[anchor=east, align=right]
      at ([xshift=3mm]$(img.north west)!0.75!(img.south west)$)
      {\rotatebox{90}{\shorttabulara{\scriptsize\textbf{Predictions}}}};
  \end{tikzpicture}%
  }
}


\newcommand{\sampleblockER}[1]{%
  \adjustbox{max width=\linewidth}{
  \begin{tikzpicture}
    \node[inner sep=0pt, anchor=north west] (img)
      at (\labw,0)
      {\includegraphics[width=\linewidth-\labw]{#1}};

    \node[anchor=east, align=right]
      at ([xshift=3mm]$(img.north west)!0.1667!(img.south west)$)
      {\rotatebox{90}{\shorttabulara{\scriptsize\textbf{E-RayZer}}}};

    \node[anchor=east, align=right]
      at ([xshift=3mm]$(img.north west)!0.5!(img.south west)$)
      {\rotatebox{90}{\shorttabulara{\scriptsize\textbf{RayDer}}}};

    \node[anchor=east, align=right]
      at ([xshift=3mm]$(img.north west)!0.8333!(img.south west)$)
      {\rotatebox{90}{\shorttabulara{\scriptsize\textbf{Targets}}}};
  \end{tikzpicture}%
  }
}

\newcommand{\sampleblockint}[1]{%
  \adjustbox{max width=\linewidth}{%
    \includegraphics{#1}%
  }%
}






\newcommand{\sampleblockdlthreedv}[1]{\sampleblock{img//imgs//samples//dl3dv10k_benchmark//rayder_b_50k_cosine/#1_gt_vs_pred.jpg}}

\begin{figure}
    \centering
    \sampleblockdlthreedv{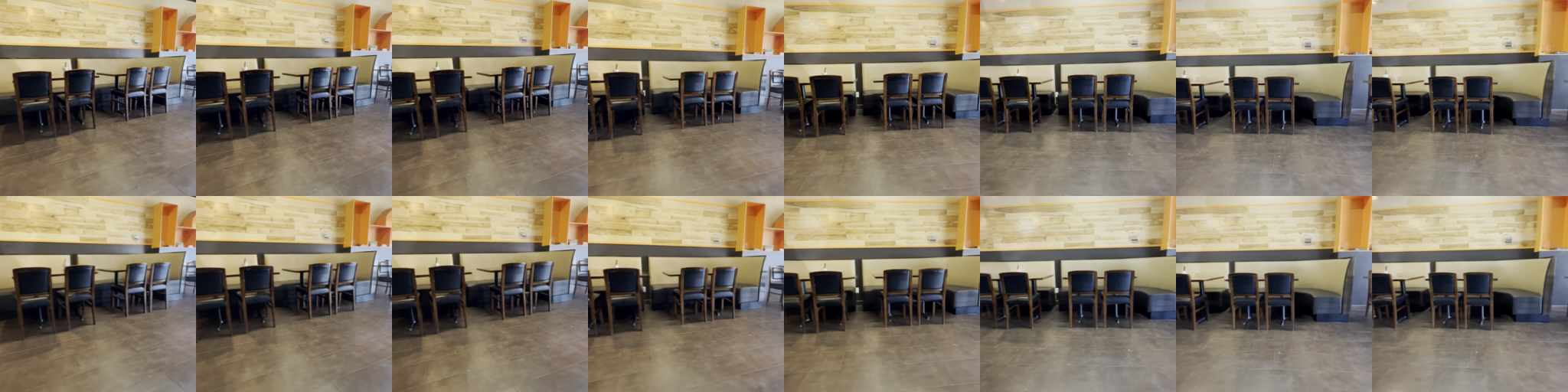}\\[2mm]
    \sampleblockdlthreedv{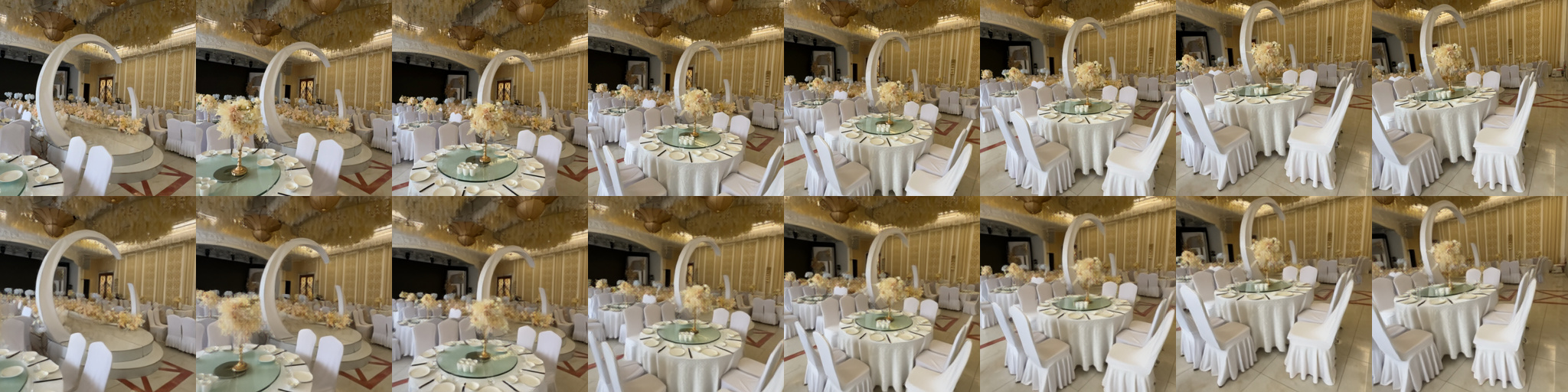}\\[2mm]
    \sampleblockdlthreedv{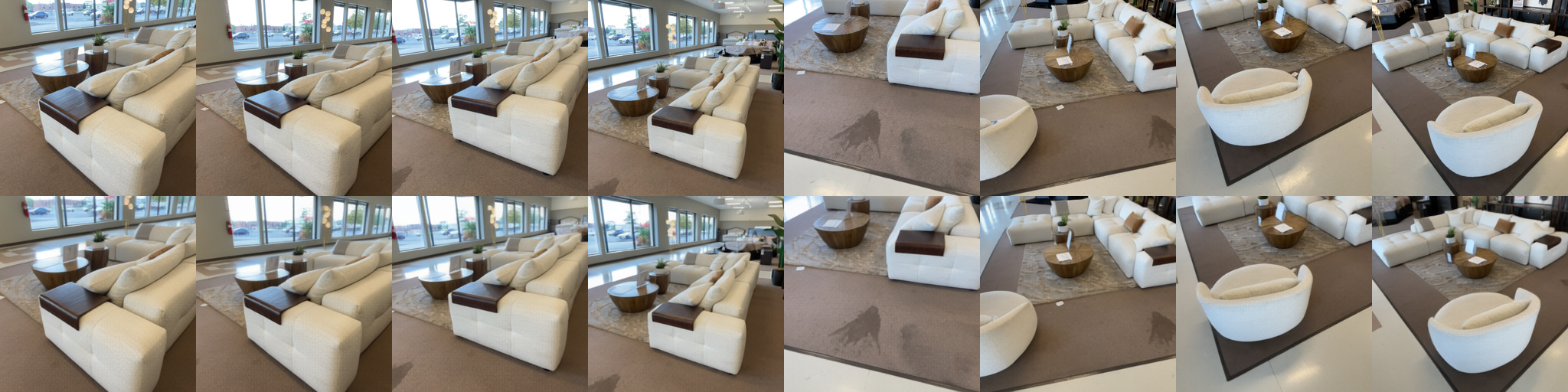}\\[2mm]
    \sampleblockdlthreedv{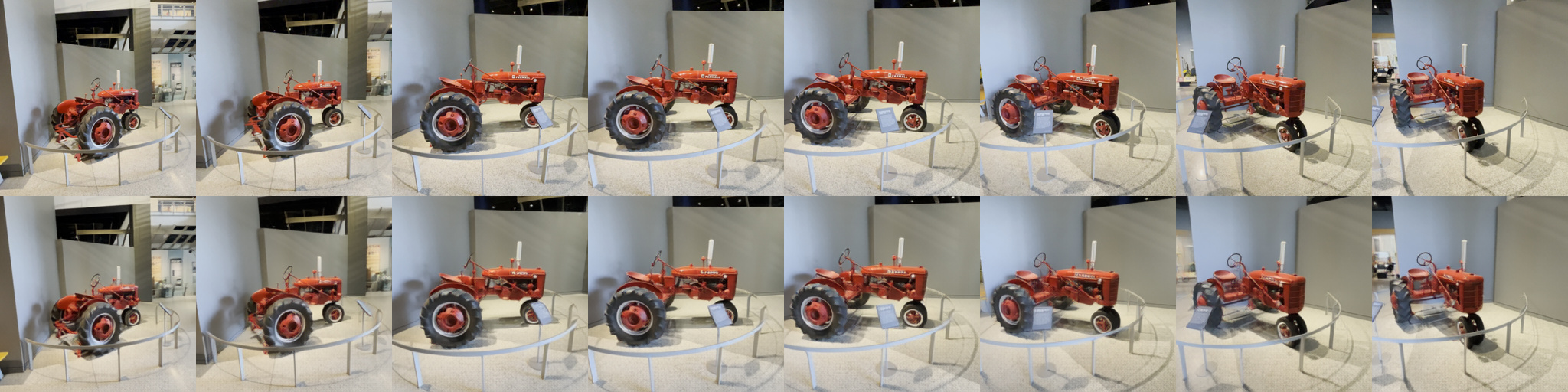}\\[2mm]
    \sampleblockdlthreedv{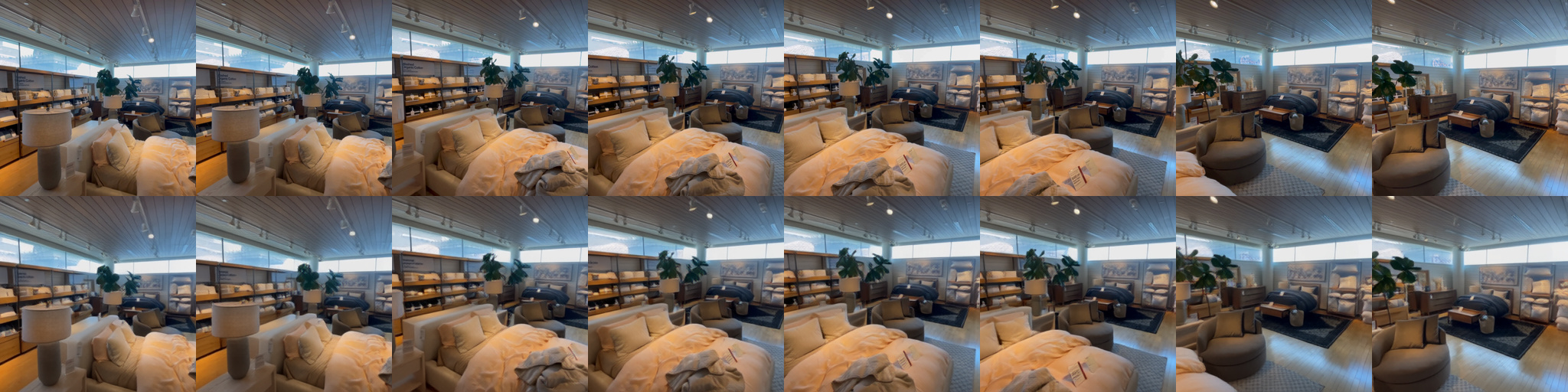}\\[2mm]
    
    
    \caption{\textbf{\methodname trained on static data.} Novel View Samples from RayDer-B trained for 50k iterations on DL3DV-10k \citep{ling2024dl3dv}. The ground-truth images are at the top, generated novel views are at the bottom. The input-images follow the official RayZer \citep{wang2025OpenRayzer} \textit{even} indices for DL3DV-10k Benchmark.}
    \label{fig:app_qual_dl3dvbench}
\end{figure}

\newcommand{\sampleblockopen}[1]{\sampleblockER{img//imgs//samples//qualitative//openset//tgt_vs_pred_#1.jpg}}

\begin{figure}[h!]
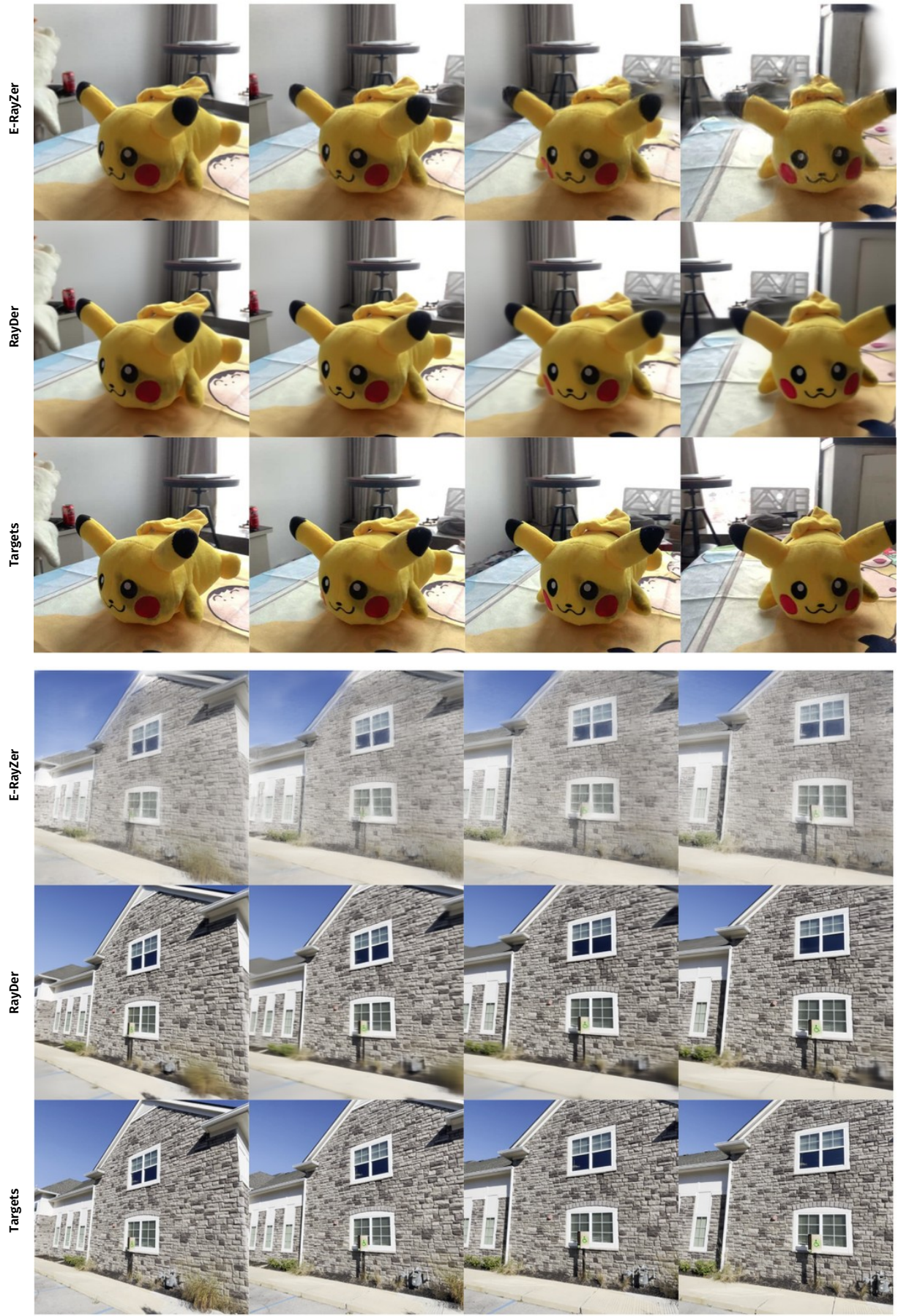

    \centering
    \sampleblockopen{1}\\[2mm]
    \sampleblockopen{2}\\[2mm]
    \caption{Zero-shot Open-set samples on WildRGBD and DL3DV-10k evaluation using \methodname-L-576 with a sparse number of input views.}
    \vspace{-1em}
    \label{fig:app_qual_wildrgbd}
\end{figure}

\newcommand{\sampleblockinterpolation}[1]{\sampleblockint{img//imgs//samples//qualitative//interpolation//frame_grid_#1.jpg}}

\begin{figure}[h!]
    \centering
    \sampleblockinterpolation{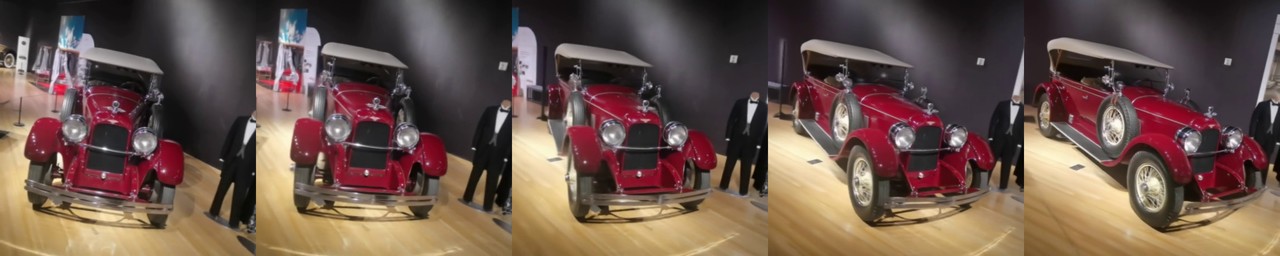}\\[2mm]
    \sampleblockinterpolation{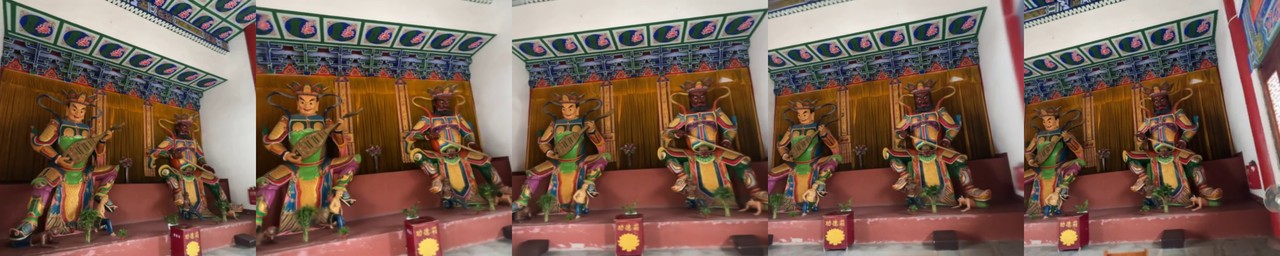}\\[2mm]
    \sampleblockinterpolation{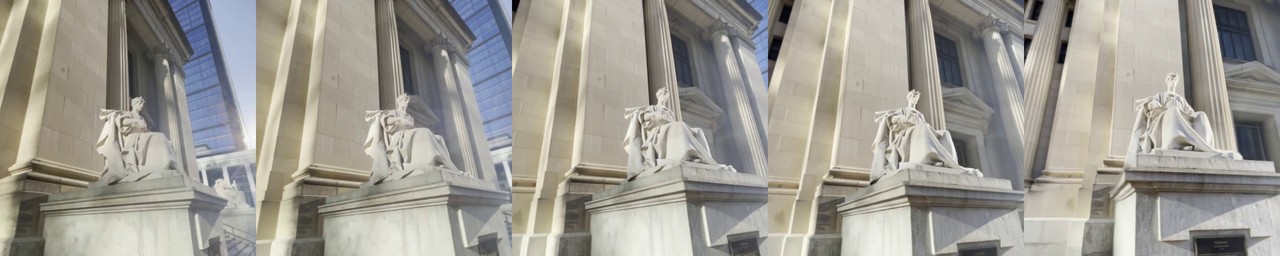}\\[2mm]
    \sampleblockinterpolation{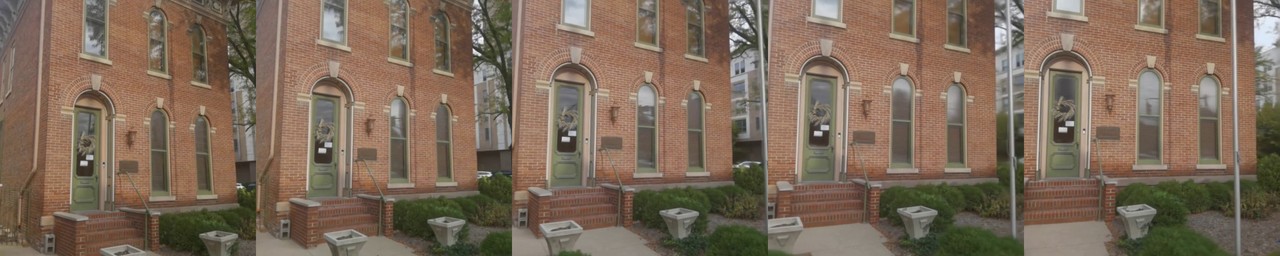}\\[2mm]
    \sampleblockinterpolation{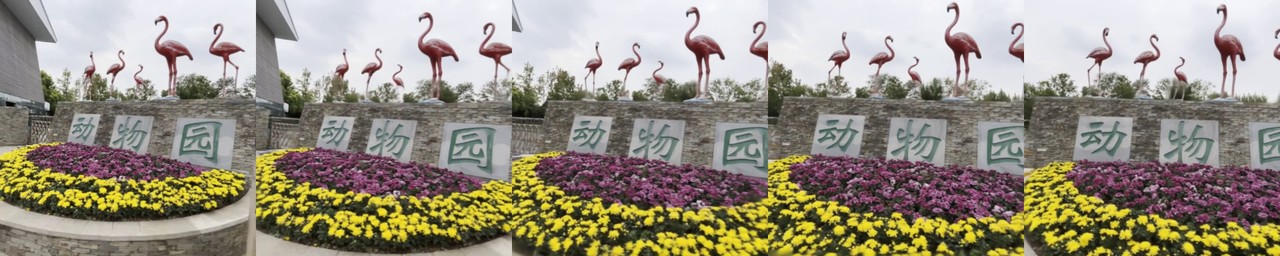}\\[2mm]
    \sampleblockinterpolation{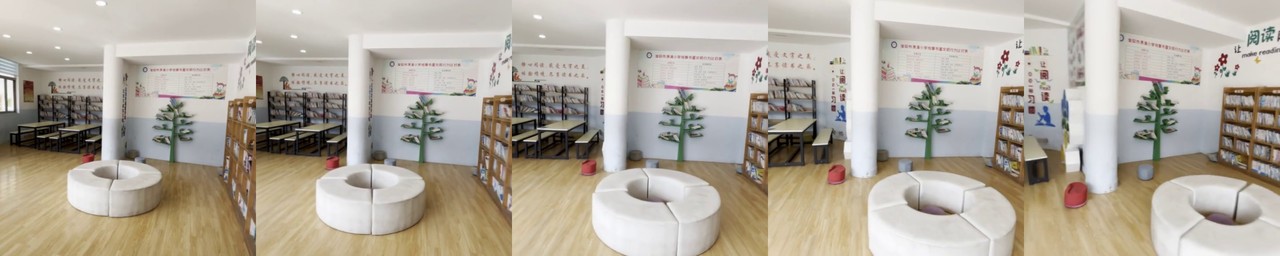}\\[2mm]
    \caption{\textbf{Zero-shot view interpolation on DL3DV-10k}. Given a sparse set of context images, our model synthesizes smooth, intermediate novel views by interpolating between the predicted camera poses.}
    \vspace{-1em}
    \label{fig:app_qual_interpolation}
\end{figure}

\newcommand{\sampleblockrealestate}[1]{\sampleblock{img//imgs//samples//r10k//rayder_L_680k//#1//gt_vs_pred.jpg}}

\begin{figure}[t]
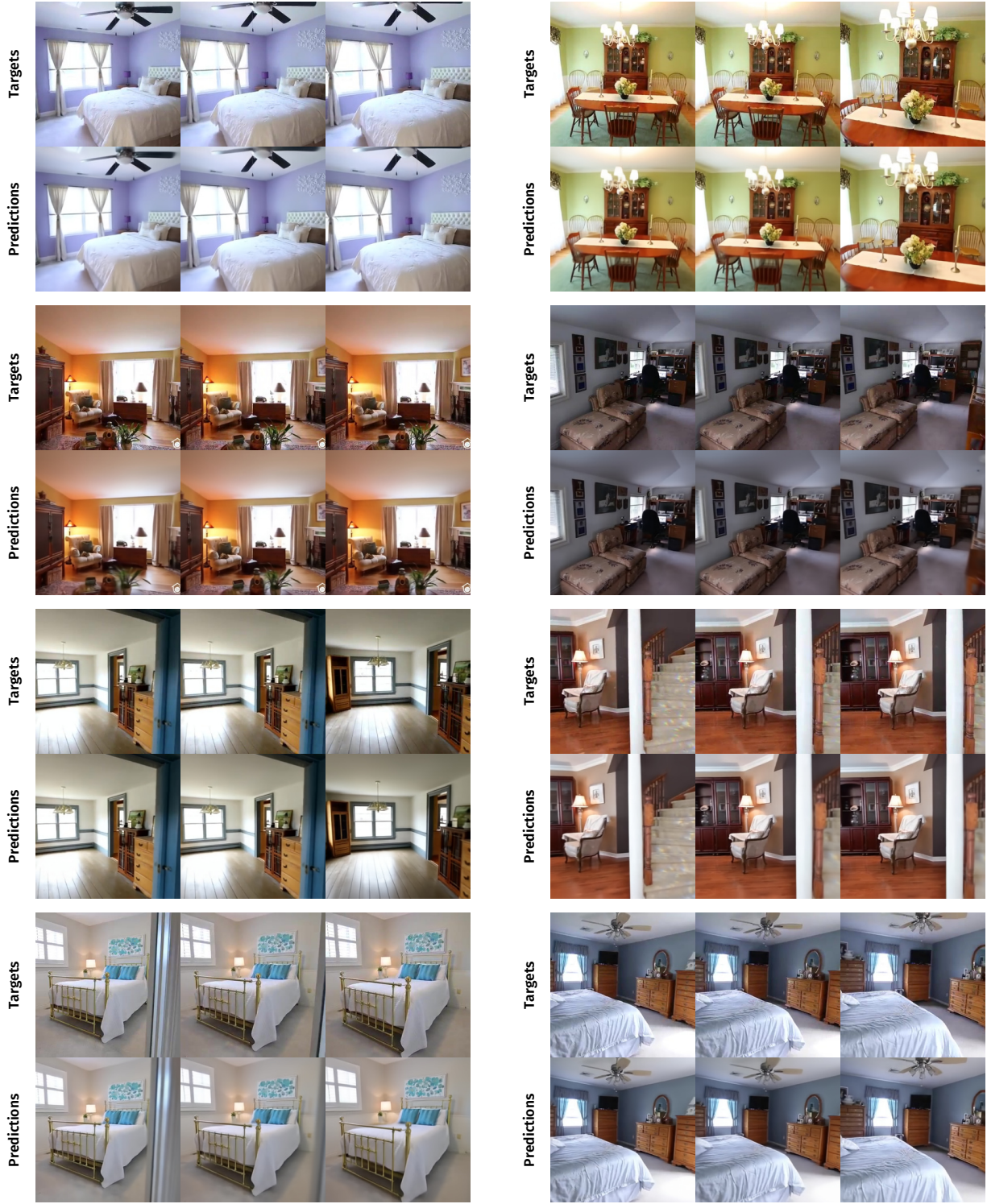

\centering

\newlength{\colw}
\setlength{\colw}{0.49\linewidth}

\begin{minipage}[t]{\colw}
\centering
\sampleblockrealestate{000034}\\[2mm]
\sampleblockrealestate{000137}\\[2mm]
\sampleblockrealestate{000284}\\[2mm]
\sampleblockrealestate{000444}\\[2mm]

\end{minipage}\hfill
\begin{minipage}[t]{\colw}
\centering
\sampleblockrealestate{000422}\\[2mm]
\sampleblockrealestate{000169}\\[2mm]
\sampleblockrealestate{000401}\\[2mm]
\sampleblockrealestate{000971}\\[2mm]
\end{minipage}

\caption{\textbf{\methodname trained on dynamic data, Zero-shot on static data}
Novel View Synthesis samples from \methodname -L on RealEstate10k using the official PixelSplat \cite{charatan2024pixelsplat} input-indices with two input images.}
\label{fig:app_qual_r10k}
\end{figure}

\newcommand{\sampleblockwildrgbd}[1]{\sampleblock{img//imgs//samples//qualitative//rayder_L_680k//WildRGBD/tgt_vs_pred_#1.jpg}}

\begin{figure}
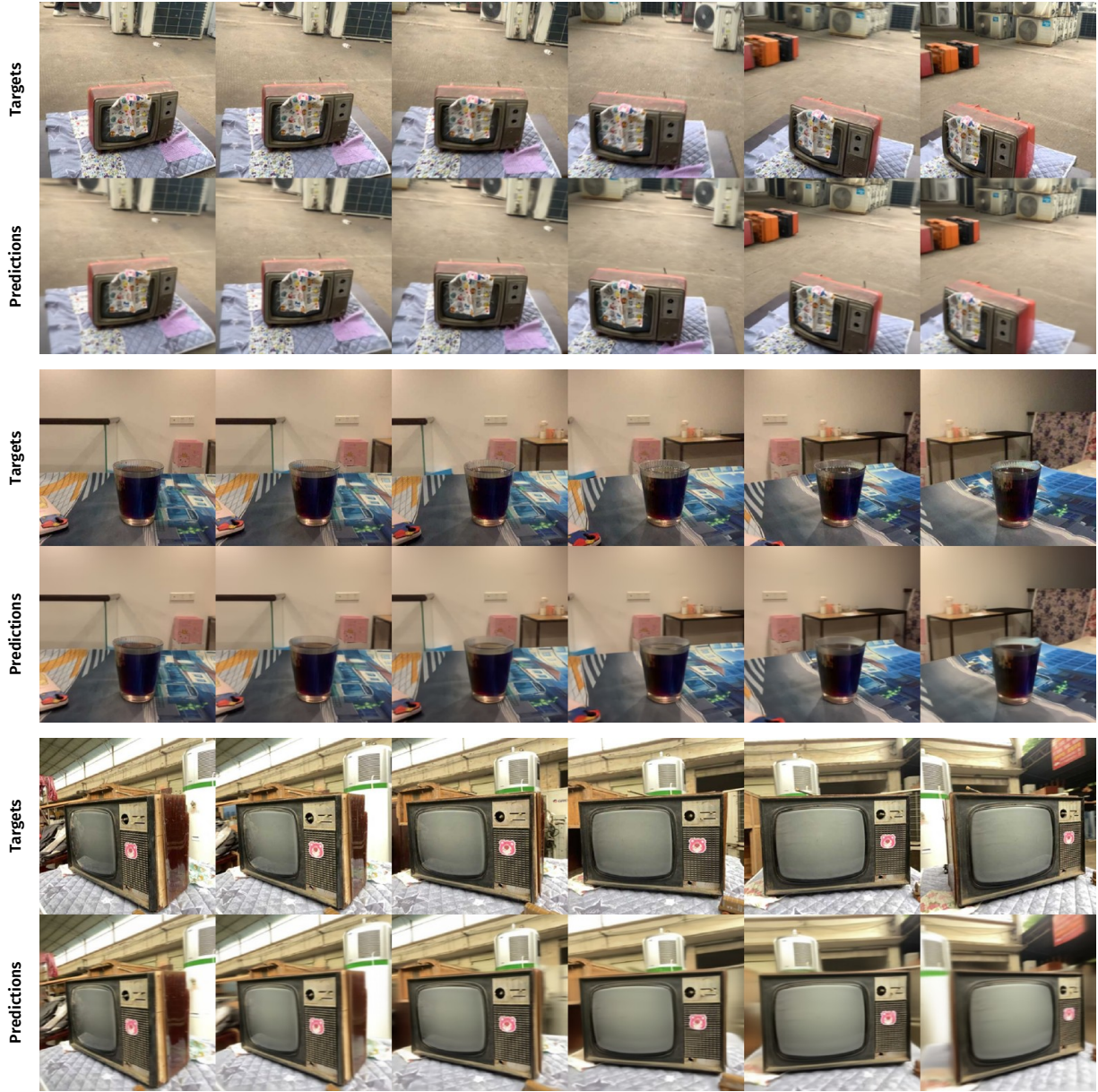

\centering
\sampleblockwildrgbd{1}\\[2mm]
\sampleblockwildrgbd{2}\\[2mm]
\sampleblockwildrgbd{3}\\[2mm]
\caption{\textbf{\methodname trained on dynamic data, Zero-shot on static data} Novel View Synthesis samples from \methodname -L on WildRGBD}
\label{fig:app_qual_wildrgbd_2}
\end{figure}

\newcommand{\sampleblockwildrgbdsparse}[1]{\sampleblock{img//imgs//samples//qualitative//rayder_L_680k//WildRGBD_2ctxview/tgt_vs_pred_#1.jpg}}

\begin{figure}
\centering
\sampleblockwildrgbdsparse{1}\\[2mm]
\sampleblockwildrgbdsparse{2}\\[2mm]
\sampleblockwildrgbdsparse{3}\\[2mm]
\sampleblockwildrgbdsparse{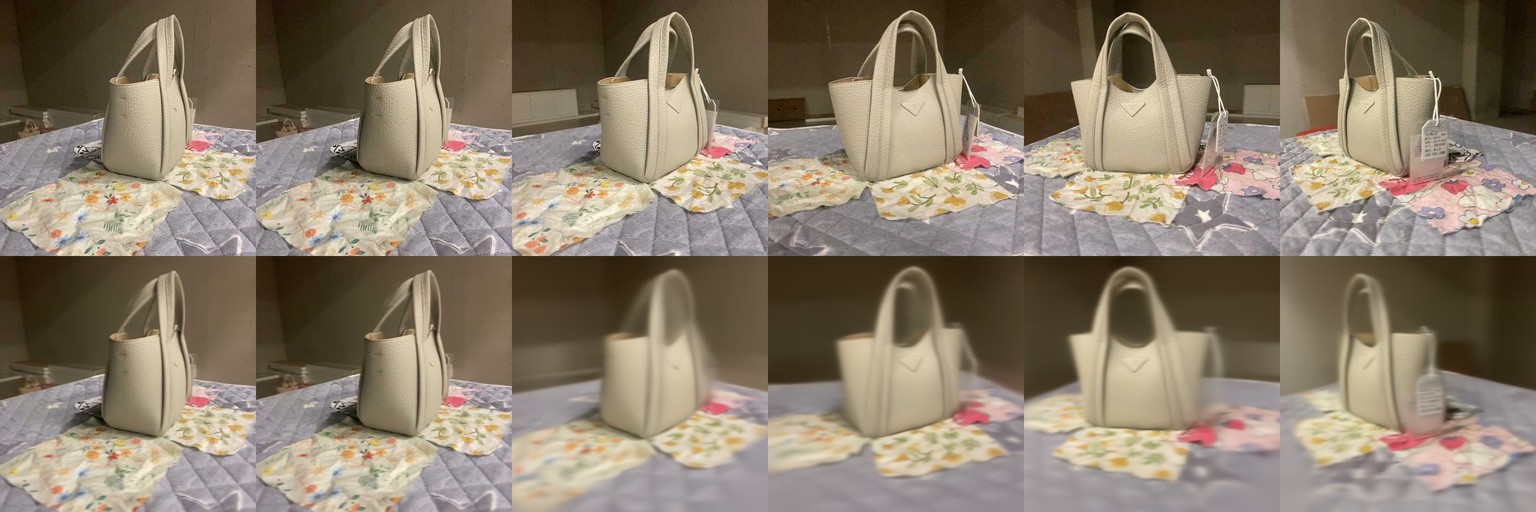}\\[2mm]

\caption{\textbf{\methodname trained on dynamic data, Zero-shot on static data, sparse view setting} Novel View Synthesis samples from \methodname -L on WildRGBD with 2 input views.}
\label{fig:app_qual_wildrgbd_3}
\end{figure}



\newcommand{\sampleblockllfff}[1]{\sampleblock{img//imgs//samples//qualitative//rayder_L_680k//LLFF//tgt_vs_pred_llff_#1.jpg}}

\begin{figure}
\centering
\sampleblockllfff{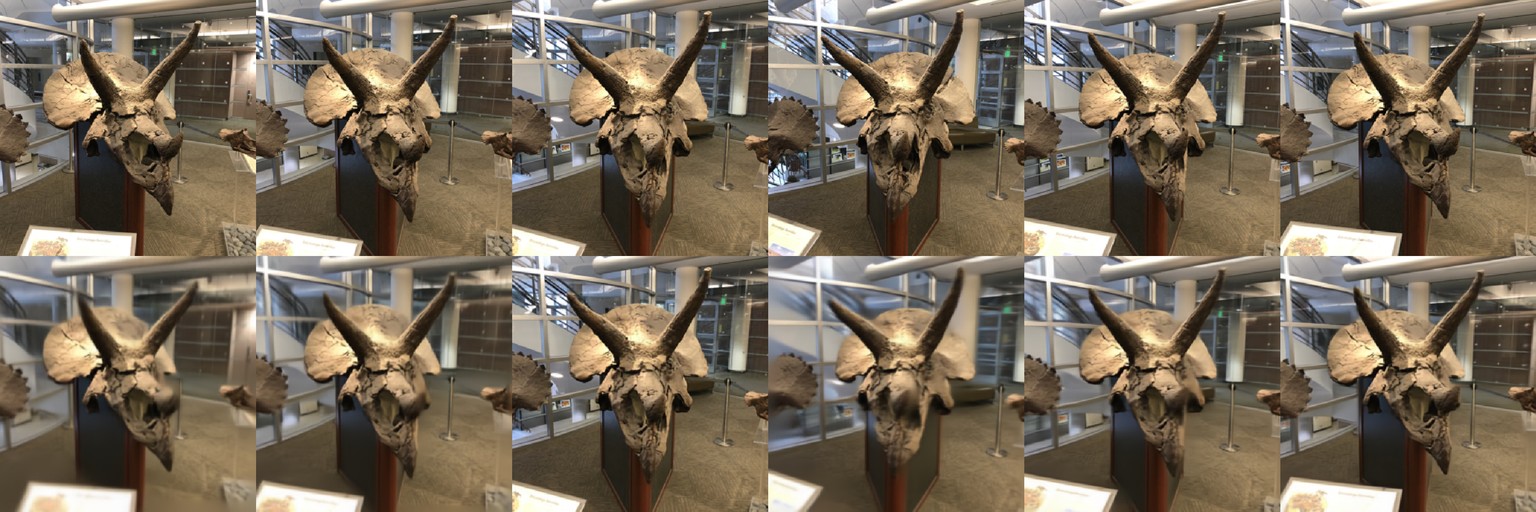}\\[2mm]
\sampleblockllfff{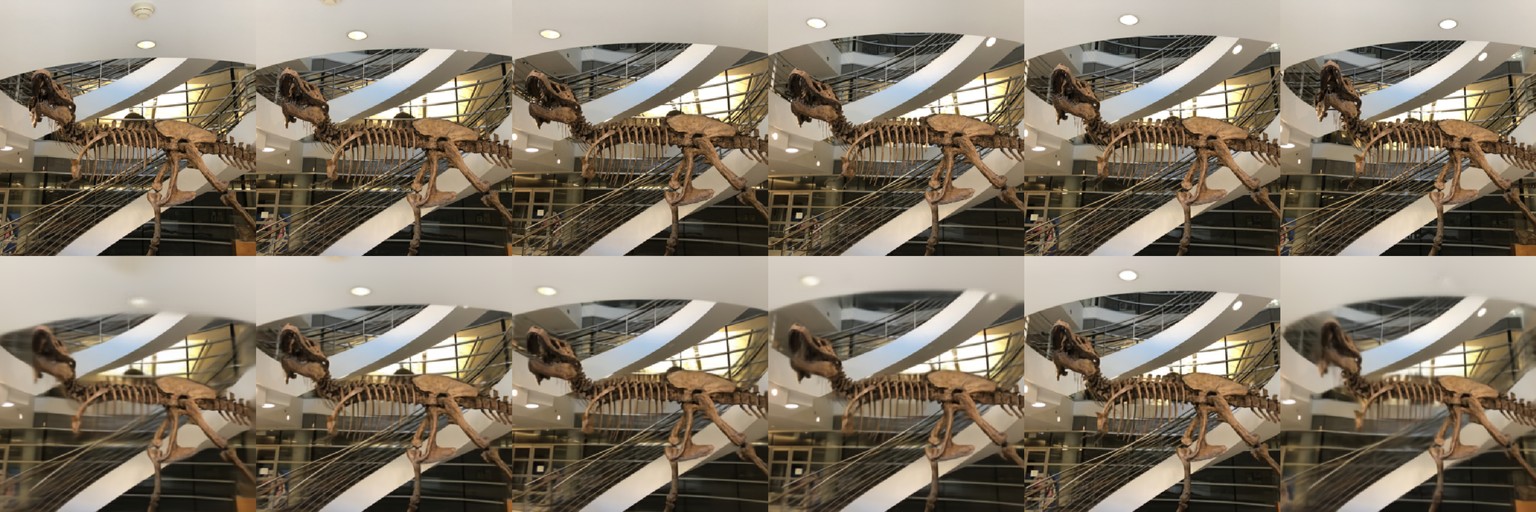}\\[2mm]
\sampleblockllfff{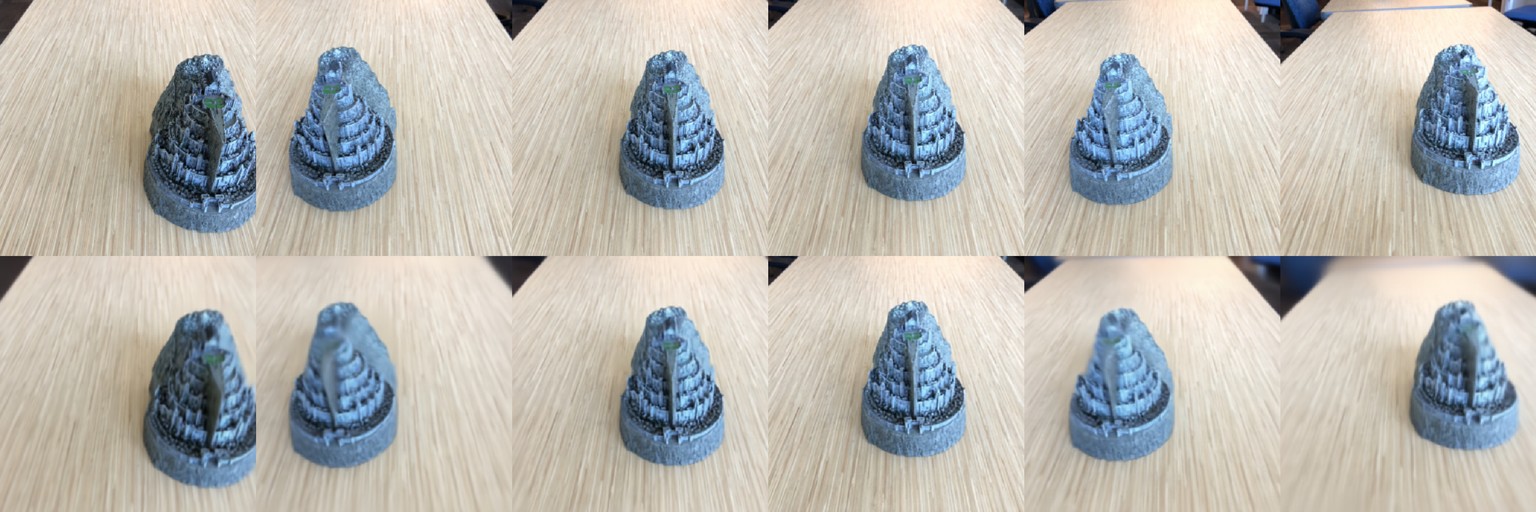}\\[2mm]
\sampleblockllfff{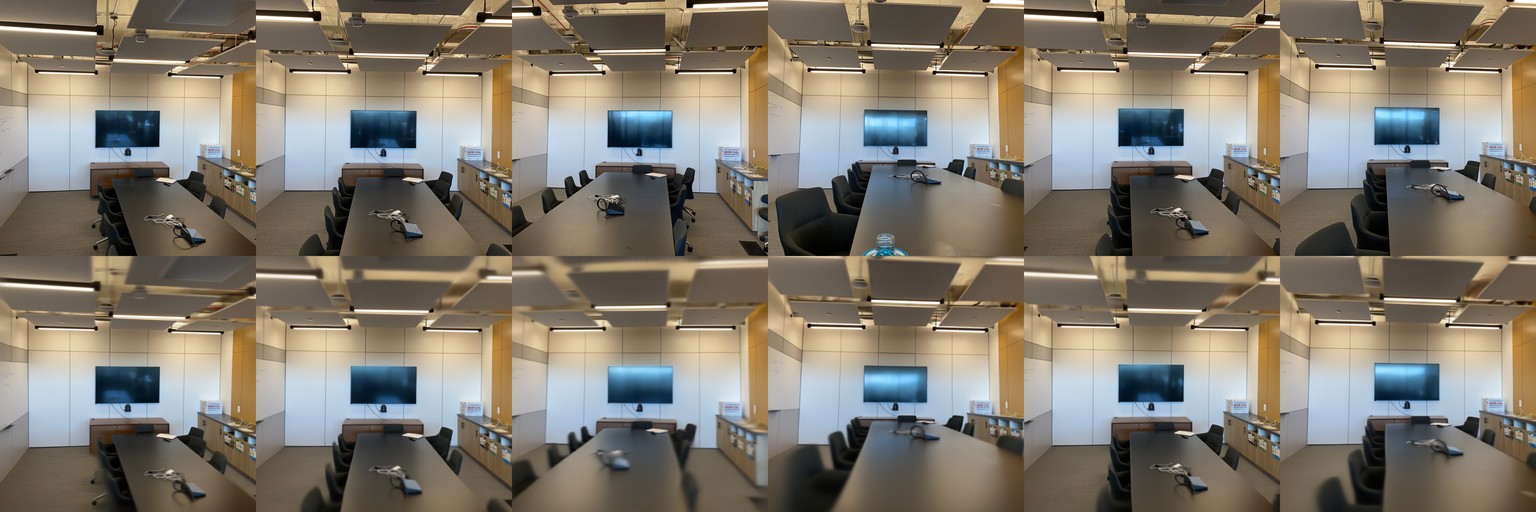}\\[2mm]
\caption{\textbf{\methodname trained on dynamic data, Zero-shot on static data} Novel View Synthesis samples from \methodname -L on LLFF dataset with 3 input views.} 
\label{fig:app_qual_llff}
\end{figure}
\clearpage

\section{Language Model Usage}
We employed large language models (Claude Opus 4.6, OpenAI GPT-5.2, Google Gemini 3 Pro) for text refinement purposes, including improving grammar and as inspiration for rephrasing sections.
They were also employed to provide feedback on early drafts and propose initial implementations for auxiliary utility functions not directly related to the paper's contributions (e.g., implementations of alternative camera intrinsics models), subsequently verified and reworked by the authors.
No scientific content, experimental results, or novel ideas were generated by LLMs -- all technical contributions were conceived, implemented, and verified by the authors.

\section{Author Contributions}
UP and SB co-led the project and developed the core method.
Beyond the core method, SB was primarily responsible for project coordination, implementation, and writing; UP was responsible for related work and all evaluations incl.\ baselines.
NS contributed to the method concept, infrastructure development, and writing; BO advised the project.

\section{Copyright}
\ifeccv\else
    The style used for this paper is adapted from the \href{https://arxiv.org/abs/2407.15595v2}{arXiv preprint \emph{Discrete Flow Matching} (Gat et al., 2024)}, licensed under \href{https://creativecommons.org/licenses/by/4.0/}{CC BY 4.0}.
    Throughout the paper and figures/plots, we use Fira Sans (licensed under the \href{https://openfontlicense.org/open-font-license-official-text/}{OFL v1.1}) for bold text.
\fi

\paragraph{Dataset Licenses}
Datasets used in this work are available under the following licenses:
\begin{itemize}
    \item \textbf{Segment Anything-Video~\cite{ravi2024sam2}:} \href{https://creativecommons.org/licenses/by/4.0/}{CC BY 4.0}.
    \item \textbf{SpatialVid~\cite{wang2025spatialvid}}: \href{https://creativecommons.org/licenses/by-nc-sa/4.0/}{CC BY NC SA 4.0}.
    \item \textbf{DL3DV-10K~\cite{ling2024dl3dv}:} DL3DV-10K Terms of use, and \href{https://creativecommons.org/licenses/by-nc/4.0/}{CC BY-NC 4.0}.
    \item \textbf{RE10k~\cite{zhou2018stereo}:} \href{https://creativecommons.org/licenses/by/4.0/}{CC BY 4.0}.
    \item \textbf{uCO3D~\cite{liu24uco3d}:} \href{https://creativecommons.org/licenses/by/4.0/}{CC BY 4.0}.
    \item \textbf{LLFF~\cite{mildenhall2019llff}:} \href{https://www.gnu.org/licenses/gpl-3.0.en.html}{GPL-3.0} on repository, but no explicit statement that this also applies to the data (only used for evaluation).
    \item \textbf{DTU MVS~\cite{jensen2014large}:} ``freely available'' (only used for evaluation).
    \item \textbf{CO3D~\cite{liu24uco3d}:} \href{https://creativecommons.org/licenses/by-nc/4.0/}{CC BY-NC 4.0}.
    \item \textbf{WildRGBD~\cite{xia2024rgbd}:} \href{https://opensource.org/license/mit}{MIT} on repository, but no explicit statement that this also applies to the data (only used for evaluation).
    \item \textbf{MipNeRF-360~\cite{barron2022mip}:} unknown (only used for evaluation).
    \item \textbf{Tanks \& Temples~\cite{knapitsch2017tanks}:} \href{https://creativecommons.org/licenses/by/4.0/}{CC BY 4.0}.
    \item \textbf{DAVIS~\cite{perazzi2016benchmark}:} \href{https://opensource.org/license/bsd-3-clause}{BSD 3-Clause}.
    \item \textbf{DyCheck~\cite{gao2022dynamic}:} \href{https://www.apache.org/licenses/LICENSE-2.0}{Apache 2.0}.
\end{itemize}

\end{document}